\documentclass[11pt]{article}
\usepackage{macros}
\usepackage{setspace}
\usepackage{bm}
\usepackage{adjustbox}
\makeatletter
\newenvironment{breakablealgorithm}
  {\par\medskip\noindent
   \refstepcounter{algorithm}%
   \hrule height.8pt depth0pt \kern2pt
   \renewcommand{\caption}[1]{%
     {\raggedright\textbf{\fname@algorithm~\thealgorithm}\enskip ##1\par}%
     \addcontentsline{loa}{algorithm}{\protect\numberline{\thealgorithm}##1}%
     \kern2pt\hrule\kern2pt}}
  {\kern2pt\hrule\par\medskip}
\makeatother

\providecommand{\meth}{}\renewcommand{\meth}[1]{\textsf{#1}}
\definecolor{tabgreen}{RGB}{0,128,64}
\definecolor{tabblue}{RGB}{31,94,170}
\definecolor{tabred}{RGB}{192,32,38}
\providecommand{\best}{}\renewcommand{\best}[1]{\textcolor{tabgreen}{\textbf{#1}}}
\providecommand{\better}{}\renewcommand{\better}[1]{\textcolor{tabblue}{#1}}
\providecommand{\worse}{}\renewcommand{\worse}[1]{\textcolor{tabred}{#1}}
\providecommand{\gain}{}\renewcommand{\gain}[1]{\textcolor{tabgreen}{#1}}

\providecommand{\cell}{}\renewcommand{\cell}[2]{\begin{tabular}{@{}c@{}}#1\\[-2pt]{\tiny(#2)}\end{tabular}}
\providecommand{\cellbold}{}\renewcommand{\cellbold}[2]{\cell{\textbf{#1}}{#2}}
\providecommand{\cellbest}{}\renewcommand{\cellbest}[2]{{\color{tabgreen}\cell{\textbf{#1}}{#2}}}
\providecommand{\cellbetter}{}\renewcommand{\cellbetter}[2]{{\color{tabblue}\cell{#1}{#2}}}
\providecommand{\cellworse}{}\renewcommand{\cellworse}[2]{{\color{tabred}\cell{#1}{#2}}}
\providecommand{\R}{}\renewcommand{\R}{\mathbb R}
\providecommand{\one}{}\renewcommand{\one}{\bm 1}
\providecommand{\norm}{}\renewcommand{\norm}[1]{\left\lVert#1\right\rVert}
\providecommand{\op}{}\renewcommand{\op}{\mathrm{op}}
\providecommand{\row}{}\renewcommand{\row}{\operatorname{row}}
\providecommand{\rank}{}\renewcommand{\rank}{\operatorname{rank}}
\providecommand{\diag}{}\renewcommand{\diag}{\operatorname{diag}}
\providecommand{\tM}{}\renewcommand{\tM}{\widetilde M}
\providecommand{\tY}{}\renewcommand{\tY}{\widetilde Y}
\providecommand{\tE}{}\renewcommand{\tE}{\widetilde{\mathcal E}}
\providecommand{\Dnorm}{}\renewcommand{\Dnorm}[1]{\left\lVert#1\right\rVert_D}
\providecommand{\ESS}{}\renewcommand{\ESS}{\operatorname{ESS}}
\providecommand{\lamt}{}\renewcommand{\lamt}{\lambda}
\providecommand{\se}{}\renewcommand{\se}{\operatorname{se}}

\providecommand{\outfig}{}\renewcommand{\outfig}[2]{\includegraphics[width=#2]{figures/#1.pdf}}

\newcommand{\TABLE}[3]{\centering{\setstretch{1}\adjustbox{max width=\textwidth}{#2}\par}\caption{#1 #3}}
\newcommand{\FIGURE}[3]{\centering#1\par\caption{#2 #3}}
\newcommand{\up}{\rule{0pt}{1.2em}}
\newcommand{\down}{\rule[-0.5em]{0pt}{0pt}}

\newcommand{\Halmos}{\ifmmode\qquad\square\else\unskip\nobreak\hfill$\square$\fi}
\renewenvironment{proof}[1]{\par\smallskip\noindent\textit{#1}\ }{\par\smallskip}
\let\endnote\footnote
\newcommand{\blfootnote}[1]{\begingroup\renewcommand\thefootnote{}\footnote{#1}\addtocounter{footnote}{-1}\endgroup}
\tcbset{prompttemplate, breakable}

\newtheorem{proposition}{\textbf{Proposition}}
\numberwithin{proposition}{section}
\theoremstyle{definition}
\newtheorem{example}{\textbf{Example}}
\newtheorem{problem}{\textbf{Problem}}
\newtheorem{remark}{\textbf{Remark}}
\numberwithin{example}{section}
\numberwithin{problem}{section}
\numberwithin{remark}{section}

\begin{document}

\title{SYN-DIGITS: A Synthetic Control Framework for \\ Calibrated Digital Twin Simulation}

\author{
    Grace Jiarui Fan\thanks{Finance Division, Columbia Business School.} \quad
    Chengpiao Huang\thanks{Department of IEOR, Columbia University} \quad
    Tianyi Peng\thanks{Decision, Risk, and Operations Division, Columbia Business School.} \\[0.3em]
    Kaizheng Wang\thanks{Department of IEOR and Data Science Institute, Columbia University} \quad
    Yuhang Wu\footnotemark[3]
}

\date{This version: October 30, 2026}

\maketitle
\vspace{-1em}

\begin{abstract}
Persona simulation based on large language models---often referred to as digital twin simulation---is increasingly used for market research, recommender systems, and social science research. However, these simulations often exhibit systematic bias and miscalibration relative to real human behavior. We propose SYN-DIGITS, a model-agnostic, post-hoc calibration framework that learns relationships among questions from digital twins and transfers them to predict human responses. We focus on a cold-start setting in which no human response to the target question is observed, so calibration must rely on digital twins and human data on reference questions. Our analysis shows that, though digital twins may not reproduce individual human responses accurately, they can capture nontrivial relationships among questions that can be effectively transferred to the human side. When individual-level data is unavailable and we only have access to marginal information on reference questions, our framework also supports distribution-level calibration for predicting the distribution of human responses on the target question. Across three datasets and thirteen persona constructions, experiments show that SYN-DIGITS achieves 11\% to 417\% relative improvements in individual-level correlation and 46\%--57\% relative reductions in distributional discrepancy compared to uncalibrated baselines.\blfootnote{Author names are ordered alphabetically. Code and data: \url{https://github.com/yw3453/syn-digits}.}
\end{abstract}


\section{Introduction}\label{sec:intro}

\subsection{Background}
Simulating human behavior is central to scientific inquiry and practical decision-making, yet it remains challenging because human responses are heterogeneous, context-dependent, and governed by latent structure. Classical approaches---such as agent-based models, cognitive architectures, discrete-choice models, and latent-variable frameworks---offer structured representations but face limitations in scalability and generality \citep{mcfadden1972conditional,laird1987soar,epstein1996growing, bonabeau2002agent,anderson2004integrated,train2009discrete, lord2012applications}.

Large language models (LLMs) have emerged as promising human behavior simulators---often called \emph{digital twins} (DTs)---because they generate coherent, context-sensitive responses across diverse domains \citep{aher2023using,horton2023large,tranchero2024theorizing, binz2025foundation,park2024generativeagentsimulations1000,toubia2025database}. However, LLMs are generally not optimized for faithful reproduction of human behavior. Na\"{i}ve deployment can therefore produce systematic bias, overconfidence, and concentrated response distributions \citep{santurkar2023whose,scherrer2023evaluating,rossi2024problems, gao2025take,li2026llm,hullman2026human,peng2026digitaltwinsfunhousemirrors}.

\subsection{Problem and Existing Approaches}\label{sec:prob_existing_approaches}
Suppose a firm wishes to optimize some product, for which it would be helpful to understand how users might respond to potential changes. During the ideation stage, it is not uncommon to consider many hypothetical potential changes, for which there is little or no user feedback data available without active, costly data collection. If a set of DTs can accurately simulate the firm's users' feedback to these potential changes, it would significantly speed up the process and reduce the cost of experimentation.

\begin{figure}[htbp]
\FIGURE
{\includegraphics[width=0.95\textwidth]{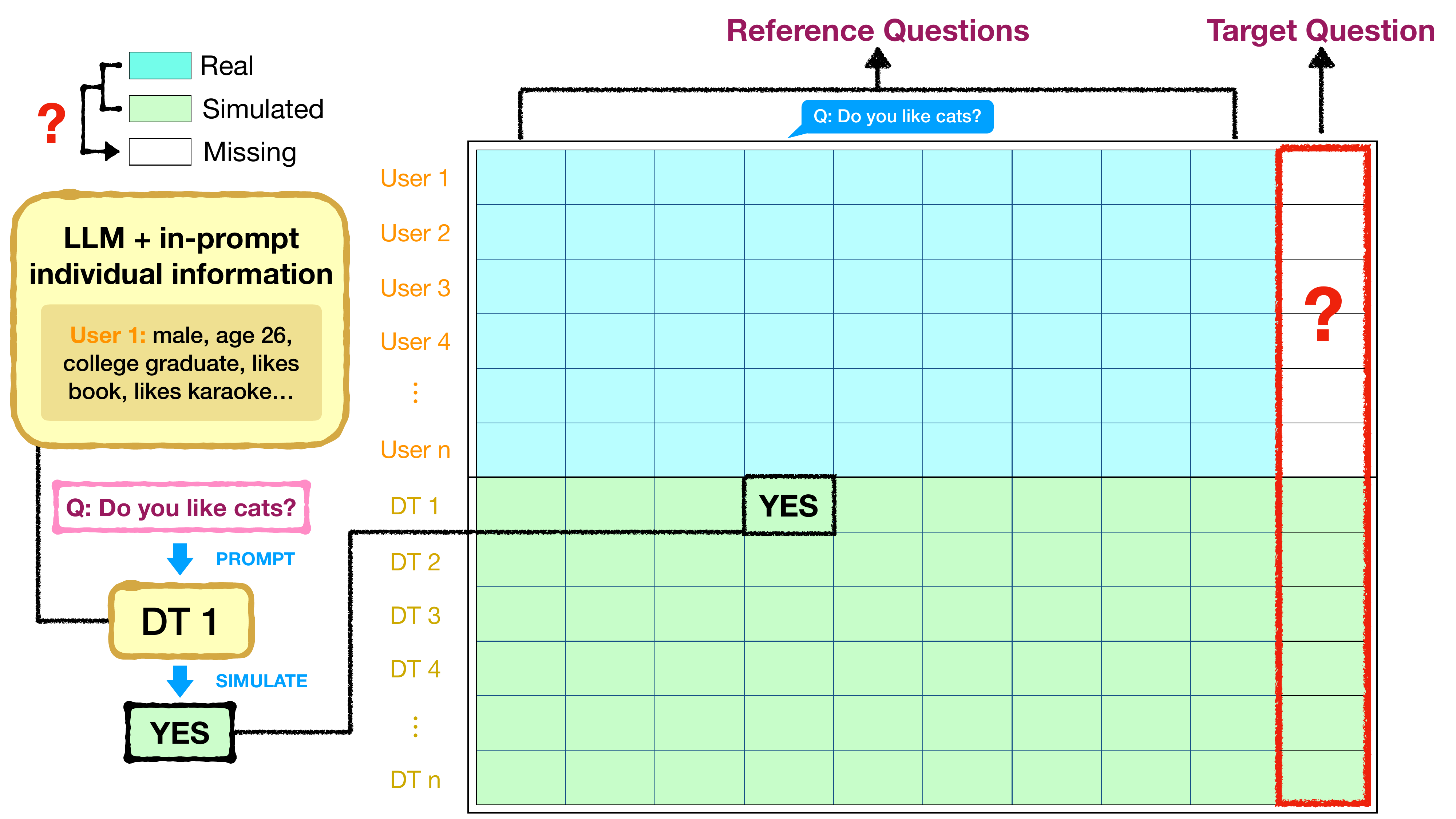}}
{The new-question calibration problem.\label{fig:motivation}}
{Human responses (blue) are observed on the reference questions, while DT responses (green) are available on both the reference and target questions, where a DT is an LLM prompted with some user's individual information. The goal is to simulate human responses to the target question (white with red question mark).}
\end{figure}

More precisely, suppose we have a set of \textbf{users}, and we are interested in simulating their responses to a set of \textbf{target questions} for which we have no human data. Often, however, we do have access to a set of \textbf{reference questions} for which we have collected human responses; for example, these might be previous product features that have been field-tested and are considered predictive of user responses to similar future changes. In addition to these real human data, we also have a set of \textbf{DTs} that have been constructed for each user by prompting an LLM with their \textbf{individual information}, such as demographic information and documented user preferences. Since these DTs are generative models, they can simulate responses to any question, including all the reference questions and the target question. We therefore ask:
\smallskip
\begin{center}
\textit{How can we use the DTs and human data on the reference questions \\ to simulate human responses to the target questions?}
\end{center}
\smallskip
See Figure \ref{fig:motivation} for an illustration. Note that this is a \emph{cold-start} problem, analogous to recommending a new item that no user has rated, so standard collaborative filtering or matrix completion on the human data alone has no anchor entries in the target column to leverage. Existing LLM-based approaches for predicting human target responses when they are \textit{entirely unobserved} include:
\begin{enumerate}
\item \textbf{Na\"{i}ve simulation:} directly prompt the digital twins to produce the target response and use it as a prediction. This is the predominant approach in existing digital twin frameworks \citep{park2024generativeagentsimulations1000, toubia2025database, peng2026digitaltwinsfunhousemirrors}, but it inherits the systematic biases of LLM simulation.
\item \textbf{Fine-tuning:} fine-tune the LLM on human responses to the reference questions, then prompt it on the target question \citep{cho2024llm,binz2025foundation,cao2025specializing,kolluri2025finetuning}. This is computationally expensive and can be infeasible in data-scarce settings. Even when fine-tuning is affordable, it can be brittle in practice; for instance, \citet{toubia2025database} find that fine-tuning can perform worse than na\"{i}ve simulation.
\item \textbf{In-context learning:} provide human responses to the reference questions as in-context examples when prompting the LLM. While cheaper than fine-tuning, this approach remains fragile and lacks principled mechanisms for correcting LLM-induced biases (see Section~\ref{sec:motivating_example} and Table~\ref{tab:movielens_methods}).
\end{enumerate}
Moreover, none of the three approaches explicitly exploits the \emph{cross-question structure} possibly shared by synthetic and real responses. We are therefore motivated to develop complementary calibration methods that are lightweight, data-efficient, model-agnostic, and amenable to theoretical analysis. 

\subsection{Core Idea and Contributions}
Our key observation is that vertically stacking the human and digital twin response matrices creates a matrix with a missing half-column that corresponds to the unobserved human responses to the target question (Figure~\ref{fig:motivation}). This viewpoint allows us to draw on the rich toolkit of matrix completion and synthetic control, which have a long history of success in recommender systems, healthcare, econometrics, and beyond \citep{abadie2010synthetic, mazumder2010spectral, candes2012exact, athey2021matrix, agarwal2026synthetic}. Building on this perspective, we propose \textbf{SYN-DIGITS} (\textbf{SYN}thetic Control Framework for Calibrated \textbf{DIGI}tal \textbf{T}win \textbf{S}imulation), a post-hoc calibration framework for any LLM-based digital twin simulator---whether zero-shot, prompt-engineered, or fine-tuned. Crucially, the DTs need not reproduce their human counterparts' responses accurately: calibration transfers the relationships among questions, which can survive substantial individual-level bias. More specifically, our contributions include:
\begin{enumerate}
\item \textbf{A modeling framework for digital twins.} We develop a coupled response-kernel model that  provides a general, principled characterization of the data generating process of digital twins.
\item \textbf{Calibration problems and methods.} Based on the model, we consider two cold-start calibration problems, in which no human data are available for the target question:
\begin{enumerate}
\item \textbf{Individual-level (Problem~\ref{prob:individual}):} simulating individual responses to the target question given individual responses to the reference questions. Transposing the response matrices, the same methods also simulate a new user's responses from reference users.
\item \textbf{Distribution-level (Problem~\ref{prob:distribution}):} simulating the marginal distribution of responses to the target question when only aggregate human marginal distributions are available for the reference questions.
\end{enumerate}
For the individual-level problem, we propose two families of calibration methods: (i) fit a predictive rule on the DT system and transfer it to the human system, or (ii) complete the human response matrix directly with the help of the DT matrix. For the distribution-level problem, we consider a mixture-of-persona approach that approximates the unobserved user population with a weighted combination of the DTs and ``intercept-like'' dummy atoms.
\item \textbf{Theoretical understanding.} We establish finite-sample error bounds for representative individual- and distribution-level calibration methods. The analysis yields a key insight: \emph{direct fidelity}---the extent to which a DT reproduces the same responses as its human counterpart---is not necessary for a simulator to be useful. Rather, \emph{transferable structure}---the extent to which the DT and human response systems encode compatible relationships among questions---is what matters for successful calibration.
\item \textbf{A systematic empirical evaluation.} We evaluate ten individual-level calibration methods on MovieLens and on thirteen Twin-2K-500 persona constructions spanning three LLMs and various prompting strategies, fine-tuning status, and persona content. The best method improves correlation from 11\% to 417\% over the uncalibrated DT. For the distribution-level problem, our approach reduces four discrepancies by 46\%--57\% on MovieLens and OpinionQA.
\end{enumerate}
The rest of the paper is organized as follows. Section~\ref{sec:litreview} reviews related work. Section~\ref{sec:framework} gives a warm-up account of the method and empirical evidence on MovieLens. Section~\ref{sec:individual_theory} develops the general model and theoretical guarantees for individual-level calibration methods, and conducts a comprehensive evaluation of the methods across Twin-2K-500 persona constructions. Section~\ref{sec:distsim} develops the distribution-level calibration method and provides both theoretical analysis and empirical verification. Section~\ref{sec:discussion} concludes with discussions of limitations and future work.

\section{Related Work}\label{sec:litreview}

\paragraph{Human digital twins.} The concept of a \textit{digital twin}---a virtual representation of a physical entity or system---originated in engineering and manufacturing and has since expanded to a wide range of applied domains. A particularly important recent extension is the concept of \textit{human digital twins}, which aims to construct faithful digital representations of human behavior and decision-making \citep{nguyen2022toward, lin2024human}. In consumer and market research, digital twins serve as AI-driven behavioral simulators of customers, valued for their speed, flexibility, and cost-effectiveness relative to traditional survey- and panel-based data collection \citep{ludwig2024large, toubia2025database, wang2026large}.

\paragraph{LLMs as human behavior simulators.} A rapidly growing line of work has explored LLMs as proxies for humans in surveys, experiments, and agent-based simulations, documenting both promises and limitations of this paradigm \citep{aher2023using, horton2023large, santurkar2023whose, scherrer2023evaluating, gao2024large, rossi2024problems, tranchero2024theorizing, binz2025foundation, cao2025specializing, gao2025take, hullman2026human, li2026llm, peng2026digitaltwinsfunhousemirrors}. Much of this literature focuses on constructing LLM-based simulators, showcasing downstream applications, or characterizing systematic failure modes such as distributional concentration, ideological biases, and sensitivity to prompt design. While these studies provide valuable insights into the capabilities and limitations of LLM-based simulation, they generally do not offer a versatile mechanism for systematically correcting the misalignment.

\paragraph{Closing the sim-to-real gap.} To improve simulation fidelity, several lines of work have emerged. One approach is fine-tuning, which re-trains the LLM on task- and domain-specific human data \citep{cho2024llm, binz2025foundation, cao2025specializing, orlikowski2025beyond, suh2025language}. While effective when sufficient training data is available, fine-tuning is computationally expensive and tightly coupled to the training distribution. A complementary line of work addresses the gap at the distribution level by reweighting LLM-generated samples to match observed human response distributions \citep{leng2024reduce, bui2025mixture, wang2026prompts}. These methods offer a lightweight alternative to fine-tuning for distributional calibration, but are not designed for individual-level calibration. A related line of work develops inference procedures that account for distributional misalignment between LLM and human responses \citep{huang2025uncertainty, iyengar2026model}. Other methods combine model-generated and limited human data for estimation and inference, including prediction-powered inference and PPI++ \citep{angelopoulos2023prediction,angelopoulos2026ppiplus}, generative augmented inference \citep{lu2026generative}, fine-tuning followed by rectification \citep{wang2025finetuning}, and optimal mixing of human and LLM-generated responses \citep{yin2026synthetic}. These methods use human labels for the target outcome to correct or combine model-generated information; our cold-start setting instead observes no human target responses and must transfer structure across questions.

\paragraph{A note on ``calibration.''} We emphasize that \emph{calibration} in our context refers to reducing the discrepancy between simulated and real human responses, i.e., improving simulation fidelity. This is related to but distinct from \emph{confidence calibration} in the machine learning literature, where methods such as Platt scaling \citep{platt1999probabilistic} and temperature scaling \citep{guo2017calibration} adjust a classifier's output probabilities so that predicted confidence levels match empirical accuracy. The goal here is to correct systematic biases in the \emph{content} of simulated responses, not in a model's self-assessed confidence. 

\paragraph{Synthetic control and matrix completion.} Our framework draws on synthetic control, which constructs weighted combinations of control units to approximate counterfactual outcomes and inform experimental design \citep{abadie2003economic, abadie2010synthetic, abadie2021using,abadie2026synthetic}, and on matrix completion and latent-factor methods for panel data \citep{candes2012exact, athey2021matrix, xiong2023latent}. \citet{agarwal2026synthetic} introduce synthetic interventions, which learn transferable structure across units; our application transfers structure from AI-generated to human responses rather than estimating causal effects. Related work develops causal matrix completion \citep{agarwal2023causal}, transfer from auxiliary panels \citep{duan2024target}, spectral and alternating-least-squares algorithms \citep{mazumder2010spectral,hastie2015matrix}, and guarantees under nonuniform observation, dependent sampling, and approximate low-rank structure \citep{du2017truncatedsvd, chen2021nonuniform,hamidi2022low,tang2026match}. The low-rank theory also connects to exact and noisy matrix recovery \citep{candesrecht2009exact,candesplan2010noise,recht2011simpler} and to group-sparse transfer across related factor models \citep{xu2021groupsparse}. We repurpose these tools to correct systematic misalignment in digital-twin simulations and develop guarantees that allow the two response systems to differ.

\section{Warm-up: Calibrating Digital Twins}\label{sec:framework}

\subsection{Problem Setup}\label{sec:problem_setup}

Consider tabular response data, where rows correspond to individuals and columns to questions or items. There are $n$ individuals, $m$ reference questions, and one target question indexed by $\star=d:=m+1$. Write the human response matrix as
$$
Y=[Y_{\mathrm{ref}}\mid Y_\star]\in\R^{n\times(m+1)},
$$
where the reference block $Y_{\mathrm{ref}} \in\R^{n\times m}$ is observed and the target column $Y_\star$ is entirely unobserved. For each individual, there is also a corresponding digital twin (DT), and DT responses are available for both the references and the target:
$$
 \tY
 =[\tY_{\mathrm{ref}}\mid\tY_\star]
 \in\R^{n\times(m+1)}.
$$
See Figure~\ref{fig:motivation} for an illustration. We consider the following problem. 

\begin{problem}[Individual-level calibration]\label{prob:individual}
Given the human reference responses $Y_{\mathrm{ref}}\in\R^{n\times m}$ and the DT responses $\tY=[\tY_{\mathrm{ref}}\mid\tY_\star]\in\R^{n\times(m+1)}$, predict the human target responses $Y_\star\in\R^n$.
\end{problem}

\subsection{Calibration Methods}\label{sec:calibration_methods}

The missing-half-column structure admits two natural families of algorithms:
\begin{itemize}
\item \textbf{Fit-and-transfer.} A predictive map is fitted on the DT system, using the DT reference responses to predict the DT target, and then applied to the human reference responses. Examples include ridge regression (\meth{Ridge}) \citep{benmichael2021augmented}, the lasso (\meth{Lasso}) \citep{hollingsworth2020tactics}, elastic net (\meth{EN}) \citep{doudchenko2016balancing}, neural network (\meth{NN}), simplex-constrained synthetic control (\meth{SC}) \citep{abadie2010synthetic}, and synthetic intervention (\meth{SI}) \citep{agarwal2026synthetic}.
\item \textbf{Direct matrix completion.} Vertically stacking the human and DT matrices, we can complete the missing human target block directly. Examples include hard SVD impute (\meth{HSV}), soft SVD impute (\meth{SSV}), and alternating least squares (\meth{ALS}) \citep{mazumder2010spectral, hastie2015matrix}. Alternatively, a synthetic-prior approach (\meth{SP}) can complete the human matrix alone, initializing all of its missing entries from the corresponding DT entries.
\end{itemize}
Algorithms~\ref{alg:fittransfer} and~\ref{alg:completion} summarize the two procedures. To account for question-specific scales, columns are standardized separately for $Y$ and $\tY$ before applying the algorithms. See Appendix~\ref{app:implementation_details_ind} for more implementation details, including preprocessing and cross-validation, and see Appendix~\ref{app:runtime} for computational costs.

\begin{algorithm}[htbp]
\caption{Fit-and-transfer}\label{alg:fittransfer}
\begin{algorithmic}[1]
\Require Human reference responses $Y_{\mathrm{ref}}$, DT responses
$\tY=[\tY_{\mathrm{ref}}\mid\tY_\star]$, method $f$
\State Fit $f$ to predict $\tY_\star$ using $\tY_{\mathrm{ref}}$.
\State Apply the fitted map to $Y_{\mathrm{ref}}$ and output the prediction $\widehat Y_\star$.
\end{algorithmic}
\end{algorithm}

\begin{algorithm}[htbp]
\caption{Direct matrix completion}\label{alg:completion}
\begin{algorithmic}[1]
\Require Human reference responses $Y_{\mathrm{ref}}$, DT responses
$\tY=[\tY_{\mathrm{ref}}\mid\tY_\star]$, completion method $f$
\State For \meth{HSV}, \meth{SSV}, and \meth{ALS}, form
$
 Z=\left[\begin{smallmatrix}Y_{\mathrm{ref}}&?\\
 \tY_{\mathrm{ref}}&\tY_\star\end{smallmatrix}\right].
$
For \meth{SP}, form $Z=Y$ with missing entries initialized from the corresponding DT entries in $\tY$.
\State Complete $Z$ by $f$, holding its observed entries fixed.
\State Extract the completed human target column $\widehat Y_\star$.
\end{algorithmic}
\end{algorithm}

\subsection{Evaluation on MovieLens}\label{sec:motivating_example}

At the outset, it is not obvious that Algorithms~\ref{alg:fittransfer} and~\ref{alg:completion} will perform well, especially if the DT answers have low fidelity. To evaluate their performance, we first conduct an illustrative experiment on the MovieLens-20M dataset \citep{harper2015movielens}. Because no single counterfactual learner uniformly dominates in panel settings \citep{athey2019ensemble}, we evaluate all ten methods mentioned above.

\paragraph{Data and digital twins.}
The dataset contains 20 million ratings for 27,000 movies by 138,000 users, where the ratings are from 0.5 to 5 stars in half-star increments. Our construction of the real-human matrix and the DTs proceeds as follows.
\begin{enumerate}
\item We select 500 users and 500 movies with the least missing data. Of the 500 movies, 250 are randomly selected for \textbf{evaluation}, and, for each user, their rating history on the remaining 250 is used as \textbf{individual information} for constructing the DTs (Figure~\ref{fig:motivation}).
\item The 500 users' real ratings on the 250 evaluation movies form the ground-truth matrix $Y \in \R^{500 \times 250}$, with 22\% of entries missing. 
\item We construct the DT matrix $\tY \in \mathbb{R}^{500 \times 250}$ by querying GPT-4.1-mini at temperature 0, as in \citet{toubia2025database}, to simulate each DT's rating on each of the 250 evaluation movies while maintaining the same missingness pattern as $Y$. The prompts are recorded in Appendix~\ref{app:movielen_individual}.
\end{enumerate}

\paragraph{Evaluation.}
Each movie is held out in turn and predicted from the remaining 249, i.e., $n=500$, $m=249$, and $Y_{\mathrm{ref}}, \tY_{\mathrm{ref}} \in \R^{500 \times 249}$ in each of the 250 folds (cf. Section~\ref{sec:problem_setup} and Figure~\ref{fig:motivation}). For fit-and-transfer methods, missing values in the covariates are imputed using the mean of the observed values in each column, separately for the human and DT matrix; direct matrix-completion methods retain missing entries. We report the mean Pearson correlation across the 250 targets following existing works \citep{park2024generativeagentsimulations1000, toubia2025database, peng2026digitaltwinsfunhousemirrors}. For completeness, we also report normalized root-mean-square error (nRMSE) in Appendix~\ref{app:nrmse}. More implementation details are available in Appendix~\ref{app:individual}.

\paragraph{Baselines.}
We compare against two LLM-based simulation baselines. In the zero-shot baseline (\meth{ZS})---na\"{i}ve simulation in Section~\ref{sec:prob_existing_approaches}---the DTs are prompted to rate the target movie based on the persona information alone. In the in-context baseline (\meth{IC}), for each target movie, the DTs are additionally provided with the user's available human ratings on the other 249 evaluation movies. Thus, \meth{IC} and our calibration methods use the same human reference information: \meth{IC} inserts it into the prompt, whereas calibration uses it after the simulator has produced the DT matrix.

\begin{table}[htbp]
\TABLE
{New-question correlation on MovieLens.\label{tab:movielens_methods}}
{
\setlength{\tabcolsep}{3pt}\begin{tabular}{@{}lcccccccccccc@{}}
\hline\up
& \multicolumn{2}{c}{\textbf{Baselines}} & \multicolumn{6}{c}{\textbf{Fit-and-transfer}} & \multicolumn{4}{c}{\textbf{Matrix Completion}} \\
& \meth{ZS} & \meth{IC} & \meth{Ridge} & \meth{Lasso} & \meth{EN} & \meth{NN} & \meth{SC} & \meth{SI} & \meth{HSV} & \meth{SSV} & \meth{ALS} & \meth{SP} \\ \hline\up
Corr. & .349 & .406 & \best{.516} & \better{.503} & \better{.514} & \better{.501} & \better{.444} & \better{.509} & \better{.471} & \better{.513} & \better{.513} & \better{.450} \\
S.E. & .005 & .005 & \best{.005} & \better{.006} & \better{.005} & \better{.005} & \better{.005} & \better{.005} & \better{.007} & \better{.005} & \better{.005} & \better{.005} \\
\%$\Delta$ \meth{ZS} & --- & +16 & \best{+48} & \better{+44} & \better{+47} & \better{+43} & \better{+27} & \better{+46} & \better{+35} & \better{+47} & \better{+47} & \better{+29} \\
\%$\Delta$ \meth{IC} & --- & --- & \best{+27} & \better{+24} & \better{+27} & \better{+23} & \better{+9} & \better{+25} & \better{+16} & \better{+26} & \better{+26} & \better{+11} \down\\
\hline
\end{tabular}
}
{Pearson correlation between predicted and human ratings; higher is better. Green bold marks the best method, and blue a method that beats the baseline of that row (\meth{ZS}, or \meth{IC} in the last row). Standard errors are computed across 250 target movies.}
\end{table}

\paragraph{Results.}
Table~\ref{tab:movielens_methods} shows that all ten methods outperform both \meth{ZS} ($.349$) and \meth{IC} ($.406$). \meth{Ridge} performs best ($.516$), closely followed by \meth{EN} ($.514$), \meth{SSV} ($.513$), and \meth{ALS} ($.513$). The comparison with \meth{IC} is especially informative: providing the LLM with the available human reference ratings raises correlation to $.406$, whereas the best calibration method reaches $.516$. The gain therefore does not arise simply from exposing the simulator to more human observations---it comes from transferring a relationship learned within the DT system.

\subsection{A Theoretical Case Study: Exact Transfer}\label{sec:exact_transfer}\label{sec:warmup}

What factors drive the performance of the calibration methods? Intuitively, the DT responses must share some cross-question structure with the human responses, even if individual responses differ. To formalize this intuition, suppose that the human and DT response matrices satisfy
$$
 Y=M +\mathcal E,
 \qquad
 \tY=\tM+\tE,
$$
where
$$
 M=[M_{\mathrm{ref}}\mid M_\star]
 \quad \text{and} \quad
 \tM=[\tM_{\mathrm{ref}}\mid\tM_\star]
 \; \in\;\R^{n\times(m+1)}
$$
are fixed human and DT ground-truth mean matrices and $\mathcal E$ and $\tE$ are noise matrices. Entries of $\mathcal E$ are independent, mean-zero, and sub-Gaussian with scale at most $\sigma_{HU}$; each entry of $\tE$ averages $S$ independent mean-zero sub-Gaussian DT draws with one-draw scale at most $\sigma_{DT}$. Equivalently, $\tY$ is the average of $S$ independent draws from the DT system, analogous to repeated Monte Carlo draws from a stochastic simulator. No independence between the human and DT systems is required. 

\paragraph{Structural relationship and performance measure.}
Throughout the paper, structural relationship will be imposed on the ground-truth matrices $M$ and $\tM$, not on realized noisy response matrices $Y$ and $\tY$. This is the same signal-plus-noise viewpoint used in noisy matrix completion, where low-rank structure concerns the underlying mean matrix \citep[see, e.g.,][]{candesplan2010noise}. For any predictor $\widehat Y_\star\in\R^n$ of the target column, we measure performance by the root-mean-square error with respect to the population mean:
\begin{equation*}
 \operatorname{Err}(\widehat Y_\star)
 =n^{-1/2}\norm{\widehat Y_\star-M_\star}_2.
\end{equation*}

\paragraph{Exact transfer under ridge regression.}
Since \meth{Ridge} performs best on MovieLens, the following proposition analyzes Algorithm~\ref{alg:fittransfer} when $f$ is ridge regression. We focus on ridge in this warm-up section for concreteness and impose exact structural conditions; Section~\ref{sec:individual_theory} develops the general model and relaxes these conditions, and Proposition~\ref{prop:indcandidate} in Appendix~\ref{app:indridgeproof} extends the analysis to any linear fit-and-transfer method, including \meth{Lasso}, \meth{EN}, \meth{SC}, and \meth{SI}. More specifically, for $\lambda>0$, we first fit
$$
 \widehat\beta_{\lambda}\in\argmin_{\beta\in\R^m}
 \left\{n^{-1}\norm{\tY_{\mathrm{ref}}\beta-
 \tY_\star}_2^2+\lambda\norm\beta_2^2\right\},
$$
where the intercept is omitted because columns of $Y$ and $\tY$ are standardized (cf. Section~\ref{sec:calibration_methods}).\endnote{If the columns are not standardized, an all-one column can be added to the reference matrices without loss of generality.} Then, we predict the target response:
$$
 \widehat Y_\star = Y_{\mathrm{ref}} \widehat\beta_{\lambda}.
$$
The following proposition gives intuitions for when the transfer performs well; proof is in Appendix~\ref{app:indexactproof}. We use $\row(\cdot)$ to denote the row space of a matrix.

\begin{proposition}[Exact question transfer]\label{prop:indexact}
Suppose $\row(M)\subseteq\row(\tM)$ and there exists $\beta^\star\in\R^m$ such that $\tM_{\mathrm{ref}}\beta^\star=\tM_\star$ and $\norm{\beta^\star}_2\lesssim m^{-1/2}$. For $\alpha\in(0,1)$, define the DT finite-sample radius
\begin{align}
 \xi_{DT,S}(\alpha)
 =\frac{\sigma_{DT}}{\sqrt S}\left(1+\sqrt{(m+1)/n}
   +\sqrt{\log(4/\alpha)/n}\right). \label{eq:inddtfinitesample}
\end{align}
If
\begin{equation}\label{eq:indexactlambda}
 \lambda\asymp m\,\xi_{DT,S}(\alpha)^2,
\end{equation}
then, with probability at least $1-\alpha$,
\begin{equation}\label{eq:indexacthighprob}
 \operatorname{Err}(Y_{\mathrm{ref}} \widehat\beta_{\lambda})
 \lesssim
 \xi_{DT,S}(\alpha)
 +
 \sigma_{HU}\left\{
 \frac1{\sqrt m}+\frac1{\sqrt n}
 +\sqrt{\frac{\log(4/\alpha)}{mn}}
 \right\}.
\end{equation}
\end{proposition}

\paragraph{Discussion.}
Proposition~\ref{prop:indexact} says that, if there is a diffuse reference-to-target rule in the DT mean matrix, and the row space of the human mean matrix is a subset of that of the DT mean matrix, then the prediction error converges to zero as $S,m,n\to\infty$ with $m/(nS)\to 0$.\endnote{For the error to converge to zero, technically we also need $\norm{M\tM^\dagger}_{\op}$ to stay bounded.} Note that even a \textbf{perfect simulator}, which provides sampling access to the correct human response law, must estimate the population mean from repeated samples, with the usual Monte Carlo error decreasing with $S$.\endnote{Simulation cost often scales linearly with $S$ in our context because $S$ simulation samples require $S$ calls to LLM APIs. In our experiments, we set $S=1$, i.e., each DT response is based on a single query, as we find the performance to be already satisfactory.} 

\paragraph{Direct fidelity vs. cross-question patterns.}
When the row-space inclusion condition $\row(M)\subseteq\row(\tM)$ holds, $\tM z=0$ for some $z$ implies $Mz=0$, i.e., all linear relationships between the columns of $\tM$ are present between those of $M$. Therefore, if there exists $\beta^\star\in\R^m$ such that $\tM_{\mathrm{ref}}\beta^\star=\tM_\star$, this linear relationship can be learned in the DT system up to statistical errors and transferred to the human system. In this sense, $\row(M)\subseteq\row(\tM)$ can be interpreted as preserving the \emph{cross-question patterns}, which, together with the diffuseness condition discussed below, is sufficient for effective calibration and obviates the need for \emph{direct fidelity} or simulation accuracy. Analogous conditions are widely used in the synthetic control literature \citep{abadie2010synthetic, duan2024target, agarwal2026synthetic}. 

\paragraph{Diffuse reference-to-target rule.}
The diffuseness condition $\norm{\beta^\star}_2\lesssim m^{-1/2}$ requires that the target be represented by many reference questions (e.g., all entries are $O(1/m)$) rather than by a few dominant references. This is necessary for consistency because each human reference response is observed only once. Even if there is one reference question that is exactly the same as the target question and ridge regression successfully learns a coefficient that is 1 at this coordinate and 0 at all other coordinates, the error would not vanish because we only sample the human responses once on that reference question, whereas the error compares against the population mean. This is also connected to the standard \emph{incoherence} assumption in the matrix completion literature \citep[see, e.g.,][]{candes2012exact}. For $\tilde r=\dim(\row(\tM))$ and $P$ the projection onto $\row(\tM)$, define
$$
 \mu_{\mathrm{inc}}:=\frac{d}{\tilde r}\max_{j\in[d]} \norm{P e_j}_2^2,
$$
which the usual incoherence assumption bounds by a constant. In our case, if $\tilde r$ is bounded, this assumption would imply $\norm{\beta^\star}_2=O(m^{-1/2})$ because the minimum-norm $\beta^\star$ satisfies $\norm{\beta^\star}_2^2 = \tfrac{\norm{P e_\star}_2^2}{1 - \norm{P e_\star}_2^2}$ (cf. Appendix~\ref{app:candidate_subspaces_defects}).

\subsection{Row-space Alignment}\label{sec:alignment}

Given the effectiveness of calibration on MovieLens and Proposition~\ref{prop:indexact}, can we observe empirical evidence for row-space inclusion? Unfortunately, the condition $\row(M)\subseteq\row(\tM)$ concerns the mean matrices, which are unobservable, and it is not directly testable on the realized matrices either: $Y$ and $\tY$ are typically full rank due to noise, so data-level inclusion would hold trivially. We therefore examine a testable proxy built in two steps. First, we consider the stronger, symmetric condition of row-space equality, $\row(M)=\row(\tM)$, which implies inclusion. Second, we replace the mean row spaces by their estimable proxies---the leading right singular subspaces of $Y$ and $\tY$---and measure their alignment.

\paragraph{Subspace alignment statistic.}
For an integer $r_0 \in [d]$, consider the $r_0$-dimensional leading right singular subspaces of the standardized response matrices $Y$ and $\tY$ (see Appendix~\ref{app:alignment} for the preprocessing), which estimate the leading question-space directions of the two systems. Between these subspaces, there are $r_0$ principal angles, and we measure the alignment between them by
\begin{equation}\label{eq:alignment}
 A(r_0)
 =\frac1{r_0}\sum_{j=1}^{r_0}\cos^2\theta_j,
\end{equation}
which is one when the two subspaces coincide and zero when they are orthogonal. We note, however, that even unrelated subspaces have nonzero alignment: Proposition~\ref{prop:haar} in Appendix~\ref{app:haar-random} shows that a Haar (uniformly) random $r_0$-dimensional subspace has expected alignment $r_0 / d$ with a fixed $r_0$-dimensional subspace. We therefore use this value as the Haar baseline for unrelated subspaces.

\begin{figure}[htbp]
\FIGURE
{\outfig{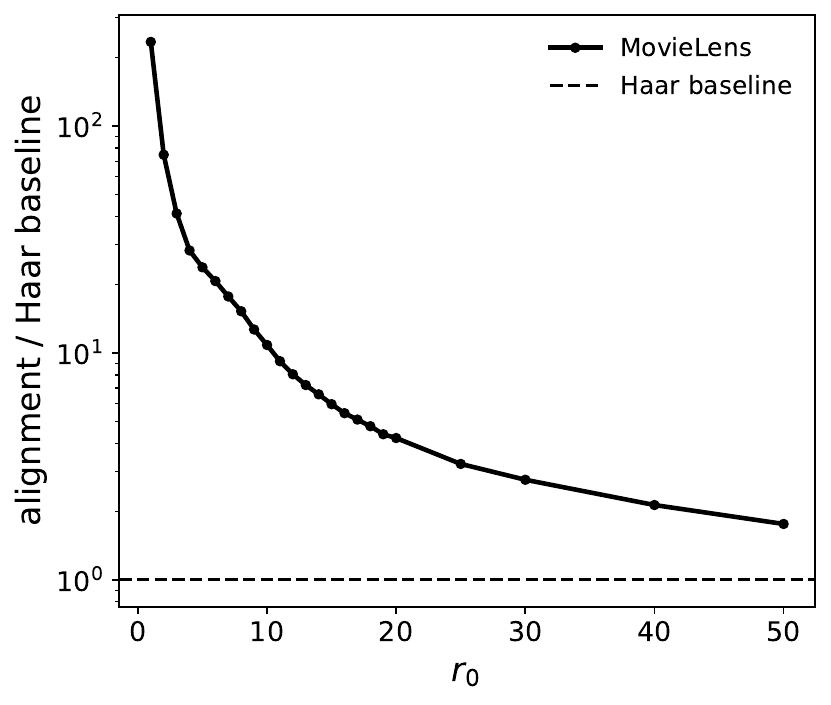}{0.56\textwidth}}
{Question-space alignment on MovieLens relative to the Haar
baseline.\label{fig:alignment_question_movielens}}
{The statistic $A(r_0)$ is divided by the Haar baseline $r_0/d$, with
$d=250$, and shown on a log scale. The curve remains above the Haar
baseline through $r_0=50$. Appendix~\ref{app:alignment} tabulates the values plotted and describes the preprocessing.}
\end{figure}

\paragraph{Results.}
Figure~\ref{fig:alignment_question_movielens} shows that MovieLens alignment substantially exceeds the Haar benchmark throughout the leading 50 singular directions. We take this as evidence that the human and DT response systems share considerably more question-space geometry than two unrelated systems. In other words, the DTs capture a significant portion of the cross-question human response structure; in light of Proposition~\ref{prop:indexact}, this is consistent with the strong performance of calibration in Section~\ref{sec:motivating_example}.

\section{Individual-level Calibration}\label{sec:individual_theory}

Building on Section~\ref{sec:framework}, we now develop a model of how human and DT responses are generated, followed by more general theoretical results. We conclude the section with a systematic empirical evaluation on a second dataset.

\subsection{A Model of Human and DT Responses}\label{sec:commonmodel}

Let $\mathcal X$ denote the space of questions and $x_j\in\mathcal X$ denote a question. Each human response unit has a latent representation $u_i\in\mathcal U_{HU}$, and each DT response unit has a representation $\widetilde u_i\in\mathcal U_{DT}$. Conceptually, $u_i$ represents the underlying characteristics of the human, while the DT representation $\widetilde u_i$ is jointly determined by the personal information in the prompt and the LLM's interpretation of that information. Let $\mathcal A\subseteq\R^{d_{\mathcal A}}$ be the answer space for some integer $d_{\mathcal A}\geq1$. We write
\begin{equation*}
 f_{HU}(\,\cdot\mid u,x) \in\mathcal P(\mathcal A),
 \qquad f_{DT}(\,\cdot\mid \widetilde u,x) \in\mathcal P(\mathcal A)
\end{equation*}
as the human and DT conditional response kernels, where $\mathcal P(\mathcal A)$ is the set of all probability distributions over $\mathcal A$. For finite sets $I_{HU}$ of human response units, $I_{DT}$ of DT response units, and $J$ of questions, define
\begin{equation*}
 \mathcal L_{I_{HU},I_{DT},J}
 =\sigma\!\left(
   (u_i)_{i\in I_{HU}},
   (\widetilde u_i)_{i\in I_{DT}},
   (x_j)_{j\in J}
  \right).
\end{equation*}
Conditional on this $\sigma$-field, the human response law is
\begin{equation*}
 Y_{ij}\sim f_{HU}(\,\cdot\mid u_i,x_j),
 \qquad (i,j)\in I_{HU}\times J.
\end{equation*}
On the DT side, the simulator can be queried repeatedly at the same persona--question pair in a Monte Carlo fashion. Fix an integer $S\geq1$ and let
\begin{equation*}
 \widetilde Y_{ij}
 :=\frac1S\sum_{s=1}^S\widetilde Y_{ij}^{(s)},
 \quad
 \text{where}
 \quad 
 \widetilde Y_{ij}^{(s)}\sim
 f_{DT}(\,\cdot\mid\widetilde u_i,x_j),
 \quad
 (i,j,s)\in I_{DT}\times J\times[S].
\end{equation*}
The two kernels may differ: a simulator $f_{DT}$ can distort both the user representation and the response mechanism. We assume the kernels have finite first moments and write
\begin{align*}
 \mu_{ij}&=\int_{\mathcal A}y\,f_{HU}(dy\mid u_i,x_j),
 &\varepsilon_{ij}&=Y_{ij}-\mu_{ij},\\
 \widetilde\mu_{ij}&=\int_{\mathcal A}y\,f_{DT}(dy\mid\widetilde u_i,x_j),
 &\widetilde\varepsilon_{ij}^{(s)}
 &=\widetilde Y_{ij}^{(s)}-\widetilde\mu_{ij}.
\end{align*}
We also write $\widetilde\varepsilon_{ij} :=\tfrac1S\sum_{s=1}^S\widetilde\varepsilon_{ij}^{(s)}$ and $\widetilde Y_{ij}=\widetilde\mu_{ij}+\widetilde\varepsilon_{ij}$. Throughout the paper, we impose the following independence assumption on the data-generating process; no independence restriction is imposed \emph{between} the human and DT responses.

\begin{assumption}[Conditional independence and sub-Gaussian noise]
\label{asm:commonlocal}
Conditional on $\mathcal L$, the human responses $\{Y_{ij}:(i,j)\in I_{HU}\times J\}$ are mutually independent, and the repeated DT responses $\{\widetilde Y_{ij}^{(s)}:(i,j,s)\in I_{DT}\times J\times[S]\}$ are mutually independent. Moreover, for constants $\sigma_{HU},\sigma_{DT}<\infty$,
$$
 \sup_{\norm{h}_2=1}
 \norm{h^\top\varepsilon_{ij}}_{\psi_2\mid\mathcal L}
 \leq\sigma_{HU},
 \qquad
 \sup_{\norm{h}_2=1}
 \norm{h^\top\widetilde\varepsilon_{ij}^{(s)}}_{
       \psi_2\mid\mathcal L}
 \leq\sigma_{DT}
$$
uniformly over the corresponding indices, where $\norm{\cdot}_{\psi_2\mid\mathcal L}$ denotes the conditional sub-Gaussian norm.
\end{assumption}

\paragraph{Model instantiation.}
The model is fairly general and can accommodate a wide range of response types and problem settings. In particular, it encompasses both individual-level (Section~\ref{sec:individual}) and distribution-level (Section~\ref{sec:distribution}) problems. We give two examples.

\begin{example}[Scalar answers]\label{ex:scalar}
Take $d_{\mathcal A}=1$. Then, $Y_{ij},\tY_{ij}\in\R$. The conditional-response model includes standard latent-factor specifications \citep{KBV09, candes2012exact, xiong2023latent}. For example, if $v_j\in\mathcal Q_{HU}$ and $\widetilde v_j\in\mathcal Q_{DT}$ are human and DT representations of question $x_j$, a scalar bilinear model is
$$
 \mu_{ij}=u_i^\top v_j,
 \qquad
 \widetilde\mu_{ij}=\widetilde u_i^\top\widetilde v_j.
$$
More generally, one may take
$$
 \mu_{ij}=g_{HU}(u_i^\top v_j),
 \qquad
 \widetilde\mu_{ij}=g_{DT}(\widetilde u_i^\top\widetilde v_j),
$$
where the link functions are appropriate to the answer space. This includes linear-probability models when the means remain in $[0,1]$, and binary logit or probit models.
\end{example}

\begin{example}[Categorical answers]\label{ex:categorical}
Take $d_{\mathcal A}=K$ and set $\mathcal A=\{e_1,\ldots,e_K\}\subseteq\R^K$; sub-Gaussian noise is thus automatic. A simple specification is an affine probability-factor model
$$
 \mu_{ij}=b_{HU}(u_i)+G_{HU}(u_i)v_j,
 \qquad
 \widetilde\mu_{ij}=b_{DT}(\widetilde u_i)+G_{DT}(\widetilde u_i)\widetilde v_j,
$$
where $b_{HU}(u_i)$ and $b_{DT}(\widetilde u_i)$ are intercepts, $G_{HU}(u_i)$ and $G_{DT}(\widetilde u_i)$ are coefficient matrices, and $\mu_{ij}$ and $\widetilde\mu_{ij}$ are simplex-valued vectors. More generally, categorical latent-factor models may assign category-specific score functions
$$
 s_{k,HU}:\mathcal U_{HU}\times\mathcal Q_{HU}\to\R,
 \qquad
 s_{k,DT}:\mathcal U_{DT}\times\mathcal Q_{DT}\to\R,
 \qquad k\in[K],
$$
and, for example, use multinomial-logit probabilities
$$
 \mathbb P(C_{ij}=k\mid u_i,v_j)
 =\frac{\exp\{s_{k,HU}(u_i,v_j)\}}
 {\sum_{\ell=1}^K\exp\{s_{\ell,HU}(u_i,v_j)\}},
 \qquad
 \mathbb P(\widetilde C_{ij}^{(s)}=k\mid\widetilde u_i,\widetilde v_j)
 =\frac{\exp\{s_{k,DT}(\widetilde u_i,\widetilde v_j)\}}
 {\sum_{\ell=1}^K\exp\{s_{\ell,DT}(\widetilde u_i,\widetilde v_j)\}},
$$
where $C_{ij}, \widetilde C_{ij}^{(s)} \in [K]$ are the human and DT category labels. With $u_i$ as a respondent's latent trait and $v_j$ as item parameters, item response theory models for polytomous items fit this framework: the nominal response and partial credit models take this multinomial-logit form, while the graded response model uses cumulative logits instead \citep{samejima1969estimation,bock1972estimating,masters1982rasch}.
\end{example}

\subsection{Individual-level Problem}\label{sec:individual}

In this section, we give an in-depth analysis of the fit-and-transfer methods to better understand the factors that drive their performance. In particular, we focus on ridge regression, though Proposition~\ref{prop:indcandidate} in Appendix~\ref{app:indridgeproof} applies to any linear fit-and-transfer method, including \meth{Ridge}, \meth{Lasso}, \meth{EN}, \meth{SC}, and \meth{SI}. Our main result in Theorem~\ref{thm:ridge} generalizes Proposition~\ref{prop:indexact} by relaxing the exact-transfer conditions.

\paragraph{Setup.}
Specialize Example~\ref{ex:scalar} to $I_{HU}=I_{DT}=[n]$ and $J=[d]$, and let $\mathcal L=\mathcal L_{[n],[n],[d]}$. Collect the conditional means into $M=(\mu_{ij})_{i\leq n,j\leq d}$ and $\tM=(\widetilde\mu_{ij})_{i\leq n,j\leq d}$, and partition
$$
 M=[M_{\mathrm{ref}}\mid M_\star],
 \qquad
 \tM=[\tM_{\mathrm{ref}}\mid \tM_\star],
$$
where the first $m$ columns correspond to the reference questions and $\star=d=m+1$ is the target question. Then, 
$$
Y = M + \mathcal E = [Y_{\mathrm{ref}}\mid Y_\star], \qquad 
 \tY = \tM + \tE = [\tY_{\mathrm{ref}}\mid \tY_\star],
$$
and $\mathcal E_{\mathrm{ref}}$, $\mathcal E_\star$, $\tE_{\mathrm{ref}}$, $\tE_\star$ follow the same partition. Put $r=\rank(M)$ and $\widetilde r=\rank(\tM)$, and assume $r,\widetilde r\geq1$, which is equivalent to requiring that $M, \tM \ne \mathbf{0}$. 

\paragraph{Connecting the two systems.}
Proposition~\ref{prop:indexact} tells us that two factors matter for good transfer performance: (i) the target question must be predictable from the reference questions in the DT system, and (ii) the question (row) spaces of the human and DT systems must be sufficiently aligned for the prediction learned in the DT system to transfer to the human system. We formalize these factors through a \textbf{candidate subspace} $\mathcal V\subseteq\R^d$ that serves as a bridge between the human and DT question spaces, which must satisfy:
\begin{itemize}
\item \textbf{Target predictability:} there exists a coefficient vector $\beta_{\mathcal V}\in\R^m$ such that, for every $v=(v_{1:m}^\top,v_\star)^\top\in\mathcal V$, $v_\star=\beta_{\mathcal V}^\top v_{1:m}$.\endnote{When there are multiple such vectors, we write $\beta_{\mathcal V}$ for the one with minimum $\ell_2$ norm.} Let $P_{\mathcal V}$ be the orthogonal projector onto $\mathcal V$ and define
$$
 \kappa_{\mathcal V} :=\norm{(I_d-P_{\mathcal V})e_\star}_2,
$$
which is the size of $e_\star$ orthogonal to the candidate subspace. It can be shown that target predictability is equivalent to $e_\star \notin \mathcal V$ or, equivalently, $\kappa_{\mathcal V}>0$ (cf. Appendix~\ref{app:candidate_subspaces_defects}). 
\item \textbf{Transferability:} linear maps are stably transferable to both the DT and the human system. Define
\begin{equation*}
 \gamma_{\mathcal V}
 :=
 \inf_{\substack{v\in\mathcal V\\ \norm v_2=1}}
 \frac{\norm{\tM v}_2}{\norm{M}_{\op}},
\end{equation*}
which is small when $\mathcal V$ contains spurious directions not well expressed in the DT system and thus the transfer can be unstable. $\gamma_{\mathcal V}$ is $\Omega(1)$ whenever the system is well-conditioned. For example, if $M=\widetilde M$ and $\mathcal V=\operatorname{row}(M)$, then $\gamma_{\mathcal V}=\sigma_r(M) / \sigma_1(M)$, where $\sigma_k(M)$ is the $k$-th largest singular value of $M$.

\end{itemize}
We therefore define the admissible family
\begin{equation*}
 \mathfrak V
 :=
 \left\{
 \mathcal V\text{ a linear subspace of }\R^d:
 \dim(\mathcal V)\geq1,\ 
 \kappa_{\mathcal V}>0,\ 
 \gamma_{\mathcal V}>0
 \right\},
\end{equation*}
which is nonempty under the standing assumption that $\tilde r \geq 1$ (cf. Appendix~\ref{app:candidate_subspaces_defects}). The performance of the transfer also depends on how \textbf{similar} $\mathcal V$ is to the human and DT question spaces. For $\mathcal V\in\mathfrak V$, define the human- and DT-side individual approximation defects by
\begin{align*}
 \Delta_{\mathrm{HU}}^{\mathrm{ind}}(\mathcal V)
 &=
 \frac1{\sqrt n}
 \norm{M(I_d-P_{\mathcal V})}_{\op},
\\
 \Delta_{\mathrm{DT}}^{\mathrm{ind}}(\mathcal V)
 &=
 \frac1{\sqrt n}
 \norm{\tM(I_d-P_{\mathcal V})}_{\op}.
\end{align*}
These defects measure the size of the human and DT (mean) signals lying in directions orthogonal to the candidate subspace (cf. Appendix~\ref{app:candidate_subspaces_defects}).

\begin{remark}[When the individual defects are small]\label{rem:inddefectexamples}
The two defects vanish if
$$
\row(M)\subseteq\mathcal V \quad \text{and} \quad \row(\tM)\subseteq\mathcal V
$$
for some common $\mathcal V\in\mathfrak V$. In that case, $M_{\mathrm{ref}}\beta_{\mathcal V}=M_\star$ and $\tM_{\mathrm{ref}}\beta_{\mathcal V}=\tM_\star$, so one common reference-to-target coefficient is exact in both systems. In particular, take the bilinear model from Example~\ref{ex:scalar}, which gives $M = U V ^\top$ and $\tM = \widetilde U \widetilde V ^\top$. Then, both defects are zero if
$$
\operatorname{col}(V) \subseteq \mathcal V \quad \text{and} \quad \operatorname{col}(\widetilde V) \subseteq \mathcal V,
$$
which also implies $v_\star=\sum_{j=1}^m\beta_{\mathcal V,j}v_j$ and $\widetilde v_\star=\sum_{j=1}^m\beta_{\mathcal V,j}\widetilde v_j$ for the same coefficient vector $\beta_{\mathcal V}$. Notably, under the bilinear model, whether the two defects vanish depends only on the question representations $v_i$ and $\widetilde v_i$, not on the user representations $u_i$ and $\widetilde u_i$. More generally, the magnitude of the defects is directly related to the alignment statistic \eqref{eq:alignment}. Suppose $e_\star\notin\row(\tM)$, i.e., the target question is predictable from the reference questions in the DT system. Then, we may take $\mathcal V=\row(\tM)\in \mathfrak V$ with $\Delta_{\mathrm{DT}}^{\mathrm{ind}}(\mathcal V)=0$. If $\dim(\row(M))=\dim(\row(\tM))=r_0$, then
$$
 \Delta_{\mathrm{HU}}^{\mathrm{ind}}(\mathcal V)
 \leq
 \frac{\norm{M}_{\op}}{\sqrt n}\sqrt{r_0 \cdot [1-A(r_0)]},
$$
which is small when the alignment $A(r_0)$ is close to one (cf. Appendix~\ref{app:candidate_subspaces_defects}).
\end{remark}

\paragraph{Excess risk.}
We evaluate target prediction by its conditional squared-error risk.  Let $Y_\star^{\mathrm{new}}\in\R^n$ be a fresh vector of human target responses, conditionally independent of all observed responses given $\mathcal L$, with $Y_{i\star}^{\mathrm{new}}\sim f_{HU}(\cdot\mid u_i,x_\star)$.\endnote{The fresh target vector $Y_\star^{\mathrm{new}}$ is only a device for defining the risk and does not enter the estimator.} For any realized predictor $g\in\R^n$, define risk
\begin{equation*}
 \mathcal R(g)
 :=\frac1n\mathbb E\!\left[
     \norm{g-Y_\star^{\mathrm{new}}}_2^2
     \mid\mathcal L,g
   \right].
\end{equation*}
Since $\mathbb E[Y_\star^{\mathrm{new}}\mid\mathcal L]=M_\star$, the Bayes predictor under squared loss is $M_\star$, and $\mathcal R(g)-\mathcal R(M_\star) =\tfrac1n\norm{g-M_\star}_2^2$. For a coefficient vector $\beta\in\R^m$, the predictor is $Y_{\mathrm{ref}}\beta$, and we therefore define its Bayes excess risk by
\begin{equation}\label{eq:indexcessrisk}
 \mathcal R^{\mathrm{exc}}(\beta)
 :=\mathcal R(Y_{\mathrm{ref}}\beta)-\mathcal R(M_\star)
 =\frac1n\norm{Y_{\mathrm{ref}}\beta-M_\star}_2^2.
\end{equation}

\paragraph{Excess risk bound for ridge regression.}
For $\lambda\geq0$, let
\begin{equation*}
 \widehat\beta_\lambda
 \in\argmin_{\beta\in\R^m}
 \left\{\widehat\tau_S(\beta)^2+\lambda\norm{\beta}_2^2\right\},
\end{equation*}
where the reference (training) error
\begin{equation*}
 \widehat\tau_S(\beta)
 :=\frac1{\sqrt n}
   \norm{\tY_{\mathrm{ref}}\beta-\tY_\star}_2.
\end{equation*}
When $\lambda=0$, $\widehat\beta_0$ denotes the minimum-$\ell_2$-norm least-squares solution. Recall the DT finite-sample radius \eqref{eq:inddtfinitesample}
$$
\xi_{DT,S}(\alpha)
 =\frac{\sigma_{DT}}{\sqrt S}\left(1+\sqrt{(m+1)/n}
   +\sqrt{\log(4/\alpha)/n}\right)
$$
and define the human finite-sample radius
\begin{equation*}
\xi_{HU}(\alpha)
 =\sigma_{HU}\left(1+\sqrt{m/n}
   +\sqrt{\log(4/\alpha)/n}\right).
\end{equation*}
The following theorem bounds the excess risk of ridge fit-and-transfer; the proof is in Appendix~\ref{app:indridgeproof}.

\begin{theorem}[Ridge transfer]
\label{thm:ridge}
Suppose Assumption~\ref{asm:commonlocal} holds and fix $\alpha\in(0,1)$. Then, for every $\eta>0$, there exists an $\mathcal L$-measurable $\lambda_\eta\geq0$ such that, conditioned on $\mathcal L$, with probability at least $1-\alpha$ over the realized human reference responses and repeated DT responses,
\begin{align}
 \sqrt{\mathcal R^{\mathrm{exc}}
        (\widehat\beta_{\lambda_\eta})}
 &\lesssim
 \inf_{\mathcal V\in\mathfrak V}
 \Bigg[
 (1+\norm{\beta_{\mathcal V}}_2)
 \left\{
   \Delta_{\mathrm{HU}}^{\mathrm{ind}}(\mathcal V)
   +\frac{1}{\gamma_{\mathcal V}}\left[
      \Delta_{\mathrm{DT}}^{\mathrm{ind}}(\mathcal V)
      +\xi_{DT,S}(\alpha)
   \right]
 \right\}
 +\norm{\beta_{\mathcal V}}_2\xi_{HU}(\alpha)
 \Bigg]
 +\eta.
 \label{eq:indoraclebound}
\end{align}
\end{theorem}

\paragraph{Discussion.}
The theorem does not assume that an optimizer over candidate subspaces exists, and the implementation neither constructs a candidate subspace nor solves the infimum. If some candidate $\mathcal V$ satisfies $\norm{\beta_{\mathcal V}}_2=O(m^{-1/2})$ and $\gamma_{\mathcal V}=\Omega(1)$, then, omitting the near-optimal term $\eta$ and dropping log and lower-order terms, the theorem reads
\begin{align*}
 \sqrt{\mathcal R^{\mathrm{exc}}
        (\widehat\beta_{\lambda_\eta})}
 \lesssim{}&
 \Delta_{\mathrm{HU}}^{\mathrm{ind}}(\mathcal V)
 +\Delta_{\mathrm{DT}}^{\mathrm{ind}}(\mathcal V)
 +\frac{1}{\sqrt S}
 +\frac1{\sqrt m}+\frac1{\sqrt n}
 +\sqrt{\frac {m}{nS}},
\end{align*}
which separates into structural misalignment between the human and DT systems (cf. Remark~\ref{rem:inddefectexamples}) and noise terms that vanish as $S,m,n\to\infty$ with $m/(nS)\to 0$. The assumption that $\norm{\beta_{\mathcal V}}_2=O(m^{-1/2})$ requires that the target is represented by many reference questions rather than by a few dominant references (see the discussion following Proposition~\ref{prop:indexact}).

\paragraph{Scale of $\lambda_\eta$.}
The proof gives a concrete instantiation of $\lambda_\eta$. For every $\eta>0$, there exists a coefficient budget $B_\eta>0$ such that restricting the candidate class in \eqref{eq:indoraclebound} to $\left\{ \mathcal V\in\mathfrak V, \,\norm{\beta_{\mathcal V}}_2\leq B_\eta\right\}$ yields an $\eta$-optimal infimum. For this budget, define the budget-constrained DT-side prediction error
$$
 e_{DT}(B_\eta)
 :=\min_{\norm\beta_2\leq B_\eta}
 \frac1{\sqrt n}
 \norm{\tM_{\mathrm{ref}}\beta-\tM_\star}_2.
$$
Then, omitting finite-sample errors, we approximately have
$$
\lambda_\eta \asymp \frac{e_{DT}(B_\eta)^2}{B_\eta^2}.
$$
Thus, $\lambda$ balances DT-side prediction error against the coefficient scale: systematic DT reference-to-target error favors stronger regularization, while a larger stable coefficient budget permits weaker regularization. In the experiments, $\lambda$ is selected by cross-validation.

\paragraph{The role of user representations.}
Theorem~\ref{thm:ridge} is conditional on the realized user representations and therefore imposes no distributional assumptions on $u_i$ or $\widetilde u_i$. This makes the guarantees applicable to arbitrary collections of users, with their characteristics entering only through the mean matrices $M$ and $\tM$. Additional assumptions on user representations can, however, give more concrete interpretations and guarantees. For example, suppose that the human--DT representation pairs $(u_i,\widetilde u_i)$ are sampled i.i.d.\ from a joint population distribution. Under suitable moment conditions, the individual approximation defects $\Delta_{\mathrm{HU}}^{\mathrm{ind}}(\mathcal V)$ and $\Delta_{\mathrm{DT}}^{\mathrm{ind}}(\mathcal V)$ converge to population-level approximation errors, and the stability condition $\gamma_{\mathcal V}>0$ can be guaranteed by sufficient diversity in the DT population. See Appendix~\ref{app:indpopulation} for a more detailed discussion.

\subsection{Evaluation on Twin-2K-500}\label{sec:twin2k}

We now evaluate the methods systematically on a second dataset. Twin-2K-500 \citep{toubia2025database} contains 2,058 U.S.\ participants and their responses to 123 demographic, psychological, behavioral, and economic questions, with $\sim$23\% missing values. The original study provides DT responses under thirteen persona constructions, which vary along persona content, persona format, LLM, and prompting strategy:
\begin{itemize}
\item \textbf{Text, GPT-4.1-mini}: complete survey responses as free text.
\item \textbf{Text, Gemini-Flash-2.5}: the same text persona on a different LLM.
\item \textbf{JSON, GPT-4.1-mini} and \textbf{JSON, GPT-4.1}: structured persona fields.
\item \textbf{Text + CoT} and \textbf{Text + repeat}: model instructed to reason or repeat the question before answering.
\item \textbf{Text, temperature 0.7}: more stochastic generation.
\item \textbf{JSON + PO}: uses OpenAI's Predicted Output feature for structured output generation.
\item \textbf{Fine-tuned GPT-4.1-mini}: fine-tuned on the survey responses of 500 participants.
\item \textbf{Demographics only}: 14 demographic variables rather than the full survey history.
\item \textbf{Summary} and \textbf{Summary + JSON}: a concise profile, alone or appended to structured persona fields.
\end{itemize}
All constructions use temperature 0 unless stated otherwise. The original study documents the exact prompt templates, persona encodings, and API settings. We follow the leave-one-question-out protocol of Section~\ref{sec:motivating_example} with the same preprocessing, cross-validation, and evaluation procedures. Questions on which every DT gives the same answer remain references but are not used as targets, so each construction has between 101 and 117 targets. Ten constructions cover all 2,058 participants; JSON + PO on GPT-4.1 covers 2,050, the fine-tuned construction 1,558 (its 500 training participants are excluded), and JSON on GPT-4.1-mini 1,000. The uncalibrated baseline (BL) is the raw DT response to the target. Appendix~\ref{app:individual} gives the complete protocol.

\begin{figure}[htbp]
\FIGURE
{\outfig{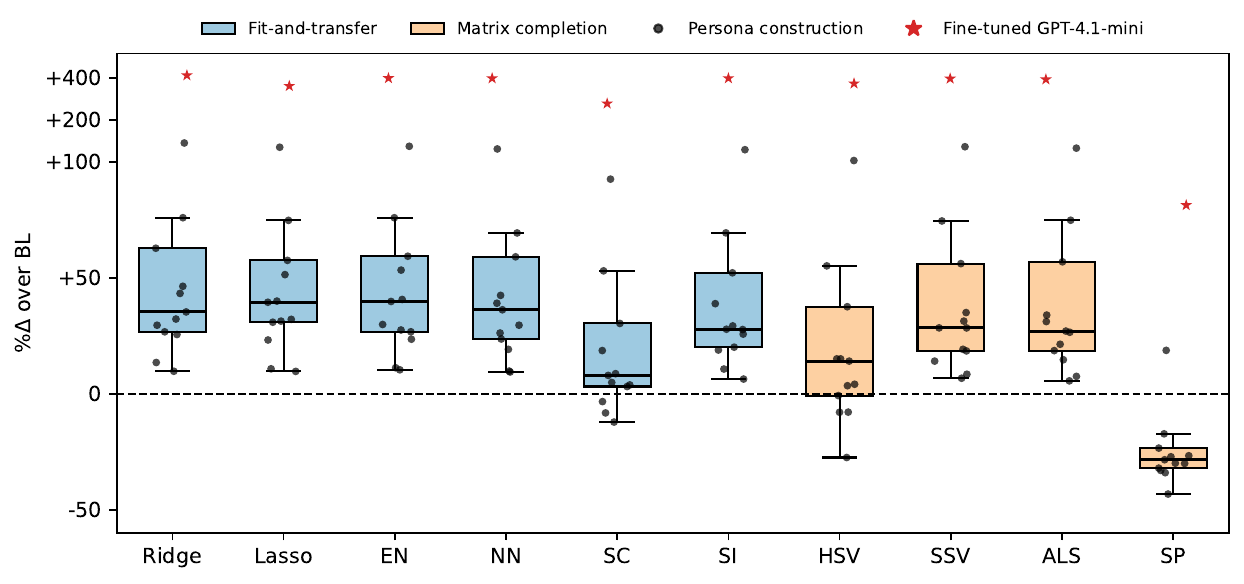}{0.95\textwidth}}
{Relative change in correlation over BL across the thirteen Twin-2K-500 persona constructions.\label{fig:2k500_gain_by_method}}
{Each box summarizes one method's \%$\Delta$ over the thirteen constructions in Table~\ref{tab:2k500_13x10}, and each point is one construction; the star marks the fine-tuned construction. The vertical axis is linear between $-100\%$ and $+100\%$ and logarithmic beyond.}
\end{figure}

\paragraph{Results.}
Figure~\ref{fig:2k500_gain_by_method} summarizes each method's relative change in correlation over BL across the thirteen constructions; Table~\ref{tab:2k500_13x10} in Appendix~\ref{app:2k500_results} reports the full results, and Appendix~\ref{app:nrmse} reports nRMSE. The best method improves on the uncalibrated DT for every persona construction. \meth{Ridge} is best for 8 constructions, \meth{Lasso} for 3, and \meth{EN} for 2. \meth{NN}, \meth{SI}, \meth{SSV}, and \meth{ALS} remain close to the best method on most constructions. \meth{SC}, \meth{HSV}, and \meth{SP} are weaker and fall below the uncalibrated DT on some constructions. Across the evaluated constructions, a regularized linear map is therefore a strong default.

\paragraph{Direct fidelity vs. transferable structure.}
The comparison across constructions illustrates why direct DT fidelity and transferable structure are separate objects---the best post-calibration results are not ordered by baseline quality. Especially noteworthy is the fine-tuned construction, which has the weakest baseline ($.058$) but achieves the highest post-calibration correlation ($.299$). Fine-tuning can therefore produce relationships among questions that are useful for transfer even when its direct predictions are poor. More broadly, DT responses are less dispersed than human responses on most questions and are shifted in location (Appendix~\ref{app:twin_moments}), yet calibration improves on the uncalibrated DT for every construction: \textbf{the relationships among questions transfer despite these individual-level biases.}

\begin{figure}[htbp]
\FIGURE
{\outfig{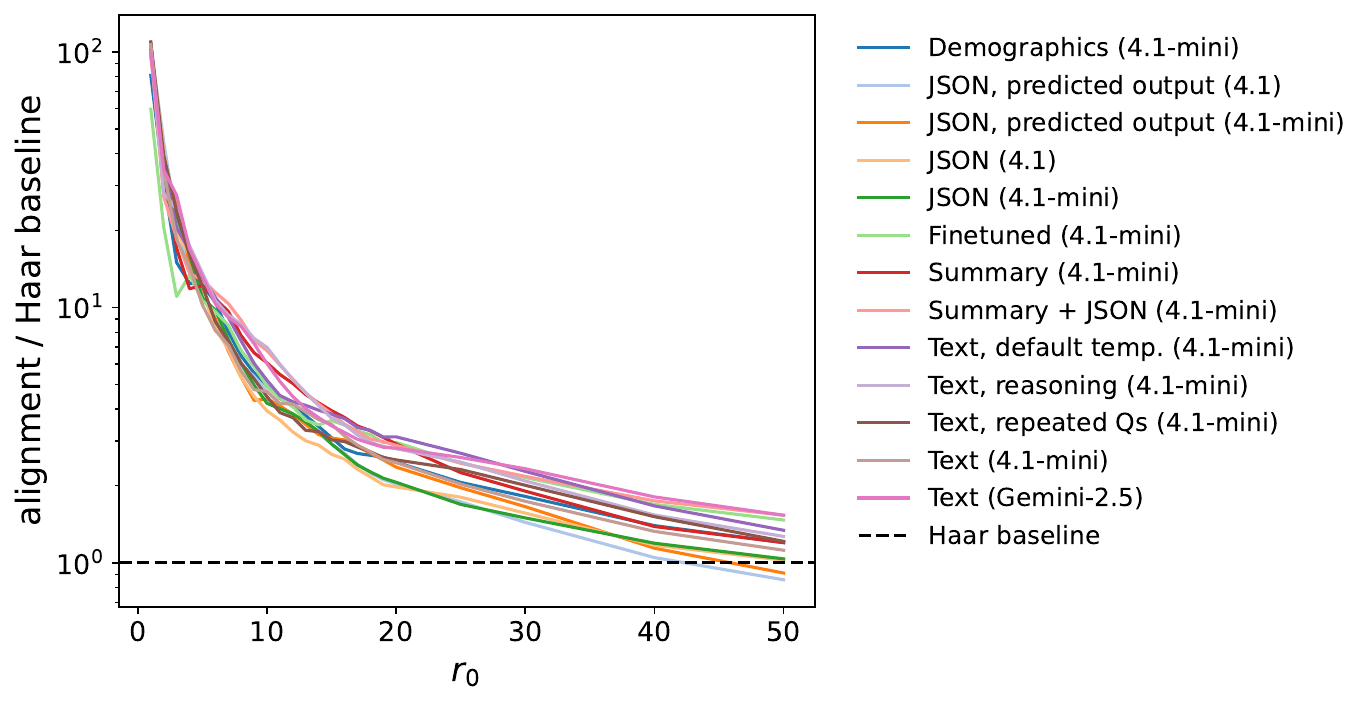}{0.90\textwidth}}
{Question-space alignment on Twin-2K-500 relative to the Haar baseline.\label{fig:alignment_question_twin2k}}
{The statistic $A(r_0)$ is divided by $r_0/d$, with $d=123$, one line per persona construction, on a log scale.}
\end{figure}

\paragraph{Question-space alignment.}
Similar to Figure~\ref{fig:alignment_question_movielens}, Figure~\ref{fig:alignment_question_twin2k} applies the statistic $A(r_0)$ from \eqref{eq:alignment} to every construction. Every construction is more aligned than the Haar baseline at every reported $r_0$, except the two JSON + PO constructions at $r_0=50$, indicating that human and DT responses share nonrandom question-space geometry. By Remark~\ref{rem:inddefectexamples}, the closer $A(r_0)$ is to one, the smaller the defects in Theorem~\ref{thm:ridge} for predictable targets. Appendix~\ref{app:alignment} tabulates the values plotted in Figure~\ref{fig:alignment_question_twin2k} and describes the preprocessing.

\begin{remark}[New users]\label{rem:indnewuser}
Our individual-level methods predict users' responses to a new question based on their responses to a set of reference questions. The same methods can address the user-side cold start, in which no human response of a new user is observed. Operating on the transposed response matrices, Problem~\ref{prob:individual} would regress the target user's DT response vector on those of the reference DTs across questions, and the fitted user weights are applied to the corresponding human users. Because Assumption~\ref{asm:commonlocal} is imposed entry-wise, the argument behind Theorem~\ref{thm:ridge} carries over to the transposed matrices, with candidate subspaces of user space in place of question space. Transfer then rests on user-space rather than question-space alignment between the two systems. Empirically, the best method improves new-user correlation over the uncalibrated DT by 18\% to 176\% across six Twin-2K-500 constructions and by 208\% on MovieLens. Appendix~\ref{app:newuser} gives the protocol, results, and evidence of user-space alignment.
\end{remark}

\section{Distribution-level Calibration}\label{sec:distsim}\label{sec:distribution}

We now turn to the distribution-level calibration problem. In many applications, individual responses are unavailable or unnecessary, and the primary object of interest is the \emph{marginal distribution} of responses across a population. Our model also provides a principled foundation for this distributional calibration task.

\subsection{Setup}\label{sec:dist_setup}

Section~\ref{sec:individual} builds on Example~\ref{ex:scalar} because the answers there are numeric. In this section, we consider categorical responses, and we therefore build on Example~\ref{ex:categorical}. Let $Q_{HU}$ be a target human-population law on $\mathcal U_{HU}$ and let $Q_{DT}$ be the source-population law on $\mathcal U_{DT}$. We reuse the notation from the individual-level setup and assume that there are $m$ reference questions with one target question indexed by $\star = d = m + 1$. For each question $j$, define the human and DT categorical response-probability vectors
\begin{equation}\label{eq:distpq}
 p_j(u)=\bigl(f_{HU}(\{e_k\}\mid u,x_j)\bigr)_{k=1}^K,
 \qquad
 q_j(\widetilde u)=\bigl(f_{DT}(\{e_k\}\mid\widetilde u,x_j)\bigr)_{k=1}^K
\end{equation}
and the human population marginals
$$
 P_j=\mathbb E_{u\sim Q_{HU}}[p_j(u)]\in\Delta^{K-1},
 \qquad
 P_+=[P_1\ \cdots\ P_m\ P_\star]\in\R^{K\times d}.
$$
In contrast to the individual-level setup, we assume $I_{HU}=\varnothing$, i.e., no individual human response units are sampled and no individual responses are observed. Instead, we assume access to marginals on the reference questions $P_1,\ldots,P_m$. We also assume access to a set of $n$ DTs ($I_{DT}=[n]$), such as the personas from Twin-2K-500, which can simulate responses to the reference and target questions. We consider the following problem.

\begin{problem}[Distribution-level calibration]\label{prob:distribution}
Given the human reference marginals $P_1,\ldots,P_m$ and the DT responses $\widetilde Y_{ij}\in\Delta^{K-1}$ for $(i,j)\in[n]\times[d]$, predict the human target marginal $P_\star$.
\end{problem}

\subsection{Algorithm}\label{sec:dist_algorithm}

\paragraph{Weighted ensemble.}
Problem~\ref{prob:distribution} offers no individual human responses to regress on, so we instead calibrate the marginal. Similar to the reweighting approaches of \citet{leng2024reduce}, \citet{bui2025mixture}, and \citet{wang2026prompts}, we approximate the human population by a weighted ensemble of the $n$ DTs, supplemented by $K$ \emph{dummy atoms}, the $k$-th of which always selects category $k$. For weights $(w,\pi)$ in
$$
 \Theta=\left\{(w,\pi)\in\R_+^n\times\R_+^K:\one_n^\top w+\one_K^\top\pi=1\right\},
$$
the ensemble marginal on question $j\in[d]$ is
\begin{equation}\label{eq:distpred}
 \widehat P_j(w,\pi)=\sum_{i=1}^nw_i\widetilde Y_{ij}+\pi\in\Delta^{K-1}.
\end{equation}
Reweighting corrects the composition of the DT population, while the dummy atoms (i) supply support on categories that the DTs rarely or never select, and (ii) serve as an intercept term for common tendencies.

\paragraph{Calibration.}
The weights are calibrated on the reference questions. Let $\mathcal H_0=\{x\in\R^K:\one_K^\top x=0\}$. Fix a norm $\Dnorm{\cdot}$ on $\mathcal H_0$, and define the discrepancy between probability vectors $P,Q$ by
\begin{equation*}
 D(P,Q)=\Dnorm{P-Q}.
\end{equation*}
Appendix~\ref{app:discrepancies} relates $D$ to total variation, Kolmogorov--Smirnov, and Wasserstein distances. For $\lambda\geq0$, the calibrated weights solve the convex program
\begin{equation}\label{eq:distalg}
 (\widehat w_\lambda,\widehat\pi_\lambda)
 \in
 \argmin_{(w,\pi)\in\Theta}
 \left\{
 \widehat\tau_{\mathrm{ref}}(w,\pi)^2+\lambda\norm w_2^2
 \right\},
\end{equation}
where the reference (training) loss is the per-question root-mean-square discrepancy
\begin{equation*}
 \widehat\tau_{\mathrm{ref}}(w,\pi)
 =
 \left\{\frac1m\sum_{j=1}^m
 D(P_j,\widehat P_j(w,\pi))^2\right\}^{1/2}.
\end{equation*}
When $\lambda=0$, $(\widehat w_0,\widehat\pi_0)$ denotes a minimizer having minimum $\norm w_2$.\endnote{Proposition~\ref{prop:diststruct} in Appendix~\ref{app:distalgproof} applies to any weights measurable with respect to the randomness in (E1)--(E2) of Section~\ref{sec:dist_theory}, not necessarily weights fitted by \eqref{eq:distalg}.} The prediction is $\widehat P_{\star,\lambda}=\widehat P_\star(\widehat w_\lambda,\widehat\pi_\lambda)$. Algorithm~\ref{alg:distensemble} summarizes the procedure, with $\lambda$ selected by cross-validation; Appendix~\ref{app:dist_protocol} gives the complete implementation, and Appendix~\ref{app:runtime} reports the computational cost.

\begin{algorithm}[htbp]
\caption{Weighted ensemble calibration}\label{alg:distensemble}
\begin{algorithmic}[1]
\Require Human reference marginals $\left\{P_j\right\}_{j \in [m]}$, DT responses $\left\{\widetilde Y_{ij}\right\}_{i \in [n], j \in [d]}$, discrepancy $D$, penalty $\lambda\geq0$
\State Fit $(\widehat w_\lambda,\widehat\pi_\lambda)$ by solving \eqref{eq:distalg} on the reference questions.
\State Output $\widehat P_{\star,\lambda}=\sum_{i=1}^n\widehat w_{\lambda,i}\widetilde Y_{i\star}+\widehat\pi_\lambda$.
\end{algorithmic}
\end{algorithm}

\subsection{Theoretical Analysis}\label{sec:dist_theory}

We now analyze Algorithm~\ref{alg:distensemble} under the categorical response model in Section~\ref{sec:dist_setup}. Intuitively, there are two factors that influence the performance of the algorithm: (i) whether the target question can be predicted from the reference questions; and (ii) whether the source distribution $Q_{DT}$, with the help of dummy atoms, can be reweighted to approximate the target distribution $Q_{HU}$. We formalize each of these factors below.

\paragraph{Target predictability and candidate subspaces.}
Analogous to the individual-level row spaces, the two row spaces in this context are $\row(P_+)$ and its DT counterpart, and the candidate subspaces that connect them are
\begin{equation*}
 \mathfrak V^{\mathrm{dist}}
 :=
 \left\{
 \mathcal V\text{ a linear subspace of }\R^d:\
 \kappa_{\mathcal V} > 0,\
 \one_d\in\mathcal V
 \right\},
\end{equation*}
where $\kappa_{\mathcal V} > 0$ has the same target predictability interpretation as in Section~\ref{sec:individual} and $\one_d\in\mathcal V$ because answers are categorical distributions (e.g., $\one_d \in \row(P_+)$). While the results in this section can be stated using $\mathfrak V^{\mathrm{dist}}$ and its associated discrepancy concepts as in Section~\ref{sec:individual}, it turns out that each $\mathcal V \in \mathfrak V^{\mathrm{dist}}$ induces a canonical $c \in \R^m$, defined by $z(c)=(c^\top,-1)^\top=-(I_d-P_{\mathcal V})e_\star/\kappa_{\mathcal V}^2$, such that $\one_m^\top c=1$ and $\mathcal V \subseteq z(c)^\perp$. Furthermore, the performance of a candidate $\mathcal V$ in Theorem~\ref{thm:distalg} below only depends on this $c$ and not its dimension or other properties. Therefore, without loss of generality and for ease of exposition, it suffices to consider the $(d-1)$-dimensional candidate subspaces
$$
 \left\{
 z(c)^\perp:\
 c\in \mathcal C_{\mathrm{aff}}
 \right\}, \quad \text{where} \quad \mathcal C_{\mathrm{aff}} :=\{c\in\R^m:\one_m^\top c=1\}
$$
is the affine coefficient class. See Appendix~\ref{app:inddistdimension} for a more detailed discussion.

\paragraph{Population defects.}
For $c \in \mathcal C_{\mathrm{aff}}$, define the human and DT population defects
\begin{equation*}
 \Delta_{\mathrm{HU}}^{\mathrm{dist}}(c)
 :=
 \Dnorm{\sum_{j=1}^m c_jP_j-P_\star},
 \quad
 \Delta_{\mathrm{DT}}^{\mathrm{dist}}(c)
 :=
 \left\{
 \mathbb E_{\widetilde u\sim Q_{DT}}
 \Dnorm{\sum_{j=1}^m c_jq_j(\widetilde u)-q_\star(\widetilde u)}^2
 \right\}^{1/2},
\end{equation*}
which measure how far the target distribution is from the affine combination of reference distributions. These are the distribution-level analogues of $\Delta_{\mathrm{HU}}^{\mathrm{ind}}$ and $\Delta_{\mathrm{DT}}^{\mathrm{ind}}$ in Section~\ref{sec:individual}. Negative coefficients may be interpreted as some reference questions being the ``opposite'' of the target question.

\begin{remark}[When the distribution-level defects are small]\label{rem:distdefectexamples}
The two defects are zero if 
$$
P_\star=\sum_jc_jP_j \quad \text{and} \quad q_\star(\widetilde u) = \sum_jc_jq_j(\widetilde u) \quad Q_{DT}\text{-almost surely}
$$
for some common coefficient vector $c\in\mathcal C_{\mathrm{aff}}$. In particular, take the affine probability-factor model from Example~\ref{ex:categorical}. Then, both defects are zero if 
$$
 v_\star=\sum_{j=1}^mc_jv_j,
 \qquad
 \widetilde v_\star=\sum_{j=1}^mc_j\widetilde v_j
$$
for a common coefficient vector $c\in\mathcal C_{\mathrm{aff}}$.
\end{remark}

\paragraph{Population reachability.}
An end-to-end error bound also requires a sampling assumption on the DT representations and a measure of how well they can approximate the target population. The following two assumptions formalize these requirements.

\begin{assumption}[DT sampling]\label{asm:disttypes}
The DT representations $\widetilde u_1,\ldots,\widetilde u_n$ are independent and identically distributed (i.i.d.) draws from $Q_{DT}$.
\end{assumption}

\begin{assumption}[Population reachability]
\label{asm:distcomparator}
There exist a measurable function $a:\mathcal U_{DT}\to[0,\bar a]$ and a vector $\pi^\circ\in\R_+^K$ such that
$$
 \mathbb E_{Q_{DT}}[a(\widetilde u)]
 =1-\one_K^\top\pi^\circ
 \in[s_0,1]
$$
and
$$
 \left\{
 \frac1m\sum_{j=1}^m
 D\!\left(
 P_j,
 \mathbb E_{Q_{DT}}[a(\widetilde u)q_j(\widetilde u)]+\pi^\circ
 \right)^2
 \right\}^{1/2}
 \leq\tau_{\mathrm{ref}}
$$
for some $\bar a\geq1$, $s_0\in(0,1]$, and $\tau_{\mathrm{ref}}\geq0$. 
\end{assumption}

\begin{remark}[Interpreting population reachability]\label{rem:interpretreach}
Reachability measures population coverage on the reference questions. If there is no response-law gap---i.e., $q_j=p_j$ for every question $j$---and $Q_{DT}=Q_{HU}$, then the DT system is a perfect simulator for the target population and $a\equiv1$, $\pi^\circ=0$ give $\tau_{\mathrm{ref}}=0$. More generally, if there is no response-law gap and $Q_{HU}\ll Q_{DT}$ with density ratio $dQ_{HU}/dQ_{DT}(u)$ bounded by $\bar a$, then importance reweighting gives
$$
 \mathbb E_{Q_{DT}}\left[q_j(u) \frac{dQ_{HU}}{dQ_{DT}}(u)\right]
 =\mathbb E_{Q_{HU}}[p_j(u)]
 =P_j,
$$
for every reference question $j$, and $a(u)=dQ_{HU}/dQ_{DT}(u)$ and $\pi^\circ=0$ again give $\tau_{\mathrm{ref}}=0$.
\end{remark}

\paragraph{Error bound for the calibrated marginal.}
Before proceeding to the error bound, we introduce some notation. For ease of exposition, we organize the relevant randomness into three stages:
\begin{itemize}
\item[(E1)] The DT representations $\widetilde u_1,\ldots,\widetilde u_n$ are realized.
\item[(E2)] Conditional on (E1), sample the reference responses $\{\widetilde Y_{ij}^{(s)}:i\leq n,\ j\leq m,\ s\leq S\}$ independently according to \eqref{eq:distpq}, and form the averages $\widetilde Y_{ij}$.
\item[(E3)] Conditional on (E1), independently of (E2), sample the target responses $\{\widetilde Y_{i\star}^{(s)}:i\leq n,\ s\leq S\}$ and form $\widetilde Y_{i\star}$.
\end{itemize}
Note that, since the target responses are not used for fitting, any implementable choice of $(w,\pi)$ is measurable with respect to (E1)--(E2). For $\delta\in(0,1)$, define the common finite-$n$ radius
\begin{equation*}
 \varepsilon_n(\delta)
 =\rho_D\sqrt K
 \left\{
 \sqrt{\frac{\log(56mK/\delta)}{n}}
 +\frac{\log(56mK/\delta)}{n}
 \right\},
\end{equation*}
where $\rho_D :=\sup_{x\in\mathcal H_0\setminus\{0\}} \tfrac{\Dnorm{x}}{\norm{x}_1} < \infty$ because $\mathcal H_0$ is finite-dimensional. For $(w,\pi)\in\Theta$, define the weighted DT response law and its corresponding reference loss:
\begin{equation*}
 \overline P_j(w,\pi)
 =\sum_{i=1}^nw_iq_j(\widetilde u_i)+\pi,
 \qquad
 \overline\tau_{\mathrm{ref}}(w,\pi)
 =
 \left\{\frac1m\sum_{j=1}^m
 D(P_j,\overline P_j(w,\pi))^2\right\}^{1/2}.
\end{equation*}
For a generic DT representation $\widetilde u\sim Q_{DT}$, define $\Sigma_j(\widetilde u) :=\diag(q_j(\widetilde u)) -q_j(\widetilde u)q_j(\widetilde u)^\top$ for $j \in [m]$ and $\Gamma_{\mathrm{ref}}(\widetilde u) :=\operatorname{blockdiag} \bigl(\Sigma_1(\widetilde u),\ldots,\Sigma_m(\widetilde u)\bigr)$ with DT reference noise level
\begin{equation*}
 \sigma_{\mathrm{ref}}^2
 :=\left\|
 \mathbb E_{\widetilde u\sim Q_{DT}}
 \Gamma_{\mathrm{ref}}(\widetilde u)
 \right\|_{\op}
 =\max_{j\leq m}
 \left\|
 \mathbb E_{\widetilde u\sim Q_{DT}}
 \Sigma_j(\widetilde u)
 \right\|_{\op},
\end{equation*}
which is at most $1/2$.\endnote{For any unit $v\in\R^K$, $v^\top\Sigma_j(\widetilde u)v$ is the variance, under $q_j(\widetilde u)$, of a random coordinate value drawn from $\{v_1,\ldots,v_K\}$. Its range is at most $\sqrt2$, so Popoviciu's inequality gives $0\preceq\Sigma_j(\widetilde u)\preceq\frac12I_K$ and hence $\sigma_{\mathrm{ref}}^2\leq1/2$.} The following theorem bounds the distribution-level error; the proof is in Appendix~\ref{app:distalgproof}.

\begin{theorem}[Distribution-level transfer]
\label{thm:distalg}
Suppose Assumptions~\ref{asm:commonlocal}, \ref{asm:disttypes}, and \ref{asm:distcomparator} hold. Fix $\delta\in(0,1)$ and suppose
\begin{equation}\label{eq:distpopn}
 n\gtrsim \frac{\bar a}{s_0}\log\left(\frac{42}{\delta}\right).
\end{equation}
Then, there exists a penalty $\lambda^\circ>0$ such that, with probability at least $1-\delta$ jointly over (E1)--(E3),
\begin{align}
 D(P_\star,\widehat P_{\star,\lambda^\circ})
 \lesssim
 \inf_{c\in\mathcal C_{\mathrm{aff}}}
 \Bigg[
 \sqrt m\norm c_2\,\tau_{\mathrm{ref}}
 +\Delta_{\mathrm{HU}}^{\mathrm{dist}}(c)
 +\Delta_{\mathrm{DT}}^{\mathrm{dist}}(c)
 +\rho_D\sqrt K\,\norm c_2
   \frac{\sigma_{\mathrm{ref}}}{\sqrt S}
 +\sqrt m\norm c_2\varepsilon_n(\delta)
 \Bigg].
 \label{eq:distoraclebound}
\end{align}
\end{theorem}

\paragraph{Discussion.}
As in Theorem~\ref{thm:ridge}, Theorem~\ref{thm:distalg} does not assume that an optimizer over affine comparators exists, and the implementation neither constructs a comparator nor solves the displayed infimum. Suppose some affine candidate $c$ satisfies $\norm c_2=O(m^{-1/2})$. Then, treating $K$ as fixed and dropping log and lower-order terms, the bound becomes
\begin{equation*}
 D(P_\star,\widehat P_{\star,\lambda^\circ})
 \lesssim
 \tau_{\mathrm{ref}}
 +\Delta_{\mathrm{HU}}^{\mathrm{dist}}(c)
 +\Delta_{\mathrm{DT}}^{\mathrm{dist}}(c)
 +\frac{1}{\sqrt{mS}}
 +\frac{1}{\sqrt{n}},
\end{equation*}
which separates into population coverage (cf. Remark~\ref{rem:interpretreach}), structural misalignment between the human and DT systems (cf. Remark~\ref{rem:distdefectexamples}), and noise terms that vanish as $mS,n\to\infty$. The assumption that $\norm c_2=O(m^{-1/2})$ is the same diffuseness requirement discussed after Proposition~\ref{prop:indexact} and Theorem~\ref{thm:ridge}. As at the individual level, the raw DT population need not match the human population. Successful transfer depends on (i) whether inter-question structures are captured by the DTs (small defects), and (ii) whether the DT population provides adequate coverage of the human population (small $\tau_{\mathrm{ref}}$).

\paragraph{Scale of $\lambda^\circ$.}
The proof gives a concrete instantiation of $\lambda^\circ$. Fix an eligible $(a,\pi^\circ)$ in Assumption~\ref{asm:distcomparator} and let $\nu:=\mathbb E_{Q_{DT}}[a(\widetilde u)^2]$ denote its transport variation. Omitting the finite-sample error $\varepsilon_n(\delta)$, we approximately have
$$
 \lambda^\circ\asymp\frac{n\,\tau_{\mathrm{ref}}^2}{\nu}.
$$
Thus, poorer population coverage favors stronger regularization, while more uneven transport weights permit weaker regularization. In the experiments, $\lambda$ is selected by cross-validation.

\subsection{Empirical Evaluation}\label{sec:dist_empirical}

\paragraph{Datasets and digital twins.}
We empirically evaluate the performance of the distribution-level calibration approach on the OpinionQA \citep{santurkar2023whose} and MovieLens datasets. OpinionQA contains $1{,}498$ multiple-choice survey questions sourced from Pew Research's American Trends Panel, covering topics such as social issues, politics, and science. We retain 489 questions with five response options ($K=5$), and use the aggregate distribution of available responses to these questions as the population marginals $P_+\in\R^{5\times 489}$.\endnote{There are between 1,014 and 10,221 respondents per question, with a median of about 3,400.} For the MovieLens dataset, on the same 500 movies as in Section~\ref{sec:motivating_example}, we use the aggregate rating distribution of the 500 users as the population marginals $P_+\in\R^{10\times 500}$ (ratings are from $\left\{0.5, 1, \ldots, 5\right\}$, so $K=10$). For the DTs, we use the 2,058 DTs from the Twin-2K-500 \textbf{Summary, GPT-4.1-mini} construction with temperature 0, so $n=2{,}058$. For OpinionQA, each DT is prompted to answer all $489$ multiple-choice questions; for MovieLens, each DT is prompted to rate all $500$ movies. The prompt templates are recorded in Appendix~\ref{app:distribution_prompts}.

\paragraph{Implementation and evaluation.}
In our experiment, we use the CDF-$\ell_2$ norm $D(P,Q)=\norm{C(P-Q)}_2$, where $C$ is the lower-triangular cumulative-sum matrix, as the objective for optimization, and report CDF-$\ell_2$, CDF-$\ell_1$, total variation (TV), and Kolmogorov--Smirnov (KS) discrepancies for evaluation (see Appendix~\ref{app:discrepancies} for definitions). We follow the same leave-one-question-out protocol of Section~\ref{sec:motivating_example} and use cross-validation to select $\lambda$ for each held-out target question. More implementation details are provided in Appendix~\ref{app:dist_protocol}. The uncalibrated baseline is the uniformly weighted average of the DT responses. In addition to the full ensemble \eqref{eq:distpred}, we also consider two ablations: DTs alone and dummy atoms alone.

\begin{table}[htbp]
\TABLE
{Distribution-level calibration.\label{tab:distsim_summary}}
{
\setlength{\tabcolsep}{4pt}\begin{tabular}{@{}lcccccccc@{}}
\toprule
 & \multicolumn{4}{c}{OpinionQA} & \multicolumn{4}{c}{MovieLens} \\
\cmidrule(lr){2-5}\cmidrule(lr){6-9}
Ensemble & CDF-$\ell_2$ & CDF-$\ell_1$ & TV & KS & CDF-$\ell_2$ & CDF-$\ell_1$ & TV & KS \\
\midrule
Both & \cellbest{0.1777}{0.0044} & \cellbest{0.2704}{0.0068} & \cellbest{0.1701}{0.0037} & \cellbest{0.1489}{0.0037} & \cellbest{0.1842}{0.0041} & \cellbest{0.4226}{0.0105} & \cellbest{0.1760}{0.0028} & \cellbest{0.1219}{0.0025} \\
Twins only & \cellbetter{0.2779}{0.0072} & \cellbetter{0.4112}{0.0107} & \cellbetter{0.2755}{0.0065} & \cellbetter{0.2388}{0.0064} & \cellbetter{0.2708}{0.0043} & \cellbetter{0.5870}{0.0100} & \cellbetter{0.3083}{0.0032} & \cellbetter{0.1979}{0.0035} \\
Dummy atoms only & \cellbetter{0.2683}{0.0073} & \cellbetter{0.4279}{0.0119} & \cellbetter{0.2256}{0.0053} & \cellbetter{0.2098}{0.0055} & \cellbetter{0.2803}{0.0076} & \cellbetter{0.6728}{0.0194} & \cellbetter{0.1835}{0.0039} & \cellbetter{0.1663}{0.0043} \\
Uncalibrated twins & \cell{0.3349}{0.0081} & \cell{0.4966}{0.0121} & \cell{0.3311}{0.0073} & \cell{0.2879}{0.0072} & \cell{0.4158}{0.0064} & \cell{0.9202}{0.0138} & \cell{0.3840}{0.0042} & \cell{0.2807}{0.0044} \\
\midrule
\%$\Delta$ best & \gain{-46.9} & \gain{-45.5} & \gain{-48.6} & \gain{-48.3} & \gain{-55.7} & \gain{-54.1} & \gain{-54.2} & \gain{-56.6} \\
\bottomrule
\end{tabular}
}
{Each question is held out in turn as the target. Mean discrepancy between the calibrated ensemble and the human population marginal, with standard error in parentheses. Lower is better. Green bold marks the best ensemble per column, and blue an ensemble that beats the uncalibrated DTs. The last row is the best ensemble's change relative to the uncalibrated DTs on each dataset.}
\end{table}

\paragraph{Results.}
Table~\ref{tab:distsim_summary} shows that combining DTs and dummy atoms has the best performance under every discrepancy on both datasets. Relative to the uncalibrated ensemble, CDF-$\ell_2$ falls by 46.9\% on OpinionQA and 55.7\% on MovieLens; the reductions under the other three measures are similar. A linear regression analogy explains this pattern: the dummy atoms act as an intercept that captures common tendencies across the population, and the DTs act as slopes that capture individual differences. Just as neither an intercept-only nor a slopes-only regression need to dominate the other, neither ablation dominates here, but the two are best combined. Appendix~\ref{app:dist_composition} reports the composition of the fitted ensembles, including effective sample sizes.

\begin{figure}[htbp]
\FIGURE
{\outfig{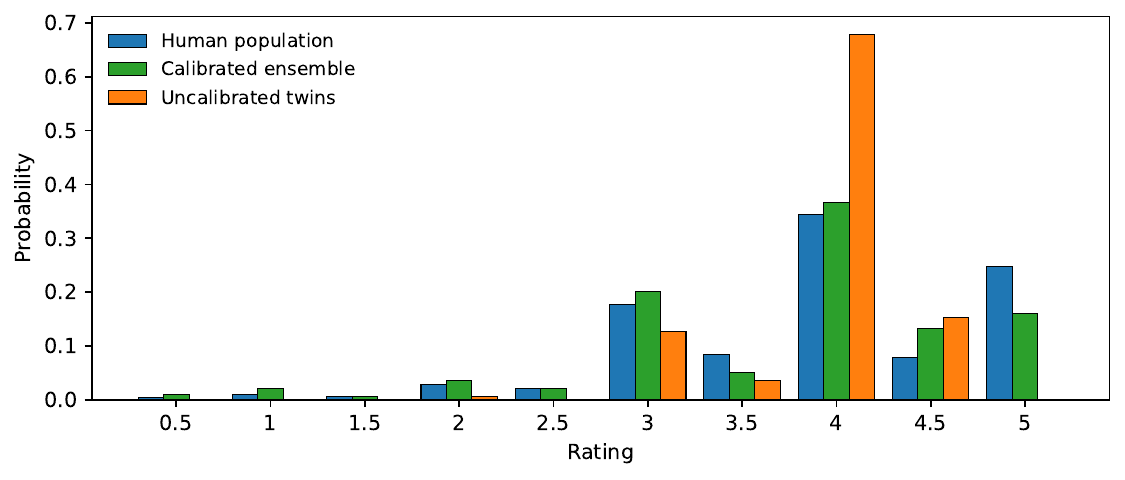}{0.82\textwidth}}
{Distribution calibration on one held-out MovieLens
film.\label{fig:distsim_example}}
{For \emph{Toy Story} (1995), the first item in dataset order, the calibrated ensemble's CDF-$\ell_2$ discrepancy is $0.1105$, compared with $0.3675$ for the uncalibrated DTs.}
\end{figure}

\paragraph{Illustration.}
Figure~\ref{fig:distsim_example} visually compares the uncalibrated and calibrated distributions on one target. The uncalibrated DTs place two-thirds of their mass on rating 4 and none on rating 5, which a quarter of the human population selects. The calibrated ensemble recovers both the shape and the missing support. Across the MovieLens movies, the uncalibrated DT population is also over-concentrated, while the calibrated ensemble restores the population spread (Appendix~\ref{app:dist_variance}).

\section{Discussion}\label{sec:discussion}

\paragraph{Summary of contributions.}
We present SYN-DIGITS, a practical, model-agnostic post-hoc calibration framework for digital twin simulation. The framework models human and DT systems with general response kernels and mild assumptions about their structure. Based on the model, we propose solutions to two calibration problems---individual-level and distribution-level---both in a cold-start setting with no human data on the target question. Across thirteen persona constructions, three LLMs, and three datasets, the best calibration method improves on the uncalibrated DTs in every setting. Theoretical analyses uncover transferable structure as a key factor for successful calibration, moving beyond classical DT fidelity measures such as accuracy or correlation. 

\paragraph{Practical guidance.}
The gains vary across methods. Overall, unconstrained linear fit-and-transfer methods perform best, followed by matrix-completion methods \meth{SSV} and \meth{ALS}. Regularized linear maps such as \meth{Ridge}, \meth{Lasso}, and \meth{EN} are therefore strong defaults.

\paragraph{Limitations and future directions.}
The framework currently applies to structured responses only---ratings on a fixed scale, numeric responses, and multiple-choice survey items. Extending calibration to free-form text responses (e.g., open-ended survey answers) is an important direction for future work. Calibration still requires DTs to carry some information about the respondents---transferable structure, if not direct fidelity---and it cannot rescue a simulator whose responses are completely unrelated to the individuals (cf. Appendix~\ref{app:null_twin}). On the theoretical side, although the response model is general, our individual-level error bounds apply only to linear fit-and-transfer methods. Analyses that apply to a broader class of methods could further improve our understanding of the conditions under which calibration is effective.

\paragraph{Broader impact.}
Calibrated digital twins can lower the cost of surveys, help pilot instruments before they are fielded, and reduce sensitive queries. However, the same tools could be misused to fabricate public-opinion data or to pass off simulated responses as genuine human ones. We therefore recommend that calibrated simulations be labeled as model predictions that inherit the biases of the underlying model and the reference data, that the reference data and calibration procedure be disclosed, and that consequential decisions continue to rest on human data.

\section*{Acknowledgments}
We thank Olivier Toubia for helpful comments. Kaizheng Wang's research is supported by the National Science Foundation grant DMS-2544147.

\bibliography{bib}
\bibliographystyle{ims}

\appendix

\section{Additional Details for the Individual-level Problem}\label{app:individual}

This section first expands on what additional assumptions on user representations add to interpretations and guarantees in Theorem~\ref{thm:ridge}, and on candidate subspaces and their defects. It then collects the experimental protocol, prompts, and additional results for the individual-level problem, including the Haar-random baseline for the alignment statistic. Algorithm~\ref{alg:newquestion} describes the complete procedure for obtaining Tables~\ref{tab:movielens_methods}, \ref{tab:2k500_13x10}, \ref{tab:movielens_nrmse}, and \ref{tab:2k500_nrmse}; Algorithms~\ref{alg:newquestion_fittransfer} and~\ref{alg:newquestion_completion} give its subroutines.

\subsection{Population Interpretations of Individual-level Conditions}\label{app:indpopulation}

Theorem~\ref{thm:ridge} is stated conditional on the realized user representations and questions. Consequently, it does not impose any distributional assumptions on the sampled human and DT representations. Here, we briefly discuss how additional assumptions on the user representations can strengthen the interpretation of the structural conditions and support further guarantees.

\paragraph{Population-level approximation defects.}
Fix the $d=m+1$ questions and define the conditional mean response vectors
$$
 g_{HU}(u)
 :=
 \left(\int_{\mathcal A}y\,f_{HU}(dy\mid u,x_j)\right)_{j=1}^d,
 \qquad
 g_{DT}(\widetilde u)
 :=
 \left(\int_{\mathcal A}y\,f_{DT}(dy\mid \widetilde u,x_j)\right)_{j=1}^d.
$$
Thus, row $i$ of $M$ is $g_{HU}(u_i)^\top$, and row $i$ of $\tM$ is $g_{DT}(\widetilde u_i)^\top$. Suppose the pairs $(u_i,\widetilde u_i)$ are sampled i.i.d.\ from a joint distribution with marginals $Q_{HU}$ and $Q_{DT}$. Define the population second-moment matrices
$$
 \Sigma_{HU}
 :=
 \mathbb E_{Q_{HU}}
 [g_{HU}(u)g_{HU}(u)^\top],
 \qquad
 \Sigma_{DT}
 :=
 \mathbb E_{Q_{DT}}
 [g_{DT}(\widetilde u)g_{DT}(\widetilde u)^\top].
$$
Under finite second moments, for fixed $d$, the strong law of large numbers gives
$$
 \frac{M^\top M}{n}\longrightarrow\Sigma_{HU},
 \qquad
 \frac{\tM^\top\tM}{n}\longrightarrow\Sigma_{DT}
$$
almost surely as $n\to\infty$. This connects the individual approximation defects directly to the underlying response model. For example,
$$
 \bigl(\Delta_{\mathrm{HU}}^{\mathrm{ind}}(\mathcal V)\bigr)^2
 =
 \norm{(I_d-P_{\mathcal V})\frac{M^\top M}{n}(I_d-P_{\mathcal V})}_{\op}
 \longrightarrow
 \norm{(I_d-P_{\mathcal V})\Sigma_{HU}(I_d-P_{\mathcal V})}_{\op},
$$
and an analogous relation holds for $\Delta_{\mathrm{DT}}^{\mathrm{ind}}(\mathcal V)$ with $\tM$ and $\Sigma_{DT}$. Thus, the empirical defects measure how well the candidate question space represents the population variation in conditional mean response vectors. 

\paragraph{DT population diversity and stability.}
The i.i.d.\ assumption can also give sufficient conditions for the stability requirement $\gamma_{\mathcal V}>0$. Let $O_{\mathcal V}$ be a matrix whose columns form an orthonormal basis of $\mathcal V$. By definition,
$$
 \gamma_{\mathcal V}^2
 =
 \frac{\lambda_{\min}\left(O_{\mathcal V}^\top(\tM^\top\tM/n)O_{\mathcal V}\right)}
 {\lambda_{\max}(M^\top M/n)}.
$$
Suppose, for a candidate $\mathcal V$, the population second moments satisfy
$$
 \lambda_{\min}\left(O_{\mathcal V}^\top\Sigma_{DT}O_{\mathcal V}\right)\geq c_{\mathcal V}>0,
 \qquad
 0<\lambda_{\max}(\Sigma_{HU})<\infty.
$$
Then, by the preceding law of large numbers and the continuity of extreme eigenvalues, $\gamma_{\mathcal V}^2$ converges almost surely to a limit of at least $c_{\mathcal V}/\lambda_{\max}(\Sigma_{HU})$, so $\gamma_{\mathcal V}$ is bounded away from zero with probability tending to one as $n\to\infty$. This has a natural interpretation: the DT population must exhibit sufficient diversity to reveal the relevant question-space directions.

\subsection{Candidate Subspaces and Defects}\label{app:candidate_subspaces_defects}

\paragraph{Predictability on a candidate subspace.}
Suppose $\kappa_{\mathcal V}>0$ and put $z_{\mathcal V} :=- \tfrac{(I_d-P_{\mathcal V})e_\star} {\kappa_{\mathcal V}^2}$. Since $e_\star^\top(I_d-P_{\mathcal V})e_\star =\kappa_{\mathcal V}^2$, the target coordinate of $z_{\mathcal V}$ is $-1$, and so we can take the first $m$ components of $z_{\mathcal V}$ as the coefficient vector $\beta_{\mathcal V}$ with $\norm{\beta_{\mathcal V}}_2 = \tfrac{\sqrt{1-\kappa_{\mathcal V}^2}}{\kappa_{\mathcal V}}$. If $\kappa_{\mathcal V}=0$, then $e_\star \in \mathcal V$, and there is no coefficient vector $\beta_{\mathcal V}$ such that $v_\star=\beta_{\mathcal V}^\top v_{1:m}$.

\paragraph{Non-emptiness of the admissible family.}
One can always choose $v\notin\operatorname{span}(e_\star)$ with $\tM v\neq0$, for which $\operatorname{span}(v)\in\mathfrak V$. To see why, note that $\tilde r \geq 1$ implies $\tM\neq0$. First suppose $\tM e_\star=0$. Then, any $v$ with $\tM v\neq0$ cannot lie in $\operatorname{span}(e_\star)$. If instead $\tM e_\star\neq0$, choose any nonzero $u\perp e_\star$ and set $v=e_\star+\varepsilon u$. For sufficiently small nonzero $\varepsilon$, continuity gives $\tM v\neq0$, while $v\notin\operatorname{span}(e_\star)$. We additionally note that the stability requirement automatically limits the candidate dimension: $\gamma_{\mathcal V}>0$ implies $\mathcal V\cap\ker(\tM)=\{0\}$, and hence $\dim(\mathcal V)\leq\rank(\tM)=\widetilde r$ for all $\mathcal V\in\mathfrak V$.

\paragraph{Interpretation of individual defects.}
If $\theta_{DT}^{\max}$ is the largest principal angle from $\row(\tM)$ to $\mathcal V$ (counting a missing row-space direction as angle $\pi/2$), then $\frac{\sigma_{\widetilde r}(\tM)}{\sqrt n}\sin\theta_{DT}^{\max} \leq \Delta_{\mathrm{DT}}^{\mathrm{ind}}(\mathcal V) \leq \frac{\sigma_1(\tM)}{\sqrt n}\sin\theta_{DT}^{\max}$. Thus, the DT defect directly reflects principal-angle misalignment, modulated by the DT singular spectrum. The identical sandwich holds for $\Delta_{\mathrm{HU}}^{\mathrm{ind}}$, with $M$, $r$, and the corresponding largest principal angle $\theta_{HU}^{\max}$ in place of $\widetilde M$, $\widetilde r$, and $\theta_{DT}^{\max}$.

\subsection{Implementation Details}\label{app:implementation_details_ind}

\paragraph{Metrics.}
For individual prediction, Pearson correlation is computed within each target and averaged across targets. Normalized root mean squared error (nRMSE) is RMSE divided by the standard deviation of the observed human target vector. Correlation is the primary metric following existing works \citep{park2024generativeagentsimulations1000, toubia2025database, peng2026digitaltwinsfunhousemirrors}.

\paragraph{Per-target cross-validation.}
For each target question, all other questions are references. A fixed 10\% of the references, at least three and at most twelve, is held out for validation. Every method is fitted on the remaining references at each point of its grid and scored by mean correlation on the validation references. The best value is then refitted on every reference to predict the target. The six penalized individual methods use $\lambda\in\{0.01,0.03,0.1,0.3\}$. \meth{HSV} and \meth{SP} use rank in $\{3,5,10,20\}$; \meth{SSV} and \meth{ALS} use shrinkage in $\{1,5,10,20\}$. The penalty $\lambda$ on the grid is on the per-sample scale of the ridge estimator in Section~\ref{sec:exact_transfer}, and each solver receives $\lambda$ in its native convention; for example, the scikit-learn ridge penalty on the unnormalized sum of squares is $\lambda n_{\mathrm{fit}}$, where $n_{\mathrm{fit}}$ is the number of respondents whose DT target is observed. A question is a target only if its human and DT columns both have more than five observed responses and positive variance.

\paragraph{Method-specific details.}
For \meth{Ridge}, \meth{Lasso}, and \meth{EN}, $f(x)=\alpha+x^\top\beta$ is fitted by minimizing squared error plus, respectively, an $\ell_2$, $\ell_1$, or mixed penalty. \meth{EN} fixes the $\ell_1$ ratio at 0.3. \meth{NN} has one hidden layer of eight ReLU units and uses Adam with learning rate $10^{-3}$, batch size 128, at most 200 epochs, patience 20, and seed 42. \meth{SC} minimizes regularized squared error subject to $\beta\geq0$ and $\one^\top\beta=1$. \meth{SI} first projects the DT reference block onto its leading 20 singular directions and fits the linear map there. \meth{HSV} alternates between filling missing entries and truncating the SVD to the selected rank; \meth{SSV} soft-thresholds the singular values and truncates at rank 20; \meth{ALS} alternates regularized least-squares updates of rank-20 row and column factors. \meth{SP} runs hard SVD completion on the human matrix alone, initializing its missing entries from the standardized DT matrix rather than stacking the two systems. All completion methods stop after 100 iterations.

\paragraph{Standardization.}
Each column of the human and DT matrices is standardized separately, using its observed entries only: it is centered and divided by the maximum of its observed standard deviation and one. For a new question, the regression response is the standardized DT target column, and the prediction is returned to the raw scale using the target's DT moments. Prediction and truth differ from their DT-standardized versions by the same affine map, so correlation and nRMSE take the same values on either scale. Unlike correlation, nRMSE therefore includes the DT--human location and dispersion gaps documented in Appendix~\ref{app:twin_moments}.

\paragraph{Missing values.}
For fit-and-transfer methods, missing entries in the reference design are filled by the column mean of observed entries before regression. Completion methods retain missing entries and fill them while holding observed entries fixed.

\paragraph{Intercepts.}
Every regression method except \meth{SC} includes an intercept. For a new question, standardization makes it numerically negligible.

\paragraph{Hyperparameter selection.}
Table~\ref{tab:hp_newquestion} reports the share of folds selecting each grid value. Values are listed in increasing order, and rows may not sum to 100 after rounding.

\begin{table}[htbp]
\TABLE
{New-question hyperparameter selection frequencies.
\label{tab:hp_newquestion}}
{\resizebox{\ifdim\width>\textwidth\textwidth\else\width\fi}{!}{
\begin{tabular}{@{}llccccccccc@{}}
\toprule
& & & \multicolumn{4}{c}{MovieLens} & \multicolumn{4}{c}{Twin-2K-500} \\
\cmidrule(lr){4-7}\cmidrule(lr){8-11}
Method & Parameter & Grid & 1st & 2nd & 3rd & 4th & 1st & 2nd & 3rd & 4th \\
\midrule
\meth{Ridge} & $\lamt$ & \{0.01, 0.03, 0.1, 0.3\} & 0 & 0 & 0 & 100 & 4 & 6 & 24 & 67 \\
\meth{Lasso} & $\lamt$ & \{0.01, 0.03, 0.1, 0.3\} & 1 & 98 & 1 & 0 & 89 & 11 & 0 & 0 \\
\meth{EN} & $\lamt$ & \{0.01, 0.03, 0.1, 0.3\} & 0 & 4 & 94 & 3 & 31 & 58 & 7 & 3 \\
\meth{NN} & $\lamt$ & \{0.01, 0.03, 0.1, 0.3\} & 0 & 13 & 87 & 0 & 8 & 79 & 12 & 0 \\
\meth{SC} & $\lamt$ & \{0.01, 0.03, 0.1, 0.3\} & 14 & 2 & 2 & 81 & 46 & 12 & 20 & 22 \\
\meth{SI} & $\lamt$ & \{0.01, 0.03, 0.1, 0.3\} & 2 & 5 & 21 & 72 & 25 & 10 & 20 & 45 \\
\meth{HSV} & rank & \{3, 5, 10, 20\} & 26 & 57 & 14 & 2 & 4 & 25 & 61 & 10 \\
\meth{SSV} & shrinkage & \{1, 5, 10, 20\} & 0 & 51 & 47 & 1 & 1 & 8 & 27 & 64 \\
\meth{ALS} & shrinkage & \{1, 5, 10, 20\} & 0 & 50 & 48 & 1 & 1 & 8 & 25 & 66 \\
\meth{SP} & rank & \{3, 5, 10, 20\} & 46 & 22 & 32 & 0 & 16 & 46 & 30 & 8 \\
\bottomrule
\end{tabular}
}}
{Percentage of folds selecting each grid value. Twin-2K-500 pools all
thirteen constructions.}
\end{table}

\begin{breakablealgorithm}
\caption{New-question prediction, Tables~\ref{tab:movielens_methods}, \ref{tab:2k500_13x10}, \ref{tab:movielens_nrmse}, and \ref{tab:2k500_nrmse}.}
\label{alg:newquestion}
\small
\begin{algorithmic}[1]
\Require $Y, \tY \in \R^{n \times d}$, both with missing entries; a set
  $\mathcal{J}$ of target questions; a method $f$ with hyperparameter grid
  $\Lambda_f$
\State $\mu_k \gets \operatorname{mean}\{\tY_{ik} : \tY_{ik} \text{ observed}\}$
  and $\sigma_k \gets \max\!\big(\operatorname{sd}\{\tY_{ik} : \tY_{ik} \text{
  observed}\},\, 1\big)$, for every $k \le d$
\State $\mu^{H}_k \gets \operatorname{mean}\{Y_{ik} : Y_{ik} \text{ observed}\}$
  and $\sigma^{H}_k \gets \max\!\big(\operatorname{sd}\{Y_{ik} : Y_{ik} \text{
  observed}\},\, 1\big)$, for every $k \le d$
\For{$j \in \mathcal{J}$}
  \State $\mathcal D \gets \{1,\dots,d\} \setminus \{j\}$
  \State draw $\mathcal D_{\mathrm{val}} \subset \mathcal D$ with $|\mathcal D_{\mathrm{val}}| = \operatorname{clip}\big(
    \operatorname{round}(0.1|\mathcal D|),\, 3,\, 12\big)$, and set $\mathcal D_{\mathrm{fit}} \gets \mathcal D \setminus \mathcal D_{\mathrm{val}}$
  \For{$\lambda \in \Lambda_f$}
    \If{$f$ is a fit-and-transfer method}
      \State $\hat y^{v}(\lambda) \gets \Call{FitTransfer}{v, \mathcal D_{\mathrm{fit}}, f, \lambda}$,
        for each $v \in \mathcal D_{\mathrm{val}}$
    \Else
      \State $\big\{\hat y^{v}(\lambda)\big\}_{v \in \mathcal D_{\mathrm{val}}} \gets
        \Call{Complete}{\mathcal D_{\mathrm{val}} \cup \{j\}, f, \lambda}$ \Comment{prediction for $j$ is not used}
    \EndIf
    \State $\Omega_v \gets \{i : Y_{iv} \text{ observed}\}$ and $\rho_v(\lambda)
      \gets \operatorname{corr}_{i \in \Omega_v}\big(Y_{iv}, \hat
      y^{v}_i(\lambda)\big)$, for each $v \in \mathcal D_{\mathrm{val}}$
  \EndFor
  \State $\hat\lambda \gets \arg\max_{\lambda \in \Lambda_f} |\mathcal D_{\mathrm{val}}|^{-1}
    \sum_{v \in \mathcal D_{\mathrm{val}}} \rho_v(\lambda)$
  \If{$f$ is a fit-and-transfer method}
    \State $\hat y \gets \Call{FitTransfer}{j, \mathcal D, f, \hat\lambda}$
  \Else
    \State $\hat y \gets \Call{Complete}{\{j\}, f, \hat\lambda}$
  \EndIf
  \State $\Omega \gets \{i : Y_{ij} \text{ observed}\}$
  \State $\rho_j \gets \operatorname{corr}_{i \in \Omega}\big(Y_{ij}, \hat y_i\big)$
    and
    $\mathrm{nRMSE}_j \gets \big(|\Omega|^{-1}\sum_{i \in \Omega}(Y_{ij} - \hat
    y_i)^2\big)^{1/2} / \operatorname{sd}_{i \in \Omega}(Y_{ij})$
  \State $\rho^{\mathrm{BL}}_j, \mathrm{nRMSE}^{\mathrm{BL}}_j \gets$ the same
    two with $\hat y_i$ replaced by $\tY_{ij}$
\EndFor
\State $\bar\rho \gets |\mathcal{J}|^{-1} \sum_{j \in \mathcal{J}} \rho_j$ and
  $\se(\bar\rho) \gets |\mathcal{J}|^{-1/2} \operatorname{sd}_{j \in
  \mathcal{J}}(\rho_j)$, and likewise $\overline{\mathrm{nRMSE}}$ and
  $\se(\overline{\mathrm{nRMSE}})$ from $\mathrm{nRMSE}_j$
\State $\bar\rho^{\mathrm{BL}}, \se(\bar\rho^{\mathrm{BL}}),
  \overline{\mathrm{nRMSE}}^{\mathrm{BL}},
  \se(\overline{\mathrm{nRMSE}}^{\mathrm{BL}}) \gets$ the same four from
  $\rho^{\mathrm{BL}}_j$ and $\mathrm{nRMSE}^{\mathrm{BL}}_j$
\State \Return $\bar\rho$, $\se(\bar\rho)$, $\overline{\mathrm{nRMSE}}$,
  $\se(\overline{\mathrm{nRMSE}})$, $\bar\rho^{\mathrm{BL}}$,
  $\se(\bar\rho^{\mathrm{BL}})$, $\overline{\mathrm{nRMSE}}^{\mathrm{BL}}$,
  $\se(\overline{\mathrm{nRMSE}}^{\mathrm{BL}})$
\end{algorithmic}
\end{breakablealgorithm}

\begin{breakablealgorithm}
\caption{New-question fit-and-transfer procedure.}
\label{alg:newquestion_fittransfer}
\small
\begin{algorithmic}[1]
\Procedure{FitTransfer}{target column $j$, reference columns $\mathcal D$, method $f$,
  hyperparameter $\lambda$}
  \State $\tY^{\mathrm{imp}}_{ik} \gets \tY_{ik}$ where observed and $\mu_k$
    where not, for $i \le n$ and $k \in \mathcal D$
  \State $Y^{\mathrm{imp}}_{ik} \gets Y_{ik}$ where observed and $\mu^{H}_k$
    where not, for $i \le n$ and $k \in \mathcal D$
  \State $\tilde X_{ik} \gets (\tY^{\mathrm{imp}}_{ik} - \mu_k)/\sigma_k$ and
    $X^{H}_{ik} \gets (Y^{\mathrm{imp}}_{ik} - \mu^{H}_k)/\sigma^{H}_k$, both $n
    \times |\mathcal D|$
  \State $R \gets \{i : \tY_{ij} \text{ observed}\}$ and $z_i \gets (\tY_{ij}
    - \mu_j)/\sigma_j$
    \Comment{the target column's own twin moments}
  \State $\hat f \gets$ fit $f$, with an intercept except for \meth{SC}, on $\big(\tilde X_{R \cdot}, z_R\big)$ at
    $\lambda$
    \Comment{the map is learned on the twin side}
  \State \Return $\hat f(X^{H}) \, \sigma_j + \mu_j$
    \Comment{applied to the humans, de-standardized on the twin scale}
\EndProcedure
\end{algorithmic}
\end{breakablealgorithm}

\begin{breakablealgorithm}
\caption{New-question completion procedure.}
\label{alg:newquestion_completion}
\small
\begin{algorithmic}[1]
\Procedure{Complete}{hidden columns $H$, method $f$, hyperparameter $\lambda$}
  \State $\tY^{\mathrm{std}}_{ik} \gets (\tY_{ik} - \mu_k)/\sigma_k$ and
    $Y^{\mathrm{std}}_{ik} \gets (Y_{ik} - \mu^{H}_k)/\sigma^{H}_k$, for $i \le n$ and
    $k \le d$
  \State $Y^{\circ}_{ik} \gets Y^{\mathrm{std}}_{ik}$ for $k \notin H$, and
    missing for $k \in H$, for $i \le n$
  \If{$f$ is \meth{SP}}
    \State $Z \gets Y^{\circ} \in \R^{n \times d}$, with missing entries filled by $\tY^{\mathrm{std}}$
  \Else
    \State $Z \gets \big[\,Y^{\circ} ; \tY^{\mathrm{std}}\,\big] \in \R^{2n
      \times d}$ stacked vertically
  \EndIf
  \State $\widehat Z \gets$ the completion of $Z$ at $\lambda$, holding $Z$'s
    observed entries fixed
  \State \Return $\hat y^{k}$ with $\hat y^{k}_i \gets \widehat Z_{ik} \, \sigma_k +
    \mu_k$, for $k \in H$ and $i \le n$
\EndProcedure
\end{algorithmic}
\end{breakablealgorithm}

\subsection{MovieLens Prompts}\label{app:movielen_individual}

Movies are represented by title, genre, and their ten most frequent tags. The prompt templates are displayed below. In the user prompt, \meth{ZS} leaves the in-context field empty, while \meth{IC} fills it with the user's available human ratings on the other 249 evaluation movies.

\begin{tcolorbox}[title=System prompt]
You, AI, are an expert in predicting human movie ratings.\\
You are given a user's history of movie ratings and a new movie for the same user to rate.\\
You are also given a format instruction that specifies the type of rating you need to provide.\\
You need to rate the movie as the user would rate it, based on the user's history of ratings and the format instruction.
\end{tcolorbox}

\begin{tcolorbox}[title=User prompt]
\texttt{PERSONA INFORMATION: \{persona\_ratings\}}\\
\texttt{IN-CONTEXT EXAMPLES: \{in\_context\_ratings\}}\\
\texttt{NEW MOVIE: \{movie\}; GENRE: \{genre\}; TOP 10 TAGS:
\{top\_10\_tags\}.}\\
\texttt{FORMAT INSTRUCTIONS: Only a number on a 5-star rating scale, with half-star increments (0.5--5.0). Larger numbers indicate higher ratings.}
\end{tcolorbox}

\subsection{New-question Correlation on Twin-2K-500}\label{app:2k500_results}

Table~\ref{tab:2k500_13x10} reports the new-question correlation for every persona construction and method summarized in Figure~\ref{fig:2k500_gain_by_method}. We note that the absolute correlation levels reflect the inherent difficulty of extrapolating to entirely unseen questions and the quality of the underlying DT simulation, which depend on many factors outside the scope of calibration, such as the LLM backbone, the persona construction, the prompt design, and the nature of the questions themselves. The contribution of calibration is the \emph{relative} improvement: the best method improves on the uncalibrated DT for every construction, recovering signal that the raw DT responses do not exploit.

\begin{table}[htbp]
\TABLE
{New-question correlation on Twin-2K-500.\label{tab:2k500_13x10}}
{\resizebox{\textwidth}{!}{
\begin{tabular}{@{}lcccccccccccc@{}}
\hline\up
 & & \multicolumn{6}{c}{\textbf{Fit-and-transfer}} & \multicolumn{4}{c}{\textbf{Matrix Completion}} & \\
\textbf{Persona} & \textbf{BL} & \textbf{\meth{Ridge}} & \textbf{\meth{Lasso}} & \textbf{\meth{EN}} & \textbf{\meth{NN}} & \textbf{\meth{SC}} & \textbf{\meth{SI}} & \textbf{\meth{HSV}} & \textbf{\meth{SSV}} & \textbf{\meth{ALS}} & \textbf{\meth{SP}} & \textbf{\%$\Delta$} \\
\hline\up
Text, GPT-4.1-mini & \cell{.203}{.016} & \cellbest{.254}{.022} & \cellbetter{.250}{.023} & \cellbetter{.250}{.023} & \cellbetter{.250}{.022} & \cellbetter{.210}{.015} & \cellbetter{.243}{.018} & \cellbetter{.211}{.019} & \cellbetter{.240}{.020} & \cellbetter{.240}{.019} & \cellworse{.138}{.018} & \gain{+25.5} \\
\hline
Text, Gemini-Flash-2.5 & \cell{.198}{.017} & \cellbetter{.251}{.019} & \cellbest{.261}{.020} & \cellbetter{.251}{.020} & \cellbetter{.236}{.020} & \cellworse{.191}{.013} & \cellbetter{.235}{.019} & \cellworse{.196}{.018} & \cellbetter{.226}{.019} & \cellbetter{.227}{.019} & \cellworse{.138}{.016} & \gain{+32.1} \\
\hline
JSON, GPT-4.1-mini & \cell{.209}{.017} & \cellbetter{.271}{.021} & \cellbest{.274}{.022} & \cellbetter{.267}{.022} & \cellbetter{.264}{.021} & \cellbetter{.220}{.015} & \cellbetter{.263}{.019} & \cellworse{.193}{.020} & \cellbetter{.250}{.020} & \cellbetter{.254}{.020} & \cellworse{.160}{.018} & \gain{+30.8} \\
\hline
JSON, GPT-4.1 & \cell{.243}{.019} & \cellbest{.276}{.019} & \cellbetter{.266}{.019} & \cellbetter{.268}{.020} & \cellbetter{.267}{.019} & \cellworse{.223}{.014} & \cellbetter{.269}{.018} & \cellworse{.223}{.019} & \cellbetter{.263}{.018} & \cellbetter{.261}{.018} & \cellworse{.138}{.018} & \gain{+13.5} \\
\hline
Text + CoT, GPT-4.1-mini & \cell{.193}{.016} & \cellbetter{.282}{.022} & \cellbetter{.292}{.023} & \cellbest{.296}{.023} & \cellbetter{.275}{.022} & \cellbetter{.210}{.015} & \cellbetter{.268}{.021} & \cellbetter{.222}{.021} & \cellbetter{.261}{.021} & \cellbetter{.259}{.021} & \cellworse{.141}{.017} & \gain{+53.3} \\
\hline
Text + repeat, GPT-4.1-mini & \cell{.204}{.016} & \cellbetter{.276}{.022} & \cellbest{.286}{.023} & \cellbetter{.285}{.023} & \cellbetter{.278}{.022} & \cellbetter{.211}{.014} & \cellbetter{.264}{.019} & \cellbetter{.233}{.020} & \cellbetter{.262}{.021} & \cellbetter{.258}{.021} & \cellworse{.150}{.017} & \gain{+39.9} \\
\hline
Text, temp=0.7, GPT-4.1-mini & \cell{.187}{.016} & \cellbest{.248}{.022} & \cellbetter{.246}{.023} & \cellbetter{.243}{.023} & \cellbetter{.243}{.022} & \cellbetter{.202}{.015} & \cellbetter{.239}{.018} & \cellbetter{.216}{.020} & \cellbetter{.241}{.019} & \cellbetter{.238}{.020} & \cellworse{.131}{.018} & \gain{+32.2} \\
\hline
JSON + PO, GPT-4.1-mini & \cell{.192}{.015} & \cellbest{.275}{.021} & \cellbetter{.268}{.022} & \cellbetter{.270}{.022} & \cellbetter{.267}{.021} & \cellbetter{.228}{.015} & \cellbetter{.246}{.018} & \cellbetter{.199}{.020} & \cellbetter{.253}{.020} & \cellbetter{.252}{.020} & \cellworse{.127}{.018} & \gain{+43.3} \\
\hline
JSON + PO, GPT-4.1 & \cell{.241}{.018} & \cellbetter{.265}{.021} & \cellbetter{.267}{.021} & \cellbest{.268}{.021} & \cellbetter{.264}{.020} & \cellworse{.212}{.013} & \cellbetter{.256}{.021} & \cellworse{.175}{.022} & \cellbetter{.257}{.020} & \cellbetter{.255}{.020} & \cellworse{.173}{.016} & \gain{+11.2} \\
\hline
Fine-tuned GPT-4.1-mini & \cell{.058}{.009} & \cellbest{.299}{.019} & \cellbetter{.260}{.019} & \cellbetter{.288}{.019} & \cellbetter{.288}{.019} & \cellbetter{.209}{.015} & \cellbetter{.288}{.019} & \cellbetter{.269}{.017} & \cellbetter{.287}{.017} & \cellbetter{.284}{.018} & \cellbetter{.105}{.020} & \gain{+417} \\
\hline
Demographics only, GPT-4.1-mini & \cell{.141}{.012} & \cellbest{.248}{.021} & \cellbetter{.247}{.021} & \cellbetter{.248}{.022} & \cellbetter{.239}{.021} & \cellbetter{.216}{.015} & \cellbetter{.239}{.019} & \cellbetter{.219}{.020} & \cellbetter{.246}{.019} & \cellbetter{.247}{.019} & \cellworse{.117}{.018} & \gain{+75.9} \\
\hline
Summary, GPT-4.1-mini & \cell{.175}{.014} & \cellbest{.285}{.020} & \cellbetter{.276}{.021} & \cellbetter{.279}{.021} & \cellbetter{.279}{.020} & \cellbetter{.228}{.015} & \cellbetter{.267}{.018} & \cellbetter{.241}{.019} & \cellbetter{.274}{.018} & \cellbetter{.275}{.018} & \cellworse{.117}{.019} & \gain{+62.7} \\
\hline
Summary + JSON, GPT-4.1-mini & \cell{.120}{.012} & \cellbest{.283}{.020} & \cellbetter{.272}{.020} & \cellbetter{.274}{.020} & \cellbetter{.267}{.020} & \cellbetter{.230}{.014} & \cellbetter{.266}{.018} & \cellbetter{.242}{.019} & \cellbetter{.273}{.018} & \cellbetter{.269}{.018} & \cellbetter{.142}{.018} & \gain{+136} \down\\
\hline
\end{tabular}
}}
{Pearson correlation for thirteen persona constructions and ten calibration methods; higher is better. Mean above, standard error across target questions in parentheses below. BL is the uncalibrated DT. Green bold marks the best method in each row, blue a method that beats BL, and red one that does not. \%$\Delta$ is the best method's change relative to BL.}
\end{table}

\subsection{New-question nRMSE}\label{app:nrmse}
Tables~\ref{tab:movielens_nrmse} and~\ref{tab:2k500_nrmse} complement the main text's correlation results. The best method improves nRMSE on MovieLens and for all thirteen Twin-2K-500 constructions. Hyperparameters are selected by validation correlation, so nRMSE is a secondary metric. Because predictions are returned on the target's DT scale, nRMSE values above one reflect the DT--human location and dispersion gaps of Appendix~\ref{app:twin_moments}. 

\begin{table}[htbp]
\TABLE
{MovieLens new-question nRMSE.\label{tab:movielens_nrmse}}
{
\setlength{\tabcolsep}{3pt}\begin{tabular}{@{}lcccccccccccc@{}}
\hline\up
& \multicolumn{2}{c}{\textbf{Baselines}} & \multicolumn{6}{c}{\textbf{Fit-and-transfer}} & \multicolumn{4}{c}{\textbf{Matrix Completion}} \\
& \meth{ZS} & \meth{IC} & \meth{Ridge} & \meth{Lasso} & \meth{EN} & \meth{NN} & \meth{SC} & \meth{SI} & \meth{HSV} & \meth{SSV} & \meth{ALS} & \meth{SP} \\ \hline\up
nRMSE & 1.092 & 1.032 & \better{.959} & \better{.970} & \better{.961} & \better{.965} & \better{.992} & \better{.962} & \better{1.043} & \better{.955} & \best{.955} & \better{.982} \\
S.E. & .008 & .006 & \better{.008} & \better{.008} & \better{.008} & \better{.007} & \better{.007} & \better{.007} & \better{.019} & \better{.007} & \best{.007} & \better{.007} \\
\%$\Delta$ \meth{ZS} & --- & -5 & \better{-12} & \better{-11} & \better{-12} & \better{-12} & \better{-9} & \better{-12} & \better{-4} & \better{-12} & \best{-13} & \better{-10} \\
\%$\Delta$ \meth{IC} & --- & --- & \better{-7} & \better{-6} & \better{-7} & \better{-6} & \better{-4} & \better{-7} & \worse{+1} & \better{-7} & \best{-7} & \better{-5} \down\\
\hline
\end{tabular}
}
{Lower is better. \meth{ZS} is the uncalibrated DT and \meth{IC} the in-context prompt. Colors as in Table~\ref{tab:movielens_methods}.
Standard errors are across 250 target movies.}
\end{table}

\begin{table}[htbp]
\TABLE
{Twin-2K-500 new-question nRMSE.\label{tab:2k500_nrmse}}
{\resizebox{0.97\textwidth}{!}{
\begin{tabular}{@{}lcccccccccccc@{}}
\hline\up
 & & \multicolumn{6}{c}{\textbf{Fit-and-transfer}} & \multicolumn{4}{c}{\textbf{Matrix Completion}} & \\
\textbf{Persona} & \textbf{BL} & \textbf{\meth{Ridge}} & \textbf{\meth{Lasso}} & \textbf{\meth{EN}} & \textbf{\meth{NN}} & \textbf{\meth{SC}} & \textbf{\meth{SI}} & \textbf{\meth{HSV}} & \textbf{\meth{SSV}} & \textbf{\meth{ALS}} & \textbf{\meth{SP}} & \textbf{\%$\Delta$} \\
\hline\up
Text, GPT-4.1-mini & \cell{1.206}{.016} & \cellbetter{1.054}{.015} & \cellbest{1.053}{.016} & \cellbetter{1.055}{.016} & \cellbetter{1.055}{.016} & \cellbetter{1.075}{.014} & \cellbetter{1.061}{.015} & \cellworse{1.240}{.051} & \cellbetter{1.063}{.015} & \cellbetter{1.062}{.015} & \cellbetter{1.090}{.015} & \gain{-12.7} \\
\hline
Text, Gemini-Flash-2.5 & \cell{1.258}{.021} & \cellbetter{1.087}{.020} & \cellbetter{1.086}{.020} & \cellbetter{1.089}{.020} & \cellbetter{1.100}{.021} & \cellbetter{1.107}{.018} & \cellbest{1.085}{.020} & \cellworse{1.280}{.045} & \cellbetter{1.107}{.021} & \cellbetter{1.106}{.023} & \cellbetter{1.110}{.018} & \gain{-13.8} \\
\hline
JSON, GPT-4.1-mini & \cell{1.188}{.015} & \cellbetter{1.048}{.015} & \cellbetter{1.047}{.016} & \cellbest{1.044}{.016} & \cellbetter{1.053}{.015} & \cellbetter{1.064}{.012} & \cellbetter{1.045}{.014} & \cellworse{1.362}{.072} & \cellbetter{1.048}{.013} & \cellbetter{1.047}{.013} & \cellbetter{1.067}{.013} & \gain{-12.1} \\
\hline
JSON, GPT-4.1 & \cell{1.168}{.019} & \cellbest{1.031}{.019} & \cellbetter{1.043}{.020} & \cellbetter{1.034}{.019} & \cellbetter{1.039}{.019} & \cellbetter{1.056}{.017} & \cellbetter{1.033}{.018} & \cellbetter{1.154}{.040} & \cellbetter{1.039}{.018} & \cellbetter{1.044}{.019} & \cellbetter{1.063}{.018} & \gain{-11.7} \\
\hline
Text + CoT, GPT-4.1-mini & \cell{1.223}{.015} & \cellbetter{1.058}{.017} & \cellbetter{1.054}{.018} & \cellbest{1.053}{.018} & \cellbetter{1.058}{.017} & \cellbetter{1.090}{.015} & \cellbetter{1.063}{.017} & \cellbetter{1.217}{.042} & \cellbetter{1.070}{.016} & \cellbetter{1.072}{.016} & \cellbetter{1.091}{.016} & \gain{-14.0} \\
\hline
Text + repeat, GPT-4.1-mini & \cell{1.201}{.017} & \cellbetter{1.047}{.018} & \cellbest{1.040}{.019} & \cellbetter{1.040}{.019} & \cellbetter{1.048}{.019} & \cellbetter{1.078}{.017} & \cellbetter{1.054}{.018} & \cellbetter{1.199}{.043} & \cellbetter{1.059}{.018} & \cellbetter{1.061}{.018} & \cellbetter{1.082}{.017} & \gain{-13.4} \\
\hline
Text, temp=0.7, GPT-4.1-mini & \cell{1.225}{.015} & \cellbetter{1.057}{.016} & \cellbest{1.054}{.016} & \cellbetter{1.057}{.016} & \cellbetter{1.059}{.016} & \cellbetter{1.080}{.014} & \cellbetter{1.064}{.015} & \cellbetter{1.163}{.035} & \cellbetter{1.069}{.016} & \cellbetter{1.069}{.016} & \cellbetter{1.094}{.015} & \gain{-14.0} \\
\hline
JSON + PO, GPT-4.1-mini & \cell{1.237}{.017} & \cellbetter{1.120}{.018} & \cellbest{1.116}{.019} & \cellbetter{1.116}{.019} & \cellbetter{1.121}{.018} & \cellbetter{1.142}{.017} & \cellbetter{1.126}{.018} & \cellworse{1.410}{.062} & \cellbetter{1.132}{.018} & \cellbetter{1.134}{.018} & \cellbetter{1.147}{.017} & \gain{-9.8} \\
\hline
JSON + PO, GPT-4.1 & \cell{1.128}{.015} & \cellbetter{1.025}{.017} & \cellbetter{1.025}{.018} & \cellbetter{1.022}{.018} & \cellbetter{1.023}{.017} & \cellbetter{1.050}{.015} & \cellbest{1.022}{.017} & \cellworse{1.627}{.100} & \cellbetter{1.029}{.017} & \cellbetter{1.033}{.017} & \cellbetter{1.045}{.016} & \gain{-9.4} \\
\hline
Fine-tuned GPT-4.1-mini & \cell{1.378}{.023} & \cellbetter{1.102}{.025} & \cellbetter{1.131}{.026} & \cellbetter{1.105}{.025} & \cellbetter{1.108}{.025} & \cellbetter{1.123}{.023} & \cellbest{1.102}{.024} & \cellbetter{1.166}{.036} & \cellbetter{1.105}{.024} & \cellbetter{1.106}{.024} & \cellbetter{1.131}{.023} & \gain{-20.0} \\
\hline
Demographics only, GPT-4.1-mini & \cell{1.320}{.020} & \cellbetter{1.185}{.022} & \cellbetter{1.186}{.023} & \cellbest{1.184}{.023} & \cellbetter{1.192}{.022} & \cellbetter{1.202}{.022} & \cellbetter{1.189}{.022} & \cellbetter{1.319}{.039} & \cellbetter{1.193}{.022} & \cellbetter{1.191}{.022} & \cellbetter{1.213}{.022} & \gain{-10.3} \\
\hline
Summary, GPT-4.1-mini & \cell{1.331}{.024} & \cellbest{1.151}{.025} & \cellbetter{1.163}{.025} & \cellbetter{1.156}{.025} & \cellbetter{1.154}{.025} & \cellbetter{1.171}{.025} & \cellbetter{1.157}{.025} & \cellbetter{1.297}{.049} & \cellbetter{1.156}{.025} & \cellbetter{1.156}{.025} & \cellbetter{1.191}{.025} & \gain{-13.5} \\
\hline
Summary + JSON, GPT-4.1-mini & \cell{1.392}{.026} & \cellbest{1.197}{.026} & \cellbetter{1.204}{.026} & \cellbetter{1.199}{.027} & \cellbetter{1.198}{.026} & \cellbetter{1.218}{.026} & \cellbetter{1.202}{.026} & \cellbetter{1.324}{.045} & \cellbetter{1.205}{.026} & \cellbetter{1.204}{.026} & \cellbetter{1.234}{.026} & \gain{-14.0} \down\\
\hline
\end{tabular}
}}
{Lower is better. Mean above, standard error across target questions in
parentheses below. BL is the uncalibrated DT. Colors as in Table~\ref{tab:2k500_13x10}.}
\end{table}

\subsection{Haar-random Baseline}\label{app:haar-random}
The following proposition shows that a uniformly random $r_0$-dimensional subspace has expected alignment $r_0 / d$ with a fixed $r_0$-dimensional subspace in a $d$-dimensional vector space. This serves as our baseline for subspace alignment in Sections~\ref{sec:alignment}, \ref{sec:twin2k}, and Appendix~\ref{app:newuser}; the proof is in Appendix~\ref{app:haarproof}.

\begin{proposition}[Haar-random baseline]\label{prop:haar}
Let $\mathcal S_1\subseteq\R^d$ be a fixed $r_0$-dimensional subspace, and let $\mathcal S_2$ be Haar-uniform on the Grassmannian of $r_0$-dimensional subspaces of $\R^d$. Then, the expected alignment between $\mathcal S_1$ and $\mathcal S_2$ is $\mathbb E\!\left[A(r_0)\right]=r_0 / d$, where $A(r_0)$ is defined in \eqref{eq:alignment}.
\end{proposition}

\subsection{Alignment Tables}
\label{app:alignment}

Table~\ref{tab:alignment_question} reports the alignment statistic, its range over Twin-2K-500 constructions, the exact Haar expectation, a Monte Carlo 90\% band, and the ratio to the expectation. The main-text figures plot the final column. Before computing singular subspaces, each human and DT question is centered and scaled separately using its observed entries, with the standard deviation floored at one; remaining missing entries are filled by the corresponding column mean.

\begin{table}[htbp]
\TABLE
{Question-space alignment.\label{tab:alignment_question}}
{\begin{tabular}{llccccc}
\toprule
Dataset & $r_0$ & Statistic & Range over constructions & Analytic baseline
& MC baseline band & Ratio \\
\midrule
Twin-2K-500 & 1 & $0.804$ & $0.486$--$0.891$ & $0.008$ & $[0.000,\,0.030]$ & $98.9$ \\
 & 2 & $0.550$ & $0.335$--$0.724$ & $0.016$ & $[0.003,\,0.038]$ & $33.8$ \\
 & 3 & $0.497$ & $0.269$--$0.671$ & $0.024$ & $[0.010,\,0.047]$ & $20.4$ \\
 & 5 & $0.484$ & $0.416$--$0.554$ & $0.041$ & $[0.025,\,0.059]$ & $11.9$ \\
 & 8 & $0.443$ & $0.346$--$0.579$ & $0.065$ & $[0.051,\,0.081]$ & $6.8$ \\
 & 11 & $0.411$ & $0.324$--$0.536$ & $0.089$ & $[0.073,\,0.104]$ & $4.6$ \\
 & 15 & $0.408$ & $0.324$--$0.480$ & $0.122$ & $[0.105,\,0.138]$ & $3.3$ \\
 & 20 & $0.418$ & $0.322$--$0.506$ & $0.163$ & $[0.145,\,0.178]$ & $2.6$ \\
 & 25 & $0.446$ & $0.344$--$0.546$ & $0.203$ & $[0.187,\,0.219]$ & $2.2$ \\
 & 30 & $0.463$ & $0.351$--$0.569$ & $0.244$ & $[0.230,\,0.257]$ & $1.9$ \\
 & 40 & $0.466$ & $0.341$--$0.589$ & $0.325$ & $[0.312,\,0.338]$ & $1.4$ \\
 & 50 & $0.491$ & $0.349$--$0.624$ & $0.407$ & $[0.397,\,0.420]$ & $1.2$ \\
\midrule
MovieLens & 1 & $0.940$ & --- & $0.004$ & $[0.000,\,0.019]$ & $235.0$ \\
 & 2 & $0.598$ & --- & $0.008$ & $[0.002,\,0.016]$ & $74.7$ \\
 & 3 & $0.494$ & --- & $0.012$ & $[0.004,\,0.022]$ & $41.2$ \\
 & 5 & $0.477$ & --- & $0.020$ & $[0.012,\,0.030]$ & $23.8$ \\
 & 8 & $0.488$ & --- & $0.032$ & $[0.023,\,0.042]$ & $15.3$ \\
 & 11 & $0.404$ & --- & $0.044$ & $[0.036,\,0.053]$ & $9.2$ \\
 & 15 & $0.357$ & --- & $0.060$ & $[0.052,\,0.068]$ & $5.9$ \\
 & 20 & $0.337$ & --- & $0.080$ & $[0.071,\,0.088]$ & $4.2$ \\
 & 25 & $0.324$ & --- & $0.100$ & $[0.092,\,0.108]$ & $3.2$ \\
 & 30 & $0.331$ & --- & $0.120$ & $[0.112,\,0.129]$ & $2.8$ \\
 & 40 & $0.342$ & --- & $0.160$ & $[0.152,\,0.168]$ & $2.1$ \\
 & 50 & $0.352$ & --- & $0.200$ & $[0.193,\,0.207]$ & $1.8$ \\
\bottomrule
\end{tabular}}
{Mean squared principal cosine between the human and digital-twin question subspaces, against the Haar-random baseline. ``Statistic'' averages over the 13 Twin-2K-500 constructions and ``Range'' gives their minimum and maximum; MovieLens is a single matrix pair, so its range is degenerate. The analytic baseline is $r_0/d$, with $d = 123$ for Twin-2K-500 and $d = 250$ for MovieLens. The ``MC baseline band'' is the $5$th--$95$th percentile band over $200$ independent pairs of Haar-uniform subspaces, whose mean agrees with the analytic baseline to three decimals at every $r_0$. The last column is the statistic divided by the analytic baseline.}
\end{table}

\section{New-user Prediction}\label{app:newuser}

This section collects the new-user experiment summarized in Remark~\ref{rem:indnewuser}. Unless stated otherwise, it follows the new-question protocol of Appendix~\ref{app:individual} with the roles of questions and users exchanged. In particular, for each target user, all other users are references, and the validation split and hyperparameter grids of Appendix~\ref{app:implementation_details_ind} apply to reference users.

\subsection{Design and Preprocessing}\label{app:newuser_design}

Twin-2K-500 uses six constructions: Text GPT-4.1-mini, Text Gemini-Flash-2.5, JSON GPT-4.1, Text + CoT, fine-tuned GPT-4.1-mini, and demographics only. For each, 200 target users are sampled once to save cost, and for each target user all other users are references. MovieLens uses all 500 users as targets in turn, with 499 references for each target user. Questions are standardized as in Appendix~\ref{app:implementation_details_ind}, except that the moments are computed from the reference users and both prediction and truth are evaluated on the per-question standardized scale; raw new-user metrics would aggregate questions measured in incomparable units. Unlike in the new-question setting, since the standardization is per-question, the intercept in new-user settings remains a genuine fitted parameter. Algorithms~\ref{alg:newuser}, \ref{alg:newuser_fittransfer}, and~\ref{alg:newuser_completion} give the complete procedures.

\begin{breakablealgorithm}
\caption{New-user prediction,
Tables~\ref{tab:newuser_corr} and~\ref{tab:newuser_nrmse}.}
\label{alg:newuser}
\small
\begin{algorithmic}[1]
\Require $Y, \tY \in \R^{n \times d}$, both with missing entries; a set
  $\mathcal{I}$ of target users; a method $f$ with hyperparameter grid
  $\Lambda_f$
\For{$i \in \mathcal{I}$}
  \State $\mathcal D \gets \{1,\dots,n\} \setminus \{i\}$
  \State draw $\mathcal D_{\mathrm{val}} \subset \mathcal D$ with $|\mathcal D_{\mathrm{val}}| = \operatorname{clip}\big(
    \operatorname{round}(0.1|\mathcal D|),\, 3,\, 12\big)$, and set $\mathcal D_{\mathrm{fit}} \gets \mathcal D \setminus \mathcal D_{\mathrm{val}}$
  \For{$\lambda \in \Lambda_f$}
    \If{$f$ is a fit-and-transfer method}
      \State $\hat y^{v}(\lambda) \gets \Call{FitTransfer}{v, \mathcal D_{\mathrm{fit}}, f, \lambda}$,
        for each $v \in \mathcal D_{\mathrm{val}}$
    \Else
      \State $\big\{\hat y^{v}(\lambda)\big\}_{v \in \mathcal D_{\mathrm{val}}} \gets
        \Call{Complete}{\mathcal D_{\mathrm{val}} \cup \{i\}, f, \lambda}$ \Comment{prediction for $i$ is not used}
    \EndIf
    \State $u^{v}_k \gets \big(Y_{vk} - \mu^{H}_k(\mathcal D_{\mathrm{fit}})\big)/\sigma^{H}_k(\mathcal D_{\mathrm{fit}})$ and
      $\hat u^{v}_k \gets \big(\hat y^{v}_k(\lambda) -
      \mu^{H}_k(\mathcal D_{\mathrm{fit}})\big)/\sigma^{H}_k(\mathcal D_{\mathrm{fit}})$, for $k \le d$ and each $v \in \mathcal D_{\mathrm{val}}$
    \State $\Omega_v \gets \{k : Y_{vk} \text{ observed}\}$ and $\rho_v(\lambda)
      \gets \operatorname{corr}_{k \in \Omega_v}\big(u^{v}_k, \hat u^{v}_k\big)$,
      for each $v \in \mathcal D_{\mathrm{val}}$
  \EndFor
  \State $\hat\lambda \gets \arg\max_{\lambda \in \Lambda_f} |\mathcal D_{\mathrm{val}}|^{-1}
    \sum_{v \in \mathcal D_{\mathrm{val}}} \rho_v(\lambda)$
  \If{$f$ is a fit-and-transfer method}
    \State $\hat y \gets \Call{FitTransfer}{i, \mathcal D, f, \hat\lambda}$
  \Else
    \State $\hat y \gets \Call{Complete}{\{i\}, f, \hat\lambda}$
  \EndIf
  \State $u_k \gets \big(Y_{ik} - \mu^{H}_k(\mathcal D)\big)/\sigma^{H}_k(\mathcal D)$ and $\hat
    u_k \gets \big(\hat y_k - \mu^{H}_k(\mathcal D)\big)/\sigma^{H}_k(\mathcal D)$, for $k \le d$,
    and $\Omega \gets \{k : Y_{ik} \text{ observed}\}$
  \State $\rho_i \gets \operatorname{corr}_{k \in \Omega}(u_k, \hat u_k)$
    and
    $\mathrm{nRMSE}_i \gets \big(|\Omega|^{-1}\sum_{k \in \Omega}(u_k - \hat
    u_k)^2\big)^{1/2} / \operatorname{sd}_{k \in \Omega}(u_k)$
  \State $\rho^{\mathrm{BL}}_i, \mathrm{nRMSE}^{\mathrm{BL}}_i \gets$ the same
    two with $\hat u_k$ replaced by $\big(\tY_{ik} - \mu_k(\mathcal D)\big)/\sigma_k(\mathcal D)$
\EndFor
\State $\bar\rho \gets |\mathcal{I}|^{-1} \sum_{i \in \mathcal{I}} \rho_i$ and
  $\se(\bar\rho) \gets |\mathcal{I}|^{-1/2} \operatorname{sd}_{i \in
  \mathcal{I}}(\rho_i)$, and likewise $\overline{\mathrm{nRMSE}}$ and
  $\se(\overline{\mathrm{nRMSE}})$ from $\mathrm{nRMSE}_i$
\State $\bar\rho^{\mathrm{BL}}, \se(\bar\rho^{\mathrm{BL}}),
  \overline{\mathrm{nRMSE}}^{\mathrm{BL}},
  \se(\overline{\mathrm{nRMSE}}^{\mathrm{BL}}) \gets$ the same four from
  $\rho^{\mathrm{BL}}_i$ and $\mathrm{nRMSE}^{\mathrm{BL}}_i$
\State \Return $\bar\rho$, $\se(\bar\rho)$, $\overline{\mathrm{nRMSE}}$,
  $\se(\overline{\mathrm{nRMSE}})$, $\bar\rho^{\mathrm{BL}}$,
  $\se(\bar\rho^{\mathrm{BL}})$, $\overline{\mathrm{nRMSE}}^{\mathrm{BL}}$,
  $\se(\overline{\mathrm{nRMSE}}^{\mathrm{BL}})$
\end{algorithmic}
\end{breakablealgorithm}

\begin{breakablealgorithm}
\caption{New-user fit-and-transfer procedure.}
\label{alg:newuser_fittransfer}
\small
\begin{algorithmic}[1]
\Procedure{FitTransfer}{target user $i$, reference users $\mathcal D$, method $f$,
  hyperparameter $\lambda$}
  \State $\tY^{\mathrm{imp}}_{uk} \gets \tY_{uk}$ where observed and $\mu_k(\mathcal D)$
    where not, for $u \in \mathcal D$ and $k \le d$
  \State $Y^{\mathrm{imp}}_{uk} \gets Y_{uk}$ where observed and $\mu^{H}_k(\mathcal D)$
    where not, for $u \in \mathcal D$ and $k \le d$
  \State $\tilde X_{ku} \gets \big(\tY^{\mathrm{imp}}_{uk} -
    \mu_k(\mathcal D)\big)/\sigma_k(\mathcal D)$ and $X^{H}_{ku} \gets \big(Y^{\mathrm{imp}}_{uk}
    - \mu^{H}_k(\mathcal D)\big)/\sigma^{H}_k(\mathcal D)$, both $d \times |\mathcal D|$
  \State $R \gets \{k : \tY_{ik} \text{ observed}\}$ and $z_k \gets
    \big(\tY_{ik} - \mu_k(\mathcal D)\big)/\sigma_k(\mathcal D)$
    \Comment{the target's own twin row}
  \State $\hat f \gets$ fit $f$, with an intercept except for \meth{SC}, on $\big(\tilde X_{R \cdot}, z_R\big)$ at
    $\lambda$
    \Comment{weights over reference \emph{users}}
  \State \Return $\hat y$ with $\hat y_k \gets \hat f(X^{H})_k \,
    \sigma^{H}_k(\mathcal D) + \mu^{H}_k(\mathcal D)$
\EndProcedure
\end{algorithmic}
\end{breakablealgorithm}

\begin{breakablealgorithm}
\caption{New-user completion procedure.}
\label{alg:newuser_completion}
\small
\begin{algorithmic}[1]
\Procedure{Complete}{hidden users $H$, method $f$, hyperparameter $\lambda$}
  \State $G \gets \{1,\dots,n\} \setminus H$
    \Comment{$\mathcal D_{\mathrm{fit}}$ under cross-validation, $\mathcal D$ at the
      refit}
  \State $\tY^{\mathrm{std}}_{uk} \gets \big(\tY_{uk} -
    \mu_k(G)\big)/\sigma_k(G)$ and $Y^{\mathrm{std}}_{uk} \gets \big(Y_{uk} -
    \mu^{H}_k(G)\big)/\sigma^{H}_k(G)$, for $u \le n$ and $k \le d$
  \State $Y^{\circ}_{uk} \gets Y^{\mathrm{std}}_{uk}$ for $u \notin H$, and
    missing for $u \in H$, for $k \le d$
  \If{$f$ is \meth{SP}}
    \State $Z \gets Y^{\circ} \in \R^{n \times d}$, with missing entries filled by $\tY^{\mathrm{std}}$
  \Else
    \State $Z \gets \big[\,Y^{\circ} \mid \tY^{\mathrm{std}}\,\big] \in \R^{n
      \times 2d}$ stacked side by side
  \EndIf
  \State $\widehat Z \gets$ the completion of $Z$ at $\lambda$, holding $Z$'s
    observed entries fixed
  \State \Return $\hat y^{u}$ with $\hat y^{u}_k \gets \widehat Z_{uk} \,
    \sigma^{H}_k(G) + \mu^{H}_k(G)$, for $u \in H$ and $k \le d$
\EndProcedure
\end{algorithmic}
\end{breakablealgorithm}

\subsection{Results}\label{app:newuser_results}

Table~\ref{tab:newuser_corr} reports the new-user correlation summarized in Remark~\ref{rem:indnewuser}. The best method improves on the uncalibrated correlation for every construction. Table~\ref{tab:newuser_nrmse} reports nRMSE, and Table~\ref{tab:hp_newuser} reports hyperparameter selection frequencies.

\begin{table}[htbp]
\TABLE
{New-user correlation.\label{tab:newuser_corr}}
{\resizebox{\textwidth}{!}{
\begin{tabular}{@{}lccccccccccccc@{}}
\hline\up
 & & \multicolumn{6}{c}{\textbf{Fit-and-transfer}} & \multicolumn{4}{c}{\textbf{Matrix Completion}} & & \\
\textbf{Persona} & \textbf{BL} & \textbf{\meth{Ridge}} & \textbf{\meth{Lasso}} & \textbf{\meth{EN}} & \textbf{\meth{NN}} & \textbf{\meth{SC}} & \textbf{\meth{SI}} & \textbf{\meth{HSV}} & \textbf{\meth{SSV}} & \textbf{\meth{ALS}} & \textbf{\meth{SP}} & \textbf{\%$\Delta$} & \textbf{Folds} \\
\hline\up
Text, GPT-4.1-mini & \cell{.181}{.013} & \cellbetter{.231}{.012} & \cellworse{.163}{.012} & \cellworse{.178}{.013} & \cellbest{.238}{.012} & \cellbetter{.223}{.015} & \cellbetter{.228}{.013} & \cellworse{.095}{.016} & \cellbetter{.213}{.014} & \cellbetter{.212}{.014} & \cellworse{.152}{.018} & \gain{+31.5} & 200 \\
\hline
Text, Gemini-Flash-2.5 & \cell{.205}{.012} & \cellbest{.265}{.012} & \cellworse{.164}{.012} & \cellworse{.180}{.013} & \cellbetter{.263}{.013} & \cellbetter{.226}{.015} & \cellbetter{.230}{.014} & \cellworse{.136}{.016} & \cellbetter{.206}{.013} & \cellbetter{.206}{.013} & \cellworse{.194}{.016} & \gain{+28.9} & 200 \\
\hline
JSON, GPT-4.1 & \cell{.232}{.012} & \cellbest{.274}{.012} & \cellworse{.156}{.012} & \cellworse{.187}{.013} & \cellbetter{.273}{.013} & \cellworse{.222}{.015} & \cellworse{.231}{.014} & \cellworse{.106}{.016} & \cellworse{.214}{.014} & \cellworse{.207}{.015} & \cellworse{.154}{.018} & \gain{+18.4} & 200 \\
\hline
Text + CoT, GPT-4.1-mini & \cell{.184}{.012} & \cellbetter{.228}{.013} & \cellworse{.147}{.012} & \cellworse{.164}{.013} & \cellbest{.230}{.013} & \cellbetter{.218}{.015} & \cellbetter{.201}{.015} & \cellworse{.061}{.016} & \cellbetter{.190}{.015} & \cellbetter{.185}{.015} & \cellworse{.154}{.018} & \gain{+24.9} & 200 \\
\hline
Fine-tuned GPT-4.1-mini & \cell{.033}{.014} & \cellbetter{.077}{.014} & \cellworse{.028}{.013} & \cellbetter{.048}{.013} & \cellbetter{.065}{.016} & \cellbest{.091}{.018} & \cellbetter{.077}{.016} & \cellbetter{.054}{.017} & \cellbetter{.069}{.015} & \cellbetter{.077}{.016} & \cellworse{.025}{.016} & \gain{+176} & 200 \\
\hline
Demographics only, GPT-4.1-mini & \cell{.115}{.012} & \cellbest{.188}{.014} & \cellworse{.113}{.013} & \cellbetter{.142}{.014} & \cellbetter{.182}{.014} & \cellbetter{.186}{.016} & \cellbetter{.173}{.015} & \cellworse{.067}{.016} & \cellbetter{.153}{.015} & \cellbetter{.152}{.015} & \cellworse{.108}{.018} & \gain{+63.2} & 200 \\
\hline
MovieLens & \cell{.026}{.005} & \cellbetter{.064}{.006} & \cellbetter{.038}{.006} & \cellbetter{.036}{.006} & \cellbetter{.073}{.006} & \cellbetter{.048}{.006} & \cellbetter{.070}{.007} & \cellbetter{.077}{.009} & \cellbetter{.080}{.007} & \cellbest{.081}{.007} & \cellbetter{.058}{.009} & \gain{+208} & 500 \down\\
\hline
\end{tabular}
}}
{Computed on the per-question standardized scale. Higher is better. Colors as in Table~\ref{tab:2k500_13x10}. BL is the uncalibrated DT; \%$\Delta$ is the best method's gain over BL. Folds is the number of target users.}
\end{table}

\begin{table}[htbp]
\TABLE
{New-user nRMSE.\label{tab:newuser_nrmse}}
{\resizebox{\textwidth}{!}{
\begin{tabular}{@{}lccccccccccccc@{}}
\hline\up
 & & \multicolumn{6}{c}{\textbf{Fit-and-transfer}} & \multicolumn{4}{c}{\textbf{Matrix Completion}} & & \\
\textbf{Persona} & \textbf{BL} & \textbf{\meth{Ridge}} & \textbf{\meth{Lasso}} & \textbf{\meth{EN}} & \textbf{\meth{NN}} & \textbf{\meth{SC}} & \textbf{\meth{SI}} & \textbf{\meth{HSV}} & \textbf{\meth{SSV}} & \textbf{\meth{ALS}} & \textbf{\meth{SP}} & \textbf{\%$\Delta$} & \textbf{Folds} \\
\hline\up
Text, GPT-4.1-mini & \cell{1.206}{0.020} & \cellbetter{0.991}{0.005} & \cellbetter{1.035}{0.006} & \cellbetter{1.017}{0.005} & \cellbetter{0.991}{0.004} & \cellbetter{0.991}{0.002} & \cellbest{0.987}{0.004} & \cellworse{1.520}{0.070} & \cellbetter{0.992}{0.004} & \cellbetter{0.998}{0.005} & \cellbetter{1.071}{0.010} & \gain{-18.1} & 200 \\
\hline
Text, Gemini-Flash-2.5 & \cell{1.257}{0.013} & \cellbest{0.981}{0.005} & \cellbetter{1.049}{0.007} & \cellbetter{1.027}{0.006} & \cellbetter{0.993}{0.005} & \cellbetter{0.991}{0.002} & \cellbetter{0.989}{0.005} & \cellworse{1.483}{0.037} & \cellbetter{0.992}{0.004} & \cellbetter{0.998}{0.006} & \cellbetter{1.064}{0.011} & \gain{-21.9} & 200 \\
\hline
JSON, GPT-4.1 & \cell{1.192}{0.016} & \cellbest{0.979}{0.005} & \cellbetter{1.054}{0.009} & \cellbetter{1.024}{0.007} & \cellbetter{0.986}{0.005} & \cellbetter{0.992}{0.002} & \cellbetter{0.989}{0.005} & \cellworse{1.318}{0.029} & \cellbetter{0.990}{0.004} & \cellbetter{0.994}{0.005} & \cellbetter{1.084}{0.013} & \gain{-17.9} & 200 \\
\hline
Text + CoT, GPT-4.1-mini & \cell{1.225}{0.013} & \cellbest{0.988}{0.005} & \cellbetter{1.036}{0.006} & \cellbetter{1.022}{0.005} & \cellbetter{0.997}{0.004} & \cellbetter{0.992}{0.002} & \cellbetter{0.990}{0.004} & \cellworse{1.566}{0.034} & \cellbetter{0.994}{0.004} & \cellbetter{1.009}{0.009} & \cellbetter{1.070}{0.011} & \gain{-19.3} & 200 \\
\hline
Fine-tuned GPT-4.1-mini & \cell{1.263}{0.024} & \cellbetter{1.020}{0.004} & \cellbetter{1.096}{0.008} & \cellbetter{1.055}{0.005} & \cellbetter{1.044}{0.004} & \cellbest{1.011}{0.002} & \cellbetter{1.014}{0.004} & \cellworse{1.346}{0.045} & \cellbetter{1.025}{0.005} & \cellbetter{1.039}{0.008} & \cellbetter{1.122}{0.017} & \gain{-20.0} & 200 \\
\hline
Demographics only, GPT-4.1-mini & \cell{1.221}{0.012} & \cellbetter{1.000}{0.004} & \cellbetter{1.058}{0.007} & \cellbetter{1.031}{0.006} & \cellbetter{1.011}{0.004} & \cellbest{0.997}{0.002} & \cellbetter{0.998}{0.004} & \cellworse{1.596}{0.038} & \cellbetter{1.007}{0.006} & \cellbetter{1.012}{0.007} & \cellbetter{1.058}{0.009} & \gain{-18.3} & 200 \\
\hline
MovieLens & \cell{1.127}{0.003} & \cellbetter{1.056}{0.003} & \cellbetter{1.079}{0.004} & \cellbetter{1.083}{0.004} & \cellbetter{1.031}{0.002} & \cellbetter{1.063}{0.005} & \cellbest{1.024}{0.002} & \cellworse{1.318}{0.015} & \cellbetter{1.056}{0.005} & \cellbetter{1.072}{0.006} & \cellbetter{1.025}{0.002} & \gain{-9.1} & 500 \down\\
\hline
\end{tabular}
}}
{Computed on the per-question standardized scale. Lower is better. The constructions and folds are those of Table~\ref{tab:newuser_corr}. Colors as in Table~\ref{tab:2k500_13x10}.}
\end{table}

\begin{table}[htbp]
\TABLE
{New-user hyperparameter selection frequencies.\label{tab:hp_newuser}}
{\resizebox{\ifdim\width>\textwidth\textwidth\else\width\fi}{!}{
\begin{tabular}{@{}llccccccccc@{}}
\toprule
& & & \multicolumn{4}{c}{MovieLens} & \multicolumn{4}{c}{Twin-2K-500} \\
\cmidrule(lr){4-7}\cmidrule(lr){8-11}
Method & Parameter & Grid & 1st & 2nd & 3rd & 4th & 1st & 2nd & 3rd & 4th \\
\midrule
\meth{Ridge} & $\lamt$ & \{0.01, 0.03, 0.1, 0.3\} & 7 & 2 & 6 & 85 & 12 & 2 & 3 & 83 \\
\meth{Lasso} & $\lamt$ & \{0.01, 0.03, 0.1, 0.3\} & 46 & 34 & 16 & 5 & 21 & 23 & 38 & 18 \\
\meth{EN} & $\lamt$ & \{0.01, 0.03, 0.1, 0.3\} & 32 & 27 & 30 & 12 & 17 & 15 & 27 & 41 \\
\meth{NN} & $\lamt$ & \{0.01, 0.03, 0.1, 0.3\} & 7 & 7 & 22 & 64 & 20 & 18 & 27 & 35 \\
\meth{SC} & $\lamt$ & \{0.01, 0.03, 0.1, 0.3\} & 28 & 1 & 5 & 65 & 36 & 0 & 1 & 62 \\
\meth{SI} & $\lamt$ & \{0.01, 0.03, 0.1, 0.3\} & 36 & 1 & 3 & 61 & 22 & 0 & 0 & 78 \\
\meth{HSV} & rank & \{3, 5, 10, 20\} & 31 & 22 & 33 & 13 & 44 & 29 & 14 & 14 \\
\meth{SSV} & shrinkage & \{1, 5, 10, 20\} & 20 & 14 & 13 & 53 & 3 & 3 & 5 & 89 \\
\meth{ALS} & shrinkage & \{1, 5, 10, 20\} & 26 & 14 & 12 & 48 & 4 & 3 & 5 & 88 \\
\meth{SP} & rank & \{3, 5, 10, 20\} & 38 & 23 & 25 & 14 & 34 & 28 & 21 & 17 \\
\bottomrule
\end{tabular}
}}
{Percentage of folds selecting each grid value. Twin-2K-500 pools the six
reported constructions.}
\end{table}

\subsection{User-space Alignment}\label{app:newuser_alignment}

\begin{figure}[htbp]
\FIGURE
{\outfig{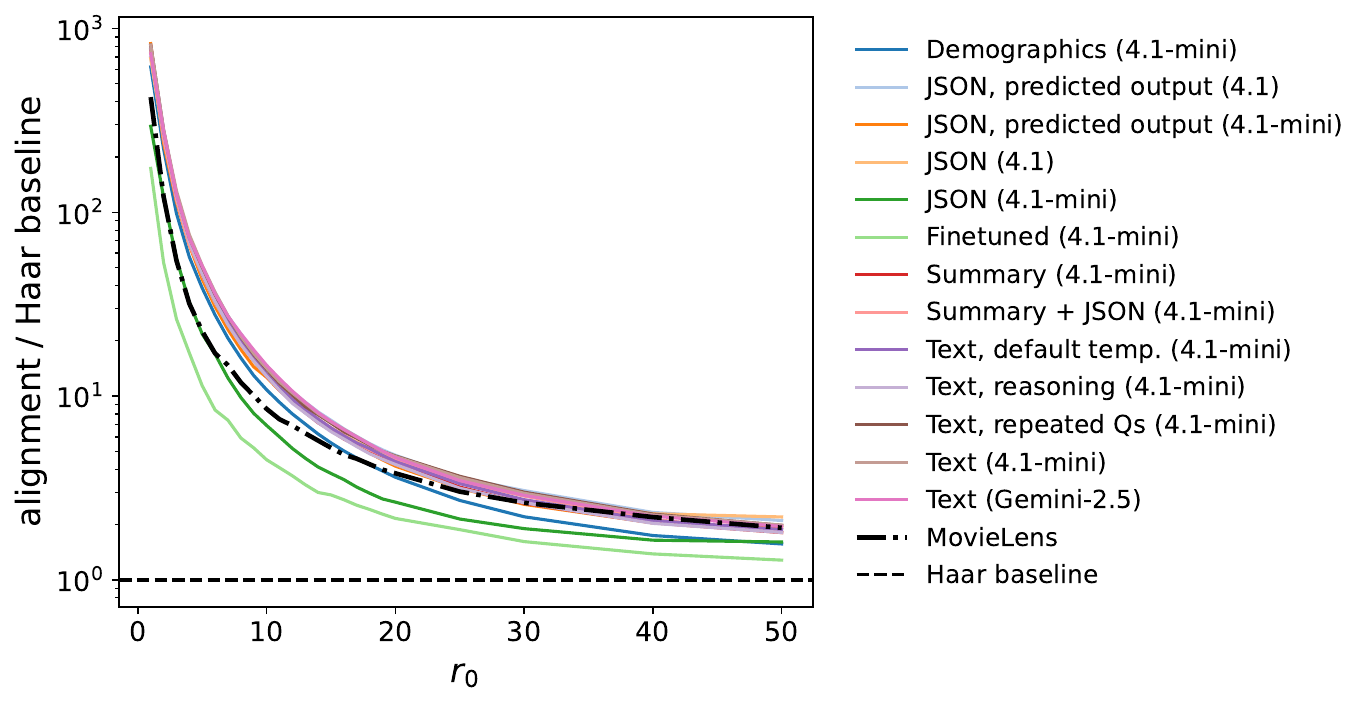}{0.90\textwidth}}
{User-space alignment relative to the Haar baseline.
\label{fig:alignment_user}}
{One line per Twin-2K-500 construction plus MovieLens, on a log scale. Every
curve remains above the unrelated-subspace benchmark through $r_0=50$.}
\end{figure}

Table~\ref{tab:alignment_user} and Figure~\ref{fig:alignment_user} apply the alignment statistic \eqref{eq:alignment} to the transposed matrices, that is, to the human and DT user subspaces, with the columns and preprocessing of Appendix~\ref{app:alignment}.

\begin{table}[htbp]
\TABLE
{User-space alignment.\label{tab:alignment_user}}
{\begin{tabular}{llccccc}
\toprule
Dataset & $r_0$ & Statistic & Range over constructions & Analytic baseline
& MC baseline band & Ratio \\
\midrule
Twin-2K-500 & 1 & $0.343$ & $0.111$--$0.402$ & $0.0005$ & $[0.0000,\,0.0022]$ & $677.7$ \\
 & 2 & $0.238$ & $0.068$--$0.273$ & $0.0011$ & $[0.0002,\,0.0027]$ & $233.6$ \\
 & 3 & $0.164$ & $0.051$--$0.191$ & $0.0016$ & $[0.0006,\,0.0030]$ & $107.3$ \\
 & 5 & $0.108$ & $0.037$--$0.125$ & $0.0027$ & $[0.0016,\,0.0041]$ & $42.4$ \\
 & 8 & $0.073$ & $0.030$--$0.085$ & $0.0043$ & $[0.0032,\,0.0056]$ & $17.9$ \\
 & 11 & $0.058$ & $0.029$--$0.067$ & $0.0059$ & $[0.0047,\,0.0073]$ & $10.3$ \\
 & 15 & $0.048$ & $0.028$--$0.057$ & $0.0081$ & $[0.0069,\,0.0093]$ & $6.2$ \\
 & 20 & $0.043$ & $0.028$--$0.053$ & $0.0108$ & $[0.0095,\,0.0119]$ & $4.1$ \\
 & 25 & $0.041$ & $0.030$--$0.054$ & $0.0134$ & $[0.0123,\,0.0147]$ & $3.1$ \\
 & 30 & $0.041$ & $0.031$--$0.057$ & $0.0161$ & $[0.0150,\,0.0172]$ & $2.6$ \\
 & 40 & $0.043$ & $0.034$--$0.066$ & $0.0215$ & $[0.0203,\,0.0227]$ & $2.0$ \\
 & 50 & $0.049$ & $0.038$--$0.080$ & $0.0269$ & $[0.0257,\,0.0280]$ & $1.8$ \\
\midrule
MovieLens & 1 & $0.847$ & --- & $0.0020$ & $[0.0000,\,0.0064]$ & $423.5$ \\
 & 2 & $0.488$ & --- & $0.0040$ & $[0.0005,\,0.0091]$ & $121.9$ \\
 & 3 & $0.328$ & --- & $0.0060$ & $[0.0021,\,0.0110]$ & $54.6$ \\
 & 5 & $0.225$ & --- & $0.0100$ & $[0.0061,\,0.0146]$ & $22.5$ \\
 & 8 & $0.190$ & --- & $0.0160$ & $[0.0120,\,0.0211]$ & $11.9$ \\
 & 11 & $0.165$ & --- & $0.0220$ & $[0.0179,\,0.0267]$ & $7.5$ \\
 & 15 & $0.157$ & --- & $0.0300$ & $[0.0260,\,0.0346]$ & $5.2$ \\
 & 20 & $0.152$ & --- & $0.0400$ & $[0.0353,\,0.0442]$ & $3.8$ \\
 & 25 & $0.151$ & --- & $0.0500$ & $[0.0461,\,0.0553]$ & $3.0$ \\
 & 30 & $0.158$ & --- & $0.0600$ & $[0.0559,\,0.0646]$ & $2.6$ \\
 & 40 & $0.176$ & --- & $0.0800$ & $[0.0761,\,0.0844]$ & $2.2$ \\
 & 50 & $0.192$ & --- & $0.1000$ & $[0.0958,\,0.1040]$ & $1.9$ \\
\bottomrule
\end{tabular}}
{Mean squared principal cosine between the human and digital-twin user subspaces, against the Haar-random baseline $r_0/n$. MovieLens has $n = 500$ throughout. The Twin-2K-500 respondent count is 2{,}058 for ten of the thirteen constructions and 2{,}050, 1{,}558, and 1{,}000 for the other three, so the baseline differs between rows of the same block. The baseline and band columns are averaged over constructions, and the ratio column is the mean of the per-construction ratios, each taken against its own $n$.}
\end{table}

\section{Additional Details for the Distribution-level Problem}\label{app:dist_details}

This section first relates $\Dnorm{\cdot}$ to classical discrepancies and explains why distribution-level candidates can be taken to be hyperplanes. It then collects the experimental protocol, prompts, and additional results for the distribution-level problem.

\subsection{Relationship between $\Dnorm{\cdot}$ and Classical Discrepancies}\label{app:discrepancies}
For every $x\in\mathcal H_0$,
\begin{equation}\label{eq:Dnormcomparison}
 \Dnorm{x}\leq\rho_D\norm{x}_1 \leq\rho_D\sqrt K\norm{x}_2.
\end{equation}
Let $C\in\R^{K\times K}$ be the lower-triangular binary cumulative-sum matrix
$$
 C_{k\ell}=\mathbf1\{\ell\leq k\},
 \qquad
 (Cx)_k=\sum_{\ell=1}^k x_\ell,
 \qquad k\in[K].
$$
For $x=P-Q$, total variation, Kolmogorov--Smirnov, CDF-$\ell_1$, and CDF-$\ell_2$ are, respectively,
$$
 \tfrac12\norm{x}_1,
 \qquad
 \norm{Cx}_\infty,
 \qquad
 \norm{Cx}_1,
 \qquad
 \norm{Cx}_2.
$$
Because $C$ is invertible, all four expressions are norms on $\mathcal H_0$. If we identify the categories with the equally spaced points $1,\ldots,K$ and use the ground metric $d(k,\ell)=|k-\ell|$, Wasserstein-1 also satisfies
$$
 W_1(P,Q)
 =\sum_{k=1}^{K-1}|F_P(k)-F_Q(k)|
 =\norm{Cx}_1,
 \qquad
 \norm{x}_{\rm KR}:=\norm{Cx}_1
 \quad(x\in\mathcal H_0).
$$
Thus, CDF-$\ell_1$ is exactly Wasserstein-1, and the last equality gives the corresponding Kantorovich--Rubinstein norm on sum-zero signed vectors.

\subsection{Why Dimension Matters Differently at the Two Levels}\label{app:inddistdimension}
For any $c\in\mathcal C_{\mathrm{aff}}$, write $z(c)=(c^\top,-1)^\top$ and $\mathcal V_c=z(c)^\perp$. Then, $\mathcal V_c$ is a $(d-1)$-dimensional hyperplane and
$$
 \one_d^\top z(c)=\one_m^\top c-1=0,
$$
so $\one_d\in\mathcal V_c$. Moreover,
$$
 \operatorname{dist}(e_\star,\mathcal V_c)
 =\frac{|e_\star^\top z(c)|}{\norm{z(c)}_2}
 =\frac1{\sqrt{1+\norm c_2^2}} > 0,
$$
and hence $e_\star\notin\mathcal V_c$. Conversely, every hyperplane $\mathcal V\subset\R^d$ that contains $\one_d$ but not $e_\star$ has a unique normal vector whose target coordinate is $-1$; writing that normal as $(c^\top,-1)^\top$ gives $\one_m^\top c=1$. Thus, $\mathcal C_{\mathrm{aff}}$ has a one-to-one correspondence to the admissible $(d-1)$-dimensional hyperplanes. More generally, any lower-dimensional subspace containing $\one_d$ and not $e_\star$ is contained in the hyperplane $z^\perp$ and has a canonical normal
$$
 z=-\frac{(I_d-P)e_\star}{\norm{(I_d-P)e_\star}_2^2}=(c^\top,-1)^\top.
$$
This explains why, while the dimension of individual-level candidate subspaces affects transfer, distribution-level candidates can be taken to be hyperplanes of dimension $d-1$. At the individual level, dimension itself participates in the transfer tradeoff. Enlarging an individual candidate can decrease the approximation defects $\Delta_{\mathrm{HU}}^{\mathrm{ind}}$ and $\Delta_{\mathrm{DT}}^{\mathrm{ind}}$, but it also adds directions on which the DT system may be weakly expressed and can decrease $\gamma_{\mathcal V}$ and increase $\norm{\beta_{\mathcal V}}_2$. In contrast, at the distribution level, any lower-dimensional candidate subspace inducing the same direction $z(c)$ can be enlarged to the hyperplane $z(c)^\perp$ of dimension $d-1$ without changing the induced coefficient $c$. Since Theorem~\ref{thm:distalg} depends on distributional structure only through this coefficient, restricting to $(d-1)$-dimensional hyperplanes is without loss of generality.

\subsection{Implementation Details}\label{app:dist_protocol}

The distribution experiment follows the per-target cross-validation rule of Appendix~\ref{app:implementation_details_ind} with a cap of 40 validation references and selects $\lamt$ by CDF-$\ell_2$ discrepancy. Algorithms~\ref{alg:distribution} and~\ref{alg:distribution_fit} give the complete procedures. The convex program~\eqref{eq:distalg} is solved by the fast iterative shrinkage-thresholding algorithm (FISTA). Algorithm~\ref{alg:distribution_fit} sums the squared reference discrepancies and scales its penalty by $n$, whereas \eqref{eq:distalg} averages the discrepancies over the $m$ references, so the penalty $\lamt$ in Algorithm~\ref{alg:distribution_fit} corresponds to $n\lamt/m$ in \eqref{eq:distalg}. The penalty acts on the twin weights alone, so it plays no role in the dummy-atoms-only ensemble. Table~\ref{tab:hp_distribution} reports the share of target questions selecting each $\lamt$; rows may not sum to 100 after rounding.

\begin{table}[htbp]
\TABLE
{Distribution-level hyperparameter selection frequencies.
\label{tab:hp_distribution}}
{
\begin{tabular}{@{}llcccc@{}}
\toprule
Dataset & Ensemble & $\lamt=0.003$ & $\lamt=0.03$ & $\lamt=0.3$ & $\lamt=3$ \\
\midrule
OpinionQA & Both & 29 & 45 & 22 & 4 \\
 & Twins only & 50 & 47 & 3 & 0 \\
MovieLens & Both & 20 & 61 & 19 & 0 \\
 & Twins only & 64 & 36 & 0 & 0 \\
\bottomrule
\end{tabular}
}
{Percentage of target questions selecting each $\lamt$ of Algorithm~\ref{alg:distribution_fit}. The dummy-only
ensemble is omitted because its fit does not depend on the twin penalty.}
\end{table}

\begin{breakablealgorithm}
\caption{Distribution calibration, Table~\ref{tab:distsim_summary}.}
\label{alg:distribution}
\small
\begin{algorithmic}[1]
\Require human population marginals $P_1, \dots, P_d \in \Delta^{K-1}$ on the $K$
  ordered answer categories; twin answers $\tY \in \{1, \dots, K\}^{n \times d}$
  from $n$ sampled twins on $d$ questions; an ensemble variant; hyperparameter
  grid $\Lambda$
\For{$j \le d$}
  \State $\mathcal D \gets \{1, \dots, d\} \setminus \{j\}$
  \State draw $\mathcal D_{\mathrm{val}} \subset \mathcal D$ with $|\mathcal D_{\mathrm{val}}| = \operatorname{clip}\big(
    \operatorname{round}(0.1|\mathcal D|),\, 3,\, 40\big)$, and set $\mathcal D_{\mathrm{fit}} \gets \mathcal D \setminus \mathcal D_{\mathrm{val}}$
  \For{$\lamt \in \Lambda$}
    \State $(w^{\lamt}, \pi^{\lamt}) \gets \Call{Fit}{\mathcal D_{\mathrm{fit}}, \lamt}$
    \State $g(\lamt) \gets |\mathcal D_{\mathrm{val}}|^{-1} \sum_{v \in \mathcal D_{\mathrm{val}}} \big\lVert C\big(P_v -
      \widehat P_v(w^{\lamt}, \pi^{\lamt})\big) \big\rVert_2$
  \EndFor
  \State $\hat\lamt \gets \argmin_{\lamt \in \Lambda} g(\lamt)$
  \State $(\hat w, \hat\pi) \gets \Call{Fit}{\mathcal D, \hat\lamt}$
    \Comment{refit on \emph{every} reference}
  \State $\delta_j \gets \big\lVert C\big(P_j - \widehat P_j(\hat w,
    \hat\pi)\big)\big\rVert_2$, and similarly under CDF-$\ell_1$,
    TV and KS
  \State $\delta^{\mathrm{BL}}_j \gets$ the same four with $(\hat w, \hat\pi)$
    replaced by $(\mathbf 1/n, 0)$
      \Comment{uniform twins, no dummy mass}
\EndFor
\State $\bar\delta \gets d^{-1} \sum_{j=1}^{d} \delta_j$ and $\se(\bar\delta)
  \gets d^{-1/2} \operatorname{sd}_j(\delta_j)$, and likewise
  $\bar\delta^{\mathrm{BL}}$ and $\se(\bar\delta^{\mathrm{BL}})$ from
  $\delta^{\mathrm{BL}}_j$, componentwise over the four discrepancies
\State \Return $\bar\delta,\ \se(\bar\delta),\ \bar\delta^{\mathrm{BL}},\
  \se(\bar\delta^{\mathrm{BL}})$
\end{algorithmic}
\end{breakablealgorithm}

\begin{breakablealgorithm}
\caption{Distributional ensemble fitting procedure.}
\label{alg:distribution_fit}
\small
\begin{algorithmic}[1]
\Procedure{Fit}{reference questions $Q$, penalty $\lamt$}
  \If{Twins only}
    \State $\Theta \gets \big\{(w, \pi) \in \R^{n}_{\ge 0} \times \R^{K}_{\ge 0}
    \,:\, \mathbf 1^\top w + \mathbf 1^\top \pi = 1, \, \pi = 0 \big\}$
  \ElsIf{Dummy atoms only}
    \State $\Theta \gets \big\{(w, \pi) \in \R^{n}_{\ge 0} \times \R^{K}_{\ge 0}
    \,:\, \mathbf 1^\top w + \mathbf 1^\top \pi = 1, \, w = 0 \big\}$
  \Else
    \State $\Theta \gets \big\{(w, \pi) \in \R^{n}_{\ge 0} \times \R^{K}_{\ge 0} : \mathbf 1^\top w + \mathbf 1^\top \pi = 1 \big\}$
  \EndIf
  \State \Return $\displaystyle \argmin_{(w, \pi) \in \Theta} \ \sum_{q \in Q}
    \big\lVert C\big(P_q - \widehat P_q(w, \pi)\big) \big\rVert_2^2 \;+\; \lamt
    \, n \, \lVert w \rVert_2^2$
\EndProcedure
\end{algorithmic}
\end{breakablealgorithm}

\subsection{Distribution-level Prompts}\label{app:distribution_prompts}

For MovieLens and OpinionQA, each of the 2,058 Twin-2K-500 summary personas is presented with one question at a time. MovieLens uses title, genre, and ten tags and requests a rating from $\left\{0.5, 1, \ldots, 5\right\}$. OpinionQA presents the survey question and requests an integer from one to five corresponding to the answer choice. GPT-4.1-mini is queried at temperature zero.

\begin{tcolorbox}[title=Distribution-level MovieLens: system prompt]
You, AI, are an expert in predicting human preferences. You are given a
persona profile and a movie to rate, and also a format instruction that
specifies the type of rating you need to provide. You need to rate the movie
as the persona would rate it, based on the persona profile and the format
instructions.
\end{tcolorbox}

\begin{tcolorbox}[title=Distribution-level MovieLens: user prompt]
\texttt{\{persona\}}\\
\texttt{MOVIE: \{movie\}}\\
\texttt{GENRE: \{genre\}}\\
\texttt{TOP 10 TAGS: \{top\_10\_tags\}}\\
\texttt{FORMAT INSTRUCTIONS: Only a number on a 5-star rating scale, with
half-star increments (0.5--5.0). Larger numbers indicate higher ratings.}
\end{tcolorbox}

\begin{tcolorbox}[title=OpinionQA: system prompt]
You, AI, are an expert in predicting human answers to survey questions. You
are given a persona profile and a survey question to answer, and also a format
instruction that specifies the type of answer you need to provide. You need
to answer the survey question as the persona would answer it, based on the
persona profile and the format instructions.
\end{tcolorbox}

\begin{tcolorbox}[title=OpinionQA: user prompt]
\texttt{\{persona\}}\\
\texttt{SURVEY QUESTION: \{question\}}\\
\texttt{FORMAT INSTRUCTIONS: Only an integer number from 1 to 5, corresponding
to the choice of answer to the survey question.}
\end{tcolorbox}

\subsection{Ensemble Composition}\label{app:dist_composition}

Table~\ref{tab:dist_diagnostics} reports the composition of the fitted ensembles.

\begin{table}[htbp]
\TABLE
{Ensemble composition.\label{tab:dist_diagnostics}}
{
\begin{tabular}{llccc}
\toprule
Dataset & Ensemble & Twin mass & Dummy mass & ESS \\
\midrule
OpinionQA & Both & 0.469 & 0.531 & 119 \\
 & Twins only & 1.000 & 0.000 & 43 \\
 & Dummy atoms only & 0.000 & 1.000 & --- \\
\midrule
MovieLens & Both & 0.627 & 0.373 & 92 \\
 & Twins only & 1.000 & 0.000 & 32 \\
 & Dummy atoms only & 0.000 & 1.000 & --- \\
\bottomrule
\end{tabular}
}
{Composition of the fitted distributional ensembles. Twin and dummy mass are the mean fitted weights over target questions. $\ESS(w)=(\one^\top w)^2/\norm w_2^2$ is the effective sample size of the DT weights and is undefined for the dummy-atoms-only ensemble.}
\end{table}

\subsection{Variance Retention}\label{app:dist_variance}

\begin{figure}[htbp]
\FIGURE
{\outfig{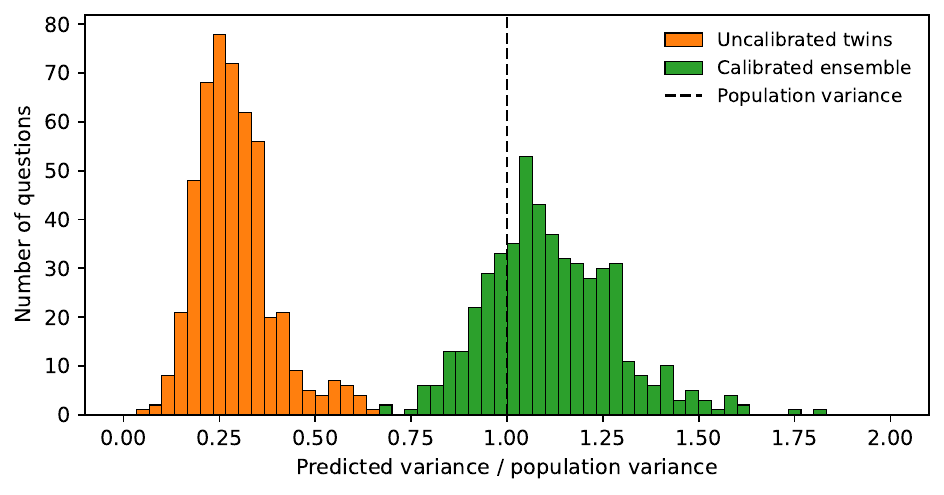}{0.74\textwidth}}
{Variance retention on MovieLens.\label{fig:movielens_variance_retention}}
{A histogram of predicted rating variance divided by the human population variance. The median is $0.282$ for uncalibrated DTs and $1.098$ for the calibrated ensemble.}
\end{figure}

Figure~\ref{fig:movielens_variance_retention} compares, for each MovieLens movie, the variance of the predicted rating distribution with that of the human population. The uncalibrated DTs are over-concentrated, while the calibrated ensemble restores the population spread.

\section{Additional Diagnostics}\label{app:diagnostics}

\subsection{Null Digital Twins}\label{app:null_twin}

Do the twins really need to simulate the corresponding users for calibration to work? We rerun MovieLens and the Text, GPT-4.1-mini construction of Twin-2K-500 with two baselines carrying no respondent-level twin signal: the \textbf{Gaussian null} draws independent entries matching the DT matrix's global mean, standard deviation, and missingness pattern; the \textbf{shuffled null} replaces the DT matrix by the human matrix with each column independently permuted, preserving its marginal distribution and missing count but breaking respondent identity. 

Table~\ref{tab:null_twin} shows the results. The reference mean is a twin-free benchmark: the correlation between the target and each respondent's average standardized reference response. Under both nulls, every method except \meth{SC} falls to mean correlation between $-.028$ and $.045$. \meth{SC} instead falls to the reference mean because its simplex-weighted prediction approaches an average of the respondent's standardized human references. With real twins, \meth{SC} remains above its null results. For these two matrix pairs, the gains over the uncalibrated DT therefore depend on the twin signal rather than on the fitting procedure alone.

\begin{table}[htbp]
\TABLE
{Null digital twins.\label{tab:null_twin}}
{\resizebox{\textwidth}{!}{
\begin{tabular}{@{}lcccccccccccc@{}}
\hline\up
 & & \multicolumn{6}{c}{\textbf{Fit-and-transfer}} & \multicolumn{4}{c}{\textbf{Matrix Completion}} & \\
\textbf{Twin matrix} & \textbf{BL} & \textbf{\meth{Ridge}} & \textbf{\meth{Lasso}} & \textbf{\meth{EN}} & \textbf{\meth{NN}} & \textbf{\meth{SC}} & \textbf{\meth{SI}} & \textbf{\meth{HSV}} & \textbf{\meth{SSV}} & \textbf{\meth{ALS}} & \textbf{\meth{SP}} & \textbf{Folds} \\
\hline\up
\emph{MovieLens} &  &  &  &  &  &  &  &  &  &  &  &  \\
Twins & \cell{.349}{.005} & \cellbest{.516}{.005} & \cellbetter{.503}{.006} & \cellbetter{.514}{.005} & \cellbetter{.501}{.005} & \cellbetter{.444}{.005} & \cellbetter{.509}{.005} & \cellbetter{.471}{.007} & \cellbetter{.513}{.005} & \cellbetter{.513}{.005} & \cellbetter{.450}{.005} & 250 \\
Gaussian null & \cell{.002}{.003} & \cell{.011}{.012} & \cell{-.002}{.010} & \cell{-.008}{.011} & \cell{.045}{.011} & \cellbold{.438}{.005} & \cell{.014}{.012} & \cell{.028}{.014} & \cell{.021}{.011} & \cell{.024}{.011} & \cell{.014}{.015} & 250 \\
Shuffled null & \cell{.006}{.004} & \cell{-.019}{.012} & \cell{.005}{.010} & \cell{.000}{.012} & \cell{-.005}{.011} & \cellbold{.436}{.005} & \cell{.006}{.011} & \cell{-.001}{.016} & \cell{-.006}{.012} & \cell{-.003}{.012} & \cell{.020}{.014} & 250 \\
Reference mean & \cell{.439}{.005} & --- & --- & --- & --- & --- & --- & --- & --- & --- & --- & 250 \down\\
\hline\up
\emph{Twin-2K-500 (Text, GPT-4.1-mini)} &  &  &  &  &  &  &  &  &  &  &  &  \\
Twins & \cell{.203}{.016} & \cellbest{.254}{.022} & \cellbetter{.250}{.023} & \cellbetter{.250}{.023} & \cellbetter{.250}{.022} & \cellbetter{.210}{.015} & \cellbetter{.243}{.018} & \cellbetter{.211}{.019} & \cellbetter{.240}{.020} & \cellbetter{.240}{.019} & \cellworse{.138}{.018} & 102 \\
Gaussian null & \cell{.001}{.003} & \cell{-.014}{.008} & \cell{-.012}{.006} & \cell{-.016}{.007} & \cell{-.010}{.007} & \cellbold{.102}{.013} & \cell{-.028}{.010} & \cell{-.021}{.013} & \cell{-.014}{.010} & \cell{-.028}{.010} & \cell{-.011}{.012} & 102 \\
Shuffled null & \cell{-.001}{.004} & \cell{-.004}{.009} & \cell{-.013}{.008} & \cell{-.003}{.009} & \cell{.007}{.007} & \cellbold{.108}{.012} & \cell{-.013}{.011} & \cell{.003}{.013} & \cell{-.004}{.009} & \cell{-.007}{.008} & \cell{.008}{.009} & 102 \\
Reference mean & \cell{.123}{.014} & --- & --- & --- & --- & --- & --- & --- & --- & --- & --- & 102 \down\\
\hline
\end{tabular}
}}
{New-question correlation with real and null DT matrices. The folds and calibration procedures are unchanged. BL is the uncalibrated matrix in that row. Colors in the twin rows are as in Table~\ref{tab:2k500_13x10}; null rows mark the best method in bold.}
\end{table}

\subsection{Twin and Human Moments}\label{app:twin_moments}

For each question, Table~\ref{tab:twin_moments} compares the DT and human moments over respondents observed on both sides. The mean shift is measured in human standard deviations, and the SD ratio is DT over human standard deviation. The median SD ratio is below one for every construction, ranging from 0.490 to 0.951 on Twin-2K-500, and is 0.685 on MovieLens. DT responses are therefore systematically less dispersed.

\begin{table}[htbp]
\TABLE
{DT and human moments.\label{tab:twin_moments}}
{
\begin{tabular}{@{}lccccc@{}}
\toprule
& \multicolumn{3}{c}{Mean shift} & \multicolumn{2}{c}{SD ratio} \\
\cmidrule(lr){2-4}\cmidrule(lr){5-6}
Persona & Median & IQR & Median abs. & Median & Share $<1$ \\
\midrule
Text, GPT-4.1-mini & 0.185 & [-0.448, 0.324] & 0.359 & 0.763 & 0.93 \\
Text, Gemini-Flash-2.5 & -0.299 & [-0.413, 0.112] & 0.365 & 0.951 & 0.83 \\
JSON, GPT-4.1-mini & -0.245 & [-0.418, 0.197] & 0.331 & 0.634 & 0.84 \\
JSON, GPT-4.1 & -0.015 & [-0.403, 0.095] & 0.177 & 0.801 & 0.84 \\
Text + CoT, GPT-4.1-mini & -0.254 & [-0.349, 0.272] & 0.320 & 0.640 & 0.81 \\
Text + repeat, GPT-4.1-mini & 0.125 & [-0.419, 0.205] & 0.251 & 0.653 & 0.96 \\
Text, temp=0.7, GPT-4.1-mini & 0.189 & [-0.407, 0.324] & 0.353 & 0.790 & 0.92 \\
JSON + PO, GPT-4.1-mini & -0.427 & [-0.613, 0.271] & 0.543 & 0.496 & 0.97 \\
JSON + PO, GPT-4.1 & -0.114 & [-0.274, 0.063] & 0.173 & 0.493 & 0.68 \\
Fine-tuned GPT-4.1-mini & -0.338 & [-0.492, -0.054] & 0.366 & 0.762 & 0.93 \\
Demographics only, GPT-4.1-mini & -0.434 & [-0.632, 0.502] & 0.588 & 0.490 & 0.99 \\
Summary, GPT-4.1-mini & -0.346 & [-0.471, 0.337] & 0.447 & 0.680 & 0.89 \\
Summary + JSON, GPT-4.1-mini & -0.479 & [-0.711, 0.187] & 0.588 & 0.737 & 0.92 \\
\midrule
MovieLens & 0.261 & [0.084, 0.494] & 0.283 & 0.685 & 0.98 \\
\bottomrule
\end{tabular}
}
{Question-level differences between DT and human moments. Mean shift reports its median, interquartile range, and median absolute value over questions. The Twin-2K-500 matrices have 123 questions; MovieLens has 250.}
\end{table}

\subsection{Computational Cost}\label{app:runtime}

We record in Table~\ref{tab:runtime} the computational costs of each method on one core of an Intel Xeon processor. Folds are mutually independent and can be run in parallel.

\begin{table}[htbp]
\TABLE
{Computational cost.\label{tab:runtime}}
{
\begin{tabular}{@{}lcccc@{}}
\toprule
& \multicolumn{2}{c}{New question} & \multicolumn{2}{c}{New user} \\
\cmidrule(lr){2-3}\cmidrule(lr){4-5}
Method & MovieLens & Twin-2K-500 & MovieLens & Twin-2K-500 \\
\midrule
\meth{Ridge} & 0.62 & 0.62 & 0.83 & 0.93 \\
\meth{Lasso} & 0.74 & 0.59 & 0.76 & 1.85 \\
\meth{EN} & 1.22 & 0.63 & 0.78 & 5.34 \\
\meth{NN} & 18.1 & 41.3 & 10.7 & 8.66 \\
\meth{SC} & 5.37 & 10.0 & 5.92 & 11.9 \\
\meth{SI} & 1.23 & 0.98 & 1.40 & 1.34 \\
\meth{HSV} & 12.1 & 10.3 & 46.0 & 20.7 \\
\meth{SSV} & 12.8 & 19.5 & 23.7 & 24.0 \\
\meth{ALS} & 3.34 & 10.5 & 2.99 & 7.45 \\
\meth{SP} & 6.39 & 4.55 & 6.50 & 4.57 \\
\midrule
& \multicolumn{2}{c}{OpinionQA} & \multicolumn{2}{c}{MovieLens} \\
\cmidrule(lr){2-3}\cmidrule(lr){4-5}
Distribution & \multicolumn{2}{c}{22.5} & \multicolumn{2}{c}{23.0} \\
\bottomrule
\end{tabular}
}
{Median wall-clock seconds for one leave-one-out fold. Each time includes grid search, refit, and prediction on one core of an Intel Xeon processor. Ten new-question and five new-user folds are timed per dataset and method; the final row uses ten distribution-level targets per dataset.}
\end{table}

\section{Proofs}\label{app:proofs}

\subsection{Proof of Proposition~\ref{prop:indexact}}\label{app:indexactproof}
Write $z(\beta)=(\beta^\top,-1)^\top$. Since $\row(M)\subseteq\row(\tM)$, $M=M\tM^\dagger\tM$. Hence, with $L:=\norm{M\tM^\dagger}_{\op}<\infty$,
\begin{equation}\label{eq:warmuptransferproof}
 \norm{Mz}_2\leq L\norm{\tM z}_2,
 \qquad z\in\R^{m+1}.
\end{equation}
In particular, if $\tM z(\beta^\star)=0$, then $Mz(\beta^\star)=0$. Denote
$$
\xi_{HU}(\alpha)
 =\sigma_{HU}\left(1+\sqrt{m/n}
   +\sqrt{\log(4/\alpha)/n}\right).
$$
By a standard sub-Gaussian operator-norm bound \citep[e.g.,][Theorem~4.4.3]{vershynin2026high} and a union bound, there is an absolute constant $C$ such that, with probability at least $1-\alpha$,
\begin{equation}\label{eq:warmupnoiseevent}
 \frac1{\sqrt n}\norm{\tE}_{\op}
 \leq C\xi_{DT,S}(\alpha),
 \qquad
 \frac1{\sqrt n}\norm{\mathcal E_{\mathrm{ref}}}_{\op}
 \leq C\xi_{HU}(\alpha).
\end{equation}
Suppose the event holds. By \eqref{eq:indexactlambda}, there are absolute constants $0<c_\lambda\leq C_\lambda<\infty$ such that $c_\lambda m\xi_{DT,S}(\alpha)^2 \leq \lambda \leq C_\lambda m\xi_{DT,S}(\alpha)^2$. Let $C_\beta$ be a fixed constant such that $\norm{\beta^\star}_2\leq C_\beta/\sqrt m$. Ridge optimality against $\beta^\star$ gives
\begin{align*}
 &\frac1n
 \norm{\tY z(\widehat\beta_{\lambda})}_2^2
 +\lambda\norm{\widehat\beta_{\lambda}}_2^2\leq
 \frac1n\norm{\tE z(\beta^\star)}_2^2
 +\lambda\norm{\beta^\star}_2^2
 \lesssim_{C_\beta}\xi_{DT,S}(\alpha)^2.
\end{align*}
The lower bound on $\lambda$ therefore implies
\begin{equation}\label{eq:warmupbetabound}
 \norm{\widehat\beta_{\lambda}}_2
 \lesssim_{C_\beta}m^{-1/2},
\end{equation}
while ridge optimality also gives
\begin{equation}\label{eq:warmupfitbound}
 \frac1{\sqrt n}
 \norm{\tY z(\widehat\beta_{\lambda})}_2
 \lesssim_{C_\beta} \xi_{DT,S}(\alpha).
\end{equation}
Using \eqref{eq:warmupnoiseevent}, \eqref{eq:warmupbetabound}, and \eqref{eq:warmupfitbound}, the triangle inequality gives
$$
 \frac1{\sqrt n}\norm{\tM z(\widehat\beta_{\lambda})}_2
 \leq
 \frac1{\sqrt n}
 \norm{\tY z(\widehat\beta_{\lambda})}_2
 +\frac1{\sqrt n}
 \norm{\tE}_{\op}
 \norm{z(\widehat\beta_{\lambda})}_2
 \lesssim_{C_\beta} \xi_{DT,S}(\alpha).
$$
Finally,
\begin{align*}
 \operatorname{Err}(Y_{\mathrm{ref}}\widehat\beta_{\lambda})
 &\leq
 \frac1{\sqrt n}\norm{M z(\widehat\beta_{\lambda})}_2
 +\frac1{\sqrt n}
 \norm{\mathcal E_{\mathrm{ref}}\widehat\beta_{\lambda}}_2\\
 &\lesssim_{M,\tM,C_\beta}
 \xi_{DT,S}(\alpha)+m^{-1/2}\xi_{HU}(\alpha)\\
 &=
 \xi_{DT,S}(\alpha)
 +
 \sigma_{HU}\left\{
 \frac1{\sqrt m}+\frac1{\sqrt n}
 +\sqrt{\frac{\log(4/\alpha)}{mn}}
 \right\},
\end{align*}
where the first term uses \eqref{eq:warmuptransferproof} and the second uses \eqref{eq:warmupnoiseevent}--\eqref{eq:warmupbetabound}. This proves \eqref{eq:indexacthighprob}. \Halmos

\subsection{Proof of Theorem~\ref{thm:ridge}}
\label{app:indridgeproof}

Before proving the theorem, we first establish a proposition that works for any linear method in the fit-and-transfer family, including \meth{Ridge}, \meth{Lasso}, \meth{EN}, \meth{SC}, and \meth{SI}. Denote $B_{\mathcal V}=\norm{\beta_{\mathcal V}}_2$.

\begin{proposition}[Candidate-space deterministic certificate]
\label{prop:indcandidate}
For every $\mathcal V\in\mathfrak V$ and $\beta\in\R^m$,
\begin{align}
 \sqrt{\mathcal R^{\mathrm{exc}}(\beta)}
 \leq
 \frac1{\gamma_{\mathcal V}}\widehat\tau_S(\beta)
 +
 \left[
 \Delta_{\mathrm{HU}}^{\mathrm{ind}}(\mathcal V)
 +\frac1{\gamma_{\mathcal V}}
 \left(
   \Delta_{\mathrm{DT}}^{\mathrm{ind}}(\mathcal V)
   +\frac1{\sqrt n}\norm{\tE}_{\op}
 \right)
 \right]
 \sqrt{1+\norm\beta_2^2}
 +\frac1{\sqrt n}\norm{\mathcal E_{\mathrm{ref}}\beta}_2.
 \label{eq:indcandidate}
\end{align}
\end{proposition}
\proof{Proof of Proposition~\ref{prop:indcandidate}.}
Fix a candidate $\mathcal V\in\mathfrak V$ and abbreviate $P=P_{\mathcal V}$ and $z=z(\beta)$.  Decompose $z=(I_d-P)z+Pz$.  The triangle inequality and the definition of $\Delta_{\mathrm{HU}}^{\mathrm{ind}}(\mathcal V)$ give
\begin{align}
 \frac1{\sqrt n}\norm{Mz}_2
 \leq
 \frac1{\sqrt n}\norm{M(I_d-P)z}_2
 +\frac1{\sqrt n}\norm{MPz}_2 \leq
 \Delta_{\mathrm{HU}}^{\mathrm{ind}}(\mathcal V)\norm z_2
 +\frac1{\sqrt n}\norm{M}_{\op}\norm{Pz}_2.
 \label{eq:indcandidateMproof}
\end{align}
If $Pz\neq0$, then $Pz/\norm{Pz}_2$ is a unit vector in $\mathcal V$, and the definition of $\gamma_{\mathcal V}$ gives
\begin{equation}\label{eq:indcandidategamma}
 \frac{\gamma_{\mathcal V}}{\sqrt n}\norm{M}_{\op}\norm{Pz}_2
 \leq\frac1{\sqrt n}\norm{\tM Pz}_2.
\end{equation}
The same inequality is trivial when $Pz=0$. Since $\tM Pz=\tM z-\tM(I_d-P)z$ and $\tM z=\tY z-\tE z$,
\begin{align}
 \frac1{\sqrt n}\norm{\tM Pz}_2
 &\leq
 \frac1{\sqrt n}\norm{\tM z}_2
 +\frac1{\sqrt n}\norm{\tM(I_d-P)z}_2 \notag\\
 &\leq
 \frac1{\sqrt n}\norm{\tY z}_2
 +\frac1{\sqrt n}\norm{\tE z}_2
 +\Delta_{\mathrm{DT}}^{\mathrm{ind}}(\mathcal V)\norm z_2 \notag\\
 &\leq
 \widehat\tau_S(\beta)
 +\left[\frac1{\sqrt n}\norm{\tE}_{\op}
 +\Delta_{\mathrm{DT}}^{\mathrm{ind}}(\mathcal V)\right]\norm z_2.
 \label{eq:indcandidatePproof}
\end{align}
Combining \eqref{eq:indcandidategamma} and \eqref{eq:indcandidatePproof} yields
$$
 \frac1{\sqrt n}\norm{M}_{\op}\norm{Pz}_2
 \leq
 \frac1{\gamma_{\mathcal V}}\widehat\tau_S(\beta)
 +\frac1{\gamma_{\mathcal V}}
 \left\{\frac1{\sqrt n}\norm{\tE}_{\op}
 +\Delta_{\mathrm{DT}}^{\mathrm{ind}}(\mathcal V)\right\}\norm z_2.
$$
Substitution in \eqref{eq:indcandidateMproof} gives
$$
 \frac1{\sqrt n}\norm{Mz}_2
 \leq \frac1{\gamma_{\mathcal V}}\widehat\tau_S(\beta)+\left[
 \Delta_{\mathrm{HU}}^{\mathrm{ind}}(\mathcal V)
 +\frac1{\gamma_{\mathcal V}}
 \left\{\frac1{\sqrt n}\norm{\tE}_{\op}
 +\Delta_{\mathrm{DT}}^{\mathrm{ind}}(\mathcal V)\right\}
 \right]\norm z_2.
$$
Finally, since $Mz=M_{\mathrm{ref}}\beta-M_\star$ and $Y_{\mathrm{ref}}=M_{\mathrm{ref}}+\mathcal E_{\mathrm{ref}}$, $Y_{\mathrm{ref}}\beta-M_\star=Mz+\mathcal E_{\mathrm{ref}}\beta$. Using \eqref{eq:indexcessrisk} and $\norm{z(\beta)}_2=\sqrt{1+\norm\beta_2^2}$ proves \eqref{eq:indcandidate}.  Every inequality holds for all $\beta$ and all candidate subspaces on the same realized matrices. \Halmos
\endproof

We now return to the proof of Theorem~\ref{thm:ridge}. Conditional on $\mathcal L$, the entries of $\mathcal E_{\mathrm{ref}}$ are independent, mean zero, and have sub-Gaussian norms at most $\sigma_{HU}$. A standard sub-Gaussian operator-norm bound \citep[e.g.,][Theorem~4.4.3]{vershynin2026high} and a union bound therefore give, with conditional probability at least $1-\alpha$,
\begin{equation}\label{eq:indhumanopproof}
 \frac1{\sqrt n}\norm{\mathcal E_{\mathrm{ref}}}_{\op}
 \leq C\xi_{HU}(\alpha),
 \qquad
 \frac1{\sqrt n}\norm{\tE}_{\op}
 \leq C\xi_{DT,S}(\alpha).
\end{equation}
We assume that this event holds for the remainder of the proof. For a coefficient budget $B>0$, define the population DT residual profile
\begin{equation}\label{eq:indpopprofile}
 e_{DT}(B)
 :=\min_{\norm\beta_2\leq B}
   \frac1{\sqrt n}\norm{\tM_{\mathrm{ref}}\beta-\tM_\star}_2
\end{equation}
and its finite-sample adjusted form
\begin{equation*}
 q_{S,\alpha}(B)
 :=e_{DT}(B)+\sqrt{1+B^2}\,\xi_{DT,S}(\alpha).
\end{equation*}
Writing
\begin{equation*}
 A_{S,\alpha}(\mathcal V)
 :=\Delta_{\mathrm{HU}}^{\mathrm{ind}}(\mathcal V)
 +\frac1{\gamma_{\mathcal V}}\left\{
   \Delta_{\mathrm{DT}}^{\mathrm{ind}}(\mathcal V)
   +\xi_{DT,S}(\alpha)
 \right\},
\end{equation*}
a budgeted version of the bound in \eqref{eq:indoraclebound} is
\begin{equation*}
 \Phi_{S,\alpha}(B)
 :=(1+B)
   \inf_{\substack{\mathcal V\in\mathfrak V\\B_{\mathcal V}\leq B}}
   A_{S,\alpha}(\mathcal V)
   +B\xi_{HU}(\alpha).
\end{equation*}
The constrained candidate family is nonempty for every $B>0$: if there exists $u\perp e_\star$ with $\tM u\neq0$, then $\operatorname{span}(u)$ is admissible and has coefficient norm zero; otherwise $\tM$ vanishes on $e_\star^\perp$, and $v=e_\star+t u$ with unit $u\perp e_\star$ and large $t$ gives an admissible one-dimensional candidate with coefficient norm $1/t\leq B$.

We first establish a bound for an arbitrary coefficient budget $B>0$.  By compactness of the Euclidean ball, let $\beta^B$ attain the minimum in \eqref{eq:indpopprofile}. Recall that $\widehat\tau_S(\beta) =\frac1{\sqrt n} \norm{\tY_{\mathrm{ref}}\beta-\tY_\star}_2$. From $\tY=\tM+\tE$, \eqref{eq:indhumanopproof}, and $\norm{z(\beta^B)}_2\leq\sqrt{1+B^2}$,
\begin{equation*}
 \widehat\tau_S(\beta^B)
 \leq
 e_{DT}(B)+C\sqrt{1+B^2}\,\xi_{DT,S}(\alpha)
 \lesssim q_{S,\alpha}(B).
\end{equation*}
Choose
\begin{equation*}
 \lambda_B^\circ:=\frac{q_{S,\alpha}(B)^2}{B^2}
\end{equation*}
and recall $\widehat\beta_\lambda \in\argmin_{\beta\in\R^m} \left\{\widehat\tau_S(\beta)^2+\lambda\norm{\beta}_2^2\right\}$. When $q_{S,\alpha}(B)>0$, ridge optimality relative to $\beta^B$ gives
\begin{align*}
 \widehat\tau_S(\widehat\beta_{\lambda_B^\circ})^2
 +\lambda_B^\circ\norm{\widehat\beta_{\lambda_B^\circ}}_2^2
 &\leq
 \widehat\tau_S(\beta^B)^2+\lambda_B^\circ\norm{\beta^B}_2^2 \lesssim q_{S,\alpha}(B)^2.
\end{align*}
Bounding each term on the left by the right-hand side, we obtain
\begin{equation}\label{eq:indridgeBproperties}
 \widehat\tau_S(\widehat\beta_{\lambda_B^\circ})
 \lesssim q_{S,\alpha}(B),
 \qquad
 \norm{\widehat\beta_{\lambda_B^\circ}}_2\lesssim B.
\end{equation}
If $q_{S,\alpha}(B)=0$ (only possible when $e_{DT}(B)=0$ and $\sigma_{DT}=0$), take $\lambda_B^\circ=0$.  Then $\beta^B$ is an exact-fit coefficient of norm at most $B$, and the minimum-norm convention for $\widehat\beta_0$ yields \eqref{eq:indridgeBproperties} as well.

Now fix any $\mathcal V\in\mathfrak V$ with $B_{\mathcal V}\leq B$. Because $P_{\mathcal V}z_{\mathcal V}=0$,
\begin{align*}
 e_{DT}(B)
 &\leq \frac1{\sqrt n}\norm{\tM z_{\mathcal V}}_2 \leq
 \Delta_{\mathrm{DT}}^{\mathrm{ind}}(\mathcal V)
 \norm{z_{\mathcal V}}_2
 =\Delta_{\mathrm{DT}}^{\mathrm{ind}}(\mathcal V)
  \sqrt{1+B_{\mathcal V}^2}.
\end{align*}
Hence,
\begin{equation}\label{eq:indqcandidate}
 q_{S,\alpha}(B)
 \lesssim
 (1+B)\left\{
 \Delta_{\mathrm{DT}}^{\mathrm{ind}}(\mathcal V)
 +\xi_{DT,S}(\alpha)
 \right\}.
\end{equation}
Apply Proposition~\ref{prop:indcandidate} to $\widehat\beta_{\lambda_B^\circ}$ and use \eqref{eq:indhumanopproof}--\eqref{eq:indqcandidate}.  Since $\sqrt{1+\norm{\widehat\beta_{\lambda_B^\circ}}_2^2}\lesssim1+B$, we obtain
\begin{align*}
 \sqrt{\mathcal R^{\mathrm{exc}}
       (\widehat\beta_{\lambda_B^\circ})}
 \lesssim{}&
 (1+B)A_{S,\alpha}(\mathcal V)
 +B\xi_{HU}(\alpha).
\end{align*}
Taking the infimum over $\mathcal V$ with $B_{\mathcal V}\leq B$ yields
\begin{equation}\label{eq:indbudgetPhiProof}
 \sqrt{\mathcal R^{\mathrm{exc}}
       (\widehat\beta_{\lambda_B^\circ})}
 \lesssim\Phi_{S,\alpha}(B).
\end{equation}

It remains to optimize over the coefficient budget.  Define the unrestricted candidate-level oracle value
$$
 \Psi_{S,\alpha}
 :=
 \inf_{\mathcal V\in\mathfrak V}
 \left[
 (1+B_{\mathcal V})A_{S,\alpha}(\mathcal V)
 +B_{\mathcal V}\xi_{HU}(\alpha)
 \right].
$$
We claim that
\begin{equation}\label{eq:indPhiIdentity}
 \inf_{B>0}\Phi_{S,\alpha}(B)=  \inf_{B>0}\left[(1+B)
   \inf_{\substack{\mathcal V\in\mathfrak V\\B_{\mathcal V}\leq B}}
   A_{S,\alpha}(\mathcal V)
   +B\xi_{HU}(\alpha)\right]= \Psi_{S,\alpha}.
\end{equation}
Fix any candidate $\mathcal V\in\mathfrak V$.  If $B_{\mathcal V}>0$, then choosing the budget $B=B_{\mathcal V}$ makes $\mathcal V$ feasible in the inner infimum defining $\Phi_{S,\alpha}(B)$. Therefore,
$$
 \inf_{B>0}\Phi_{S,\alpha}(B)
 \leq \Phi_{S,\alpha}(B_{\mathcal V})
 \leq
 (1+B_{\mathcal V})A_{S,\alpha}(\mathcal V)
 +B_{\mathcal V}\xi_{HU}(\alpha).
$$
If instead $B_{\mathcal V}=0$, then $\mathcal V$ is feasible for every $B>0$, so
$$
 \Phi_{S,\alpha}(B)
 \leq
 (1+B)A_{S,\alpha}(\mathcal V)+B\xi_{HU}(\alpha)
 \qquad\text{for every }B>0.
$$
Letting $B\downarrow0$ gives the same conclusion.  Thus, for every $\mathcal V\in\mathfrak V$,
$$
 \inf_{B>0}\Phi_{S,\alpha}(B)
 \leq
 (1+B_{\mathcal V})A_{S,\alpha}(\mathcal V)
 +B_{\mathcal V}\xi_{HU}(\alpha).
$$
Taking the infimum over $\mathcal V$ yields
\begin{equation}\label{eq:indPhiUpper}
 \inf_{B>0}\Phi_{S,\alpha}(B)\leq\Psi_{S,\alpha}.
\end{equation}
For the reverse inequality, fix an arbitrary budget $B>0$.  Since the term $B\xi_{HU}(\alpha)$ does not depend on the candidate,
$$
 \Phi_{S,\alpha}(B)
 =
 \inf_{\substack{\mathcal V\in\mathfrak V\\B_{\mathcal V}\leq B}}
 \left\{
 (1+B)A_{S,\alpha}(\mathcal V)+B\xi_{HU}(\alpha)
 \right\}.
$$
Now take any candidate $\mathcal V$ in this constrained family.  Because $B_{\mathcal V}\leq B$ while $A_{S,\alpha}(\mathcal V)\geq0$ and $\xi_{HU}(\alpha)\geq0$,
$$
 (1+B)A_{S,\alpha}(\mathcal V)+B\xi_{HU}(\alpha)
 \geq
 (1+B_{\mathcal V})A_{S,\alpha}(\mathcal V)
 +B_{\mathcal V}\xi_{HU}(\alpha)
 \geq \Psi_{S,\alpha}.
$$
Taking the infimum over all candidates satisfying $B_{\mathcal V}\leq B$ therefore gives $\Phi_{S,\alpha}(B)\geq\Psi_{S,\alpha}$.  Since this holds for every $B>0$,
\begin{equation}\label{eq:indPhiLower}
 \inf_{B>0}\Phi_{S,\alpha}(B)\geq\Psi_{S,\alpha}.
\end{equation}
Combining \eqref{eq:indPhiUpper} and \eqref{eq:indPhiLower} proves \eqref{eq:indPhiIdentity}. Fix $\eta>0$. By the definition of an infimum, choose a deterministic $B_\eta>0$ such that
$$
 \Phi_{S,\alpha}(B_\eta)
 \leq \inf_{B>0}\Phi_{S,\alpha}(B)+\eta.
$$
Set $\lambda_\eta:=\lambda_{B_\eta}^\circ$.  Then \eqref{eq:indbudgetPhiProof} and \eqref{eq:indPhiIdentity} give \eqref{eq:indoraclebound}. \Halmos

\subsection{Proof of Proposition~\ref{prop:haar}}\label{app:haarproof}
The expectation below is over the Haar draw of $\mathcal S_2$. Let $P_1,P_2$ be the orthogonal projectors of $\mathcal S_1$ and $\mathcal S_2$ respectively. For every orthogonal matrix $O\in\R^{d\times d}$, Haar invariance implies that $O\mathcal S_2$ has the same distribution as $\mathcal S_2$.  If $S_2\in\R^{d\times r_0}$ has orthonormal columns spanning $\mathcal S_2$, then the orthogonal projector onto $O\mathcal S_2$ is $OP_2O^\top$ because $(OS_2)(OS_2)^\top =OS_2S_2^\top O^\top$. Consequently,
$$
 O\,\mathbb E[P_2]O^\top
 =\mathbb E[P_2]
 \qquad\text{for every orthogonal }O.
$$
The only matrices invariant under every orthogonal conjugation are scalar multiples of the identity, so $\mathbb E[P_2]=cI_d$ for some scalar $c$. Taking traces and using $\operatorname{tr}(P_2)=r_0$ almost surely gives
$$
 cd =\operatorname{tr}\{\mathbb E[P_2]\} =\mathbb E\{\operatorname{tr}(P_2)\} =r_0.
$$
Hence, $c=r_0/d$. Using the trace representation of the mean squared principal cosine and the fact that $P_1$ is fixed, we have
\begin{align*}
\mathbb E \left[A(r_0)\right] = 
 \mathbb E\left[\frac{1}{r_0}\operatorname{tr}(P_1P_2)\right]
 =\frac{1}{r_0}\operatorname{tr}\!\left(
   P_1\mathbb E[P_2]
   \right)=\frac{1}{d}\operatorname{tr}(P_1)
 =\frac{r_0}{d}. \Halmos
\end{align*}

\subsection{Proof of Theorem~\ref{thm:distalg}}\label{app:distalgproof}

Before proceeding with the proof of Theorem~\ref{thm:distalg}, we first establish a proposition. Denote $r_c(\widetilde u) :=\sum_{j=1}^m c_jq_j(\widetilde u)-q_\star(\widetilde u)$ for $c\in\mathcal C_{\mathrm{aff}}$ and the empirical counterpart of the DT population defect 
\begin{equation*}
 \widehat\Delta_{\mathrm{DT}}^{\mathrm{dist}}(c)
 :=
 \left\{
 \frac1n\sum_{i=1}^n
 \Dnorm{r_c(\widetilde u_i)}^2
 \right\}^{1/2}.
\end{equation*}
For fixed $(w,\pi)$, we write $\widehat P_j$ and $\overline P_j$ for $\widehat P_j(w,\pi)$ and $\overline P_j(w,\pi)$ respectively. Throughout, fix an eligible $(a,\pi^\circ)$ in Assumption~\ref{asm:distcomparator}, let $\nu:=\mathbb E_{Q_{DT}}[a(\widetilde u)^2]$ denote its transport variation, and define transport diffuseness
\begin{equation}\label{eq:distWoracle}
 W_{n,\delta}
 :=\left\{\nu+\frac{\bar a^2\log(42/\delta)}{n}\right\}^{1/2},
\end{equation}
which is $O(1)$ since $s_0^2\leq\nu\leq \bar a$.

\begin{proposition}[Distribution-level deterministic certificate]
\label{prop:diststruct}
Suppose Assumption~\ref{asm:commonlocal} holds and let $(w,\pi)$ be any feasible pair measurable with respect to (E1)--(E2). Then, for every $\alpha\in(0,1)$, conditional on (E1)--(E2), with probability at least $1-\alpha$ over (E3),
\begin{equation}\label{eq:diststruct}
\begin{aligned}
 D(P_\star,\widehat P_\star)
 &\leq
 \inf_{c\in\mathcal C_{\mathrm{aff}}}
 \Big[
 \sqrt m\norm c_2\,\overline\tau_{\mathrm{ref}}(w,\pi)
 +\Delta_{\mathrm{HU}}^{\mathrm{dist}}(c)
 +\sqrt n\norm w_2\,\widehat\Delta_{\mathrm{DT}}^{\mathrm{dist}}(c)
 \Big]\\
 &\qquad+\frac{\rho_D\left\{\sqrt K+\sqrt{2\log(2/\alpha)}\right\}}
 {\sqrt S}\norm w_2.
\end{aligned}
\end{equation}
\end{proposition}
\proof{Proof of Proposition~\ref{prop:diststruct}.}
Fix $c\in\mathcal C_{\mathrm{aff}}$. Since $\one_m^\top c=1$, direct expansion gives
$$
 P_\star-\overline P_\star
 =
 \sum_{j=1}^m c_j(P_j-\overline P_j)
 -P_+z(c)
 +\sum_{i=1}^nw_i
 \left\{
 \sum_{j=1}^m c_jq_j(\widetilde u_i)-q_\star(\widetilde u_i)
 \right\}.
$$
Taking $\Dnorm{\cdot}$ and applying the triangle inequality and Cauchy--Schwarz yield
\begin{align*}
 D(P_\star,\overline P_\star)
 &\leq
 \sum_{j=1}^m|c_j|D(P_j,\overline P_j)
 +\Dnorm{P_+z(c)}
 +\sum_{i=1}^nw_i\Dnorm{r_c(\widetilde u_i)}\\
 &\leq
 \sqrt m\norm c_2\,\overline\tau_{\mathrm{ref}}(w,\pi)
 +\Delta_{\mathrm{HU}}^{\mathrm{dist}}(c)
 +\sqrt n\norm w_2\,\widehat\Delta_{\mathrm{DT}}^{\mathrm{dist}}(c).
\end{align*}
Taking the infimum over $c\in\mathcal C_{\mathrm{aff}}$ gives a deterministic bound for $D(P_\star,\overline P_\star)$. It remains to control the fresh averaged DT target responses. Conditioned on (E1)--(E2), the weights are fixed and the $nS$ target draws are independent, with means $q_\star(\widetilde u_i)$. Put
$$
 F=D(\overline P_\star,\widehat P_\star)
 =\Dnorm{
 \sum_{i=1}^nw_i
 \{\widetilde Y_{i\star}-q_\star(\widetilde u_i)\}}.
$$
For $k\in[K]$, conditional independence and centering give
$$
 \mathbb E\left[
 \left\{
 \sum_{i=1}^nw_i
 (\widetilde Y_{i\star,k}-q_\star(\widetilde u_i,k))
 \right\}^2
 \right]
 =\frac1S\sum_{i=1}^nw_i^2q_\star(\widetilde u_i,k)
 \{1-q_\star(\widetilde u_i,k)\}.
$$
Thus, by the definition of $\rho_D$, Jensen's inequality, and Cauchy--Schwarz,
$$
 \mathbb E[F]
 \leq\frac{\rho_D\sqrt K}{\sqrt S}\norm w_2.
$$
If one of the $nS$ target draws, corresponding to twin $i$, is changed from one one-hot vector to another, the averaged target marginal changes by $(w_i/S)(e_k-e_\ell)$. Hence, the bounded-differences constant for that draw is at most $2\rho_Dw_i/S$. Since the sum of the squared constants is $4\rho_D^2\norm w_2^2/S$, the bounded-differences inequality \citep[e.g.,][Corollary~2.21]{Wainwright_2019} gives, for every $t>0$,
$$
 \mathbb P(|F-\mathbb E[F]|\geq t)
 \leq
 2\exp\!\left\{-\frac{St^2}{2\rho_D^2\norm w_2^2}\right\}.
$$
When $\norm w_2=0$, the target-sampling error is identically zero. Otherwise, take $t=\tfrac{\rho_D\sqrt{2\log(2/\alpha)}}{\sqrt S}\norm w_2$. Combining this bound with $D(P_\star,\widehat P_\star)\leq D(P_\star,\overline P_\star)+D(\overline P_\star,\widehat P_\star)$ proves \eqref{eq:diststruct}. \Halmos
\endproof

We now return to the proof of Theorem~\ref{thm:distalg}.

\paragraph{Step 1: A finite-sample population comparator.} 
For the fixed reachability witness, put
$$
 s=\mathbb E_{Q_{DT}}[a(\widetilde u)]
 =1-\one_K^\top\pi^\circ\in[s_0,1],
 \qquad
 P_j^\circ=\mathbb E_{Q_{DT}}[a(\widetilde u)q_j(\widetilde u)]+\pi^\circ,
 \quad j\in[d].
$$
Also put
\begin{equation*}
 S_n=\frac1n\sum_{i=1}^na(\widetilde u_i),
 \qquad
 V_n=\frac1n\sum_{i=1}^na(\widetilde u_i)^2.
\end{equation*}
On $S_n>0$, define
\begin{equation}\label{eq:distwcirc}
 w_i^\circ=\frac{s\,a(\widetilde u_i)}{nS_n};
\end{equation}
on $S_n=0$, define $w^\circ$ arbitrarily. We first show that, under \eqref{eq:distpopn}, with probability at least $1-\delta/4$ over (E1)--(E2), $(w^\circ,\pi^\circ)$ is feasible,
\begin{align}
 \widehat\tau_{\mathrm{ref}}(w^\circ,\pi^\circ)
 &\leq\tau_{\mathrm{ref}}+C_{\bar a,s_0}\varepsilon_n(\delta),
 \label{eq:distwfit}\\
 n\norm{w^\circ}_2^2
 &\leq C\left\{\nu+\frac{\bar a^2\log(42/\delta)}{n}\right\},
 \label{eq:distwprops}
\end{align}
and the empirical one-draw conditional covariance scale defined below satisfies
\begin{equation}\label{eq:distrefsigmaconcentration}
 \widehat\sigma_{\mathrm{ref}}
 \leq
 \sigma_{\mathrm{ref}}
 +C\sqrt{\frac{\log(42mK/\delta)}{n}}.
\end{equation}
Because $0\leq a\leq \bar a$, $\nu=\mathbb E_{Q_{DT}}[a(\widetilde u)^2] \leq \bar a\mathbb E_{Q_{DT}}[a(\widetilde u)]=As$. Because $\sum_i\operatorname{Var}(a(\widetilde u_i))\leq n\nu$, the bounded, two-sided scalar Bernstein inequality \citep[e.g.,][Theorem~2.9.5 and Exercise~2.47]{vershynin2026high} gives, with failure probability at most $\delta/16$ over (E1),
\begin{equation}\label{eq:distSconcentration}
 |S_n-s|
 \leq
 C\left\{
 \sqrt{\frac{\nu\log(42/\delta)}{n}}
 +\frac{\bar a\log(42/\delta)}{n}
 \right\}
 \leq\frac{s}{2}.
\end{equation}
For the empirical second moment, the same inequality applied to $a(\widetilde u_i)^2$ gives, with failure probability at most $\delta/16$ over (E1),
$$
 V_n
 \leq
 C\left\{\nu+\frac{\bar a^2\log(42/\delta)}{n}\right\}.
$$
Now stack the averaged observed reference responses:
$$
 U_{\mathrm{ref}}
 =\frac1n\sum_{i=1}^na(\widetilde u_i)
 (\widetilde Y_{i1},\ldots,
  \widetilde Y_{im})
 \in\R^{mK},
$$
and let
$$
 \mu_{\mathrm{ref}}
 =\bigl(
 \mathbb E_{Q_{DT}}[a(\widetilde u)q_1(\widetilde u)],\ldots,
 \mathbb E_{Q_{DT}}[a(\widetilde u)q_m(\widetilde u)]
 \bigr).
$$
Let
$$
 X_i^{\mathrm{ref}}
 :=a(\widetilde u_i)
   (\widetilde Y_{i1},\ldots,
    \widetilde Y_{im})\in\R^{mK},
 \qquad
 \xi_i:=X_i^{\mathrm{ref}}-\mu_{\mathrm{ref}}.
$$
Each averaged block is a probability vector, hence has Euclidean norm at most one. Therefore, $\norm{X_i^{\mathrm{ref}}}_2\leq \bar a\sqrt m$. Jensen's inequality similarly gives $\norm{\mu_{\mathrm{ref}}}_2 \leq\mathbb E\norm{X_i^{\mathrm{ref}}}_2\leq \bar a\sqrt m$, and consequently $\norm{\xi_i}_2\leq2\bar a\sqrt m$. Moreover, because $\xi_i$ is centered, $\mathbb E\norm{\xi_i}_2^2 \leq\mathbb E\norm{X_i^{\mathrm{ref}}}_2^2 \leq m\,\mathbb E[a(\widetilde u_i)^2] =m\nu$. Hilbert-space Bernstein \citep[e.g.,][Theorem~6.14]{steinwart2008support} therefore gives, with failure probability at most $\delta/16$ over (E1)--(E2),
\begin{equation}\label{eq:distUconcentration}
 \norm{U_{\mathrm{ref}}-\mu_{\mathrm{ref}}}_2
 \leq
 C\sqrt m
 \left\{
 \sqrt{\frac{\nu\log(42/\delta)}{n}}
 +\frac{\bar a\log(42/\delta)}{n}
 \right\}.
\end{equation}
Intersect the first three events. Equation~\eqref{eq:distSconcentration} implies $S_n\geq s/2$. Since $\sum_iw_i^\circ=s$ and $\one_K^\top\pi^\circ=1-s$, the pair $(w^\circ,\pi^\circ)$ is feasible. By \eqref{eq:distwcirc} and the definition of $U_{\mathrm{ref}}$, the stack of its observed reference marginals and the corresponding population-comparator stack are
$$
 \bigl(\widehat P_1(w^\circ,\pi^\circ),\ldots,
       \widehat P_m(w^\circ,\pi^\circ)\bigr)
 =\frac{s}{S_n}U_{\mathrm{ref}}+\one_m\otimes\pi^\circ, 
 \qquad 
 (P_1^\circ,\ldots,P_m^\circ)
 =\mu_{\mathrm{ref}}+\one_m\otimes\pi^\circ.
$$
Thus, the common dummy component cancels, and the stacked observed-minus-population comparator marginals equal $\tfrac{s}{S_n}U_{\mathrm{ref}}-\mu_{\mathrm{ref}} =\tfrac{s}{S_n}(U_{\mathrm{ref}}-\mu_{\mathrm{ref}}) +\tfrac{s-S_n}{S_n}\mu_{\mathrm{ref}}$. Since each block of $\mu_{\mathrm{ref}}$ has Euclidean norm at most $s$, \eqref{eq:distSconcentration} and \eqref{eq:distUconcentration} imply
$$
 \left\|\frac{s}{S_n}U_{\mathrm{ref}}-\mu_{\mathrm{ref}}\right\|_2
 \leq
 C\sqrt m
 \left\{
 \sqrt{\frac{\nu\log(42/\delta)}{n}}
 +\frac{\bar a\log(42/\delta)}{n}
 \right\}.
$$
Dividing the blockwise discrepancy by $\sqrt m$, then combining this with reachability and \eqref{eq:Dnormcomparison} blockwise gives \eqref{eq:distwfit}; substituting \eqref{eq:distwcirc} and the bound on $V_n$ gives \eqref{eq:distwprops}. It remains to compare the realized one-draw conditional reference covariance scale with its population counterpart. Put
\begin{equation*}
 \widehat\sigma_{\mathrm{ref}}^2
 :=\left\|
 \frac1n\sum_{i=1}^n\Gamma_{\mathrm{ref}}(\widetilde u_i)
 \right\|_{\op}
\end{equation*}
and let
$$
 \overline\Gamma_{\mathrm{ref}}
 :=\mathbb E_{Q_{DT}}[\Gamma_{\mathrm{ref}}(\widetilde u)],
 \qquad
 X_i:=\Gamma_{\mathrm{ref}}(\widetilde u_i)
      -\overline\Gamma_{\mathrm{ref}}.
$$
Because $0\preceq\Gamma_{\mathrm{ref}}(\widetilde u)\preceq\frac12I_{mK}$, we have $\norm{X_i}_{\op}\leq1/2$. Furthermore, $\mathbb E[X_i^2] =\mathbb E[\Gamma_{\mathrm{ref}}(\widetilde u)^2] -\overline\Gamma_{\mathrm{ref}}^2 \preceq \mathbb E[\Gamma_{\mathrm{ref}}(\widetilde u)^2] \preceq\frac12\overline\Gamma_{\mathrm{ref}}$, so $\left\|\sum_{i=1}^n\mathbb E[X_i^2]\right\|_{\op} \leq\tfrac n2\sigma_{\mathrm{ref}}^2$. The self-adjoint matrix Bernstein inequality \citep[e.g.,][Theorem~5.4.1]{vershynin2026high} therefore gives, with failure probability at most $\delta/16$ over (E1),
$$
 \left\|\frac1n\sum_{i=1}^nX_i\right\|_{\op}
 \leq
 C\left\{
 \sigma_{\mathrm{ref}}
 \sqrt{\frac{\log(42mK/\delta)}{n}}
 +\frac{\log(42mK/\delta)}{n}
 \right\}.
$$
Consequently,
\begin{align*}
 \widehat\sigma_{\mathrm{ref}}^2
 &\leq
 \sigma_{\mathrm{ref}}^2
 +C\left\{
 \sigma_{\mathrm{ref}}
 \sqrt{\frac{\log(42mK/\delta)}{n}}
 +\frac{\log(42mK/\delta)}{n}
 \right\}\leq
 \left\{
 \sigma_{\mathrm{ref}}
 +C\sqrt{\frac{\log(42mK/\delta)}{n}}
 \right\}^2,
\end{align*}
which proves \eqref{eq:distrefsigmaconcentration} after possibly redefining the constant $C$. Intersecting this fourth event with the first three completes the Step~1 event, which has probability at least $1-\delta/4$ over (E1)--(E2).

\paragraph{Step 2: Uniform reference generalization.} 
For $i\leq n$ and $j\leq m$, let
$$
 \overline E_{ij}
 :=\widetilde Y_{ij}-q_j(\widetilde u_i)\in\mathcal H_0,
 \qquad
 \zeta_i=(\overline E_{i1}^\top,\ldots,\overline E_{im}^\top)^\top,
$$
and let $Z_{\mathrm{ref}}\in\R^{n\times mK}$ have row $i$ equal to $\zeta_i^\top$. By construction, $\mathbb E[\zeta_i\zeta_i^\top\mid\widetilde u_i]=\tfrac1S\Gamma_{\mathrm{ref}}(\widetilde u_i)$. For each reference question, put
$$
 d_j:=\widehat P_j(w,\pi)-\overline P_j(w,\pi)
     =\sum_{i=1}^nw_i\overline E_{ij}\in\mathcal H_0.
$$
The reverse triangle inequality for $\Dnorm{\cdot}$ gives $\left|D(P_j,\overline P_j)-D(P_j,\widehat P_j)\right| \leq\Dnorm{d_j}$. Applying the reverse triangle inequality once more to the Euclidean norms of the $m$ question-level discrepancies, followed by \eqref{eq:Dnormcomparison}, yields deterministically and simultaneously for all feasible $(w,\pi)$,
\begin{align}
 \left|\overline\tau_{\mathrm{ref}}(w,\pi)
  -\widehat\tau_{\mathrm{ref}}(w,\pi)\right|
  &\leq
  \frac1{\sqrt m}\left\{\sum_{j=1}^m\Dnorm{d_j}^2\right\}^{1/2}\leq\rho_D\sqrt{\frac Km}
  \left\{\sum_{j=1}^m\norm{d_j}_2^2\right\}^{1/2}\notag\\
  &=\rho_D\sqrt{\frac Km}\norm{Z_{\mathrm{ref}}^\top w}_2\leq\rho_D\sqrt{\frac Km}\norm{Z_{\mathrm{ref}}}_{\op}\norm w_2.\label{eq:distrefgen}
\end{align}
Conditional on (E1), we claim that, with probability at least $1-\delta/4$ over (E2),
\begin{equation}\label{eq:distZboundhat}
 \frac{\norm{Z_{\mathrm{ref}}}_{\op}}{\sqrt n}
 \leq
 C\left\{
 \frac{\widehat\sigma_{\mathrm{ref}}}{\sqrt S}
 +\sqrt{\frac{m\log(42mK/\delta)}{n}}
 \right\}.
\end{equation}
To prove this, note first that $\norm{\zeta_i}_2^2\leq2m$, while the conditional covariance of row $i$ is $S^{-1}\Gamma_{\mathrm{ref}}(\widetilde u_i)$. Put
$$
 X_i=\zeta_i\zeta_i^\top -\mathbb E[\zeta_i\zeta_i^\top\mid\widetilde u_i].
$$
Then, $\norm{X_i}_{\op}\leq4m$ and, using $(\zeta_i\zeta_i^\top)^2\preceq2m\zeta_i\zeta_i^\top$, $\left\|\sum_{i=1}^n \mathbb E[X_i^2\mid\widetilde u_i]\right\|_{\op} \leq\tfrac{2mn}{S}\widehat\sigma_{\mathrm{ref}}^2$. Matrix Bernstein \citep[e.g.,][Theorem~5.4.1]{vershynin2026high} therefore gives
$$
 \left\|\sum_iX_i\right\|_{\op}
 \leq
 C\sqrt{\frac{mn\widehat\sigma_{\mathrm{ref}}^2
                 \log(42mK/\delta)}{S}}
 +Cm\log(42mK/\delta).
$$
Since the conditional mean of $Z_{\mathrm{ref}}^\top Z_{\mathrm{ref}}$ has operator norm $n\widehat\sigma_{\mathrm{ref}}^2/S$, the triangle inequality and $2\sqrt{ab}\leq a+b$ yield $\norm{Z_{\mathrm{ref}}}_{\op}^2 / n \leq C(\widehat\sigma_{\mathrm{ref}}^2 / S + m\log(42mK/\delta) / n)$, which proves \eqref{eq:distZboundhat}. On the Step~1 event \eqref{eq:distrefsigmaconcentration}, since $S,m\geq1$,
\begin{equation}\label{eq:distZbound}
 \frac{\norm{Z_{\mathrm{ref}}}_{\op}}{\sqrt n}
 \leq
 C\left\{
 \frac{\sigma_{\mathrm{ref}}}{\sqrt S}
 +\sqrt{\frac{m\log(42mK/\delta)}{n}}
 \right\}.
\end{equation}
Combining \eqref{eq:distrefgen} and \eqref{eq:distZbound} gives, on the same event,
\begin{equation}\label{eq:distrefgenW}
 \overline\tau_{\mathrm{ref}}(w,\pi)
 \leq\widehat\tau_{\mathrm{ref}}(w,\pi)
 +C\rho_D\sqrt K
 \left\{
 \frac{\sigma_{\mathrm{ref}}}{\sqrt{mS}}
 +\sqrt{\frac{\log(42mK/\delta)}{n}}
 \right\}
 \sqrt n\norm w_2
\end{equation}
simultaneously for every feasible pair.

\paragraph{Step 3: Oracle ridge comparison and transfer.} 
On the Step~1 event, \eqref{eq:distwfit}, \eqref{eq:distwprops}, and the definition \eqref{eq:distWoracle} give
$$
 \widehat\tau_{\mathrm{ref}}(w^\circ,\pi^\circ)
 \leq C_{\bar a,s_0}\left\{\tau_{\mathrm{ref}}
 +\varepsilon_n(\delta)\right\},
 \qquad
 \sqrt n\norm{w^\circ}_2\leq C W_{n,\delta}.
$$
Set
\begin{equation}\label{eq:distlambdaoracle}
 \lambda^\circ
 :=\frac{n\left[\tau_{\mathrm{ref}}+\varepsilon_n(\delta)\right]^2}{W_{n,\delta}^2}
 =\frac{\left[\tau_{\mathrm{ref}}+\varepsilon_n(\delta)\right]^2 n^2}
      {\bar a^2\log(42/\delta) + \nu n},
\end{equation}
which balances reference mismatch against transport variation. Ridge optimality against $(w^\circ,\pi^\circ)$ then gives
\begin{align*}
 \widehat\tau_{\mathrm{ref}}(\widehat w_{\lambda^\circ},
                   \widehat\pi_{\lambda^\circ})^2
 +\lambda^\circ\norm{\widehat w_{\lambda^\circ}}_2^2
 &\leq
 C_{\bar a,s_0}\left\{\tau_{\mathrm{ref}}
 +\varepsilon_n(\delta)\right\}^2
 +C\lambda^\circ W_{n,\delta}^2/n\lesssim_{\bar a,s_0}\left\{\tau_{\mathrm{ref}}
 +\varepsilon_n(\delta)\right\}^2.
\end{align*}
Dropping each nonnegative term in turn yields
\begin{equation}\label{eq:distoracleproofbounds}
 \widehat\tau_{\mathrm{ref}}(\widehat w_{\lambda^\circ},
                  \widehat\pi_{\lambda^\circ})
 \lesssim_{\bar a,s_0}
 \left\{\tau_{\mathrm{ref}}+\varepsilon_n(\delta)\right\},
 \qquad
 \sqrt n\norm{\widehat w_{\lambda^\circ}}_2
 \lesssim_{\bar a,s_0}W_{n,\delta}.
\end{equation}
Combining \eqref{eq:distrefgenW} and \eqref{eq:distoracleproofbounds}, on the intersection of the Step~1--2 events,
\begin{align}
 \overline\tau_{\mathrm{ref}}(\widehat w_{\lambda^\circ},
                    \widehat\pi_{\lambda^\circ})
 \lesssim_{\bar a,s_0}{}&
 \tau_{\mathrm{ref}}+\varepsilon_n(\delta)
 +\rho_D\sqrt K\,W_{n,\delta}
 \left\{
 \frac{\sigma_{\mathrm{ref}}}{\sqrt{mS}}
 +\sqrt{\frac{\log(42mK/\delta)}{n}}
 \right\}. \label{eq:distoraclelatentfit} 
\end{align}
The Step~1 event has probability at least $1-\delta/4$ over (E1)--(E2), while, conditional on (E1), the Step~2 event has probability at least $1-\delta/4$ over (E2). Hence, their intersection has probability at least $1-\delta/2$ over (E1)--(E2). By Cauchy--Schwarz, every $c\in\mathcal C_{\mathrm{aff}}$ satisfies $\norm c_2\geq m^{-1/2}$. The logarithm in $\varepsilon_n(\delta)$ dominates both $1$ and $\log(8/\delta)$, so
\begin{align}
 \rho_D\sqrt K
 \sqrt{\frac{\log(42mK/\delta)}{n}}
 &\leq C\varepsilon_n(\delta),
 \label{eq:distfiniteabsorb}\\
 \frac{\rho_D\{\sqrt K+\sqrt{2\log(8/\delta)}\}}
      {\sqrt{nS}}
 &\leq C\sqrt m\norm c_2\,\varepsilon_n(\delta).
 \label{eq:disttargetabsorb}
\end{align}
We next relate the empirical DT affine defect to its population counterpart. Fix a deterministic $c\in\mathcal C_{\mathrm{aff}}$ and put $Z_i(c)=\Dnorm{r_c(\widetilde u_i)}^2$. Since $q_j(\widetilde u)$ and $q_\star(\widetilde u)$ are probability vectors, $\Dnorm{r_c(\widetilde u)} \leq2\rho_D\sqrt m\norm c_2 =:B_c$. Therefore, $0\leq Z_i(c)\leq B_c^2$, $\mathbb EZ_i(c)=\Delta_{\mathrm{DT}}^{\mathrm{dist}}(c)^2$, and $\operatorname{Var}(Z_i(c))\leq B_c^2\Delta_{\mathrm{DT}}^{\mathrm{dist}}(c)^2$. The bounded scalar Bernstein \citep[e.g.,][Theorem~2.9.5 and Exercise~2.47]{vershynin2026high} gives, with failure probability at most $\delta/4$ over (E1),
\begin{equation}\label{eq:distrmsconcentration}
 \widehat\Delta_{\mathrm{DT}}^{\mathrm{dist}}(c)
 \leq
 \Delta_{\mathrm{DT}}^{\mathrm{dist}}(c)
 +C\sqrt m\norm c_2\,\varepsilon_n(\delta).
\end{equation}
Define the deterministic oracle criterion
\begin{align*}
 G^\circ(c)
 :=\;&\sqrt m\norm c_2\tau_{\mathrm{ref}}
 +\Delta_{\mathrm{HU}}^{\mathrm{dist}}(c)
 +W_{n,\delta}\Delta_{\mathrm{DT}}^{\mathrm{dist}}(c)+\rho_D\sqrt K\norm c_2W_{n,\delta}
   \frac{\sigma_{\mathrm{ref}}}{\sqrt S}
 +\sqrt m\norm c_2(1+W_{n,\delta})\varepsilon_n(\delta).
\end{align*}
For any $\eta>0$, choose a deterministic $c_\eta\in\mathcal C_{\mathrm{aff}}$ such that
$$
 G^\circ(c_\eta)
 \leq\inf_{c\in\mathcal C_{\mathrm{aff}}}G^\circ(c)+\eta.
$$
Intersect the (E1)--(E2) Step~1--2 event with the (E1) event \eqref{eq:distrmsconcentration} for this fixed $c_\eta$, and then apply Proposition~\ref{prop:diststruct} with $(w,\pi)=(\widehat w_{\lambda^\circ},\widehat\pi_{\lambda^\circ})$ and $\alpha=\delta/4$ over (E3). The total failure probability over (E1)--(E3) is at most $\delta$. Using \eqref{eq:distoracleproofbounds}, \eqref{eq:distoraclelatentfit}, \eqref{eq:distrmsconcentration}, \eqref{eq:distfiniteabsorb}, and \eqref{eq:disttargetabsorb} gives
$$
 D(P_\star,\widehat P_{\star,\lambda^\circ})
 \leq C_{\bar a,s_0}G^\circ(c_\eta).
$$
The RMS contribution is bounded by
$$
 \sqrt n\norm{\widehat w_{\lambda^\circ}}_2
 \widehat\Delta_{\mathrm{DT}}^{\mathrm{dist}}(c_\eta)
 \lesssim_{\bar a,s_0}
 W_{n,\delta}
 \left\{
 \Delta_{\mathrm{DT}}^{\mathrm{dist}}(c_\eta)
 +\sqrt m\norm{c_\eta}_2\varepsilon_n(\delta)
 \right\},
$$
and all remaining finite-$n$ terms are absorbed by $\sqrt m\norm{c_\eta}_2(1+W_{n,\delta})\varepsilon_n(\delta)$. Take $\eta_k=2^{-k}$. The events
$$
 \left\{
 D(P_\star,\widehat P_{\star,\lambda^\circ})
 \leq C_{\bar a,s_0}\left(
 \inf_{c\in\mathcal C_{\mathrm{aff}}}G^\circ(c)+\eta_k
 \right)\right\}
$$
are decreasing and each has probability at least $1-\delta$. Continuity of probability from above therefore yields \eqref{eq:distoraclebound}. \Halmos

\end{document}